\documentclass[runningheads]{llncs}

\usepackage[mobile]{eccv}

\usepackage{eccvabbrv}

\usepackage{graphicx}
\usepackage{booktabs}
\usepackage{multirow}

\usepackage[accsupp]{axessibility}  

\usepackage{wrapfig}
\usepackage{xr}
\usepackage[pagebackref,breaklinks,colorlinks,citecolor=eccvblue]{hyperref}

\usepackage{orcidlink}

\begin{document}

\title{Fed3DGS: Scalable 3D Gaussian Splatting with Federated Learning}


\author{Teppei Suzuki}

\authorrunning{T.~Suzuki}

\institute{Denso IT Laboratory, Inc.}

\maketitle

\begin{abstract}
In this work, we present Fed3DGS, a scalable 3D reconstruction framework based on 3D Gaussian splatting (3DGS) with federated learning.
Existing city-scale reconstruction methods typically adopt a centralized approach, which gathers all data in a central server and reconstructs scenes.
The approach hampers scalability because it places a heavy load on the server and demands extensive data storage when reconstructing scenes on a scale beyond city-scale.
In pursuit of a more scalable 3D reconstruction, we propose a federated learning framework with 3DGS, which is a decentralized framework and can potentially use distributed computational resources across millions of clients.
We tailor a distillation-based model update scheme for 3DGS and introduce appearance modeling for handling non-IID data in the scenario of 3D reconstruction with federated learning.
We simulate our method on several large-scale benchmarks, and our method demonstrates rendered image quality comparable to centralized approaches.
In addition, we also simulate our method with data collected in different seasons, demonstrating that our framework can reflect changes in the scenes and our appearance modeling captures changes due to seasonal variations.
The code is available at \url{https://github.com/DensoITLab/Fed3DGS}.
  \keywords{federated learning \and neural fields \and 3D reconstruction}
\end{abstract}

\section{Introduction}
\label{sec:intro}
Since differentiable rendering~\cite{kato2018neural,niemeyer2020differentiable,kopanas2021point} has emerged, 3D reconstruction with neural fields~\cite{xie2022neural} has attracted considerable interest.
Neural fields can render photo-realistic images, leading to a diverse range of applications, such as localization~\cite{yen2021inerf,lin2023parallel,maggio2023loc}, navigation~\cite{kwon2023renderable,adamkiewicz2022vision}, and physics simulation~\cite{li2022climatenerf,li2023pac,xie2023physgaussian,wang2023neural}.

Recently, city-scale 3D reconstruction using neural fields~\cite{tancik2022block,Turki_2022_CVPR,turki2023suds} has been explored for its broad applications, such as urban planning and VR/AR applications.
Many studies employ centralized approaches, wherein all data are gathered at the central server for scene reconstruction.
However, these approaches pose challenges as they would place a heavy load on the server and require extensive data storage, particularly when reconstructing scenes on a scale beyond city-scale.
For instance, Block-NeRF~\cite{tancik2022block} trains 35 models with 2.8M images for an area of only 960 m $\times$ 570 m.
Moreover, many existing studies do not focus on the maintenance of the model to reflect changes in the scene, a crucial aspect for various applications such as localization.

\begin{figure}[t]
    \centering
    \includegraphics[clip,width=0.95\hsize]{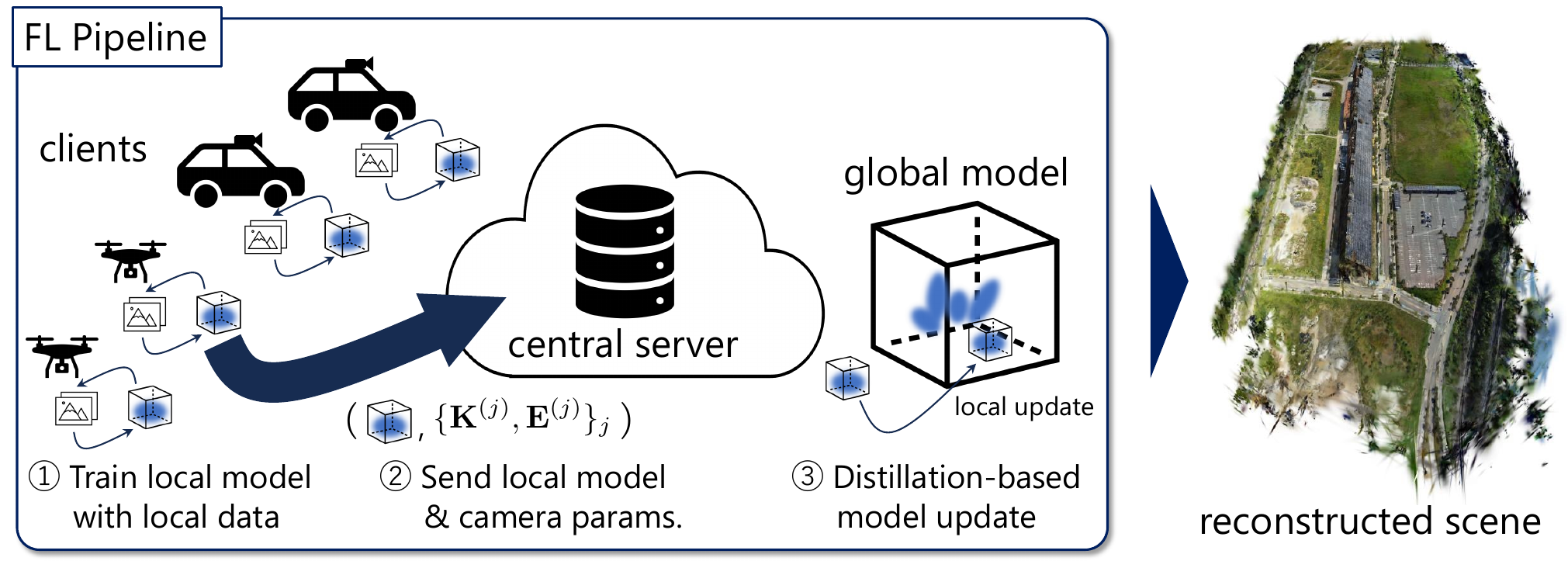}
    \caption{The proposed federated learning framework. Clients collaboratively train a global model under the orchestration of a central server. Our framework continuously updates the global model by repeating steps 1 to 3.}
    \label{fig:overview}
\end{figure}

To achieve more scalable and maintainable photo-realistic 3D reconstruction, we propose Fed3DGS, a federated learning framework with 3D Gaussian splatting (3DGS)~\cite{3dgs}.
We show the proposed framework in Fig. \ref{fig:overview}.
In our framework, many clients collaboratively reconstruct 3D scenes under the orchestration of a central server.
Consequently, the framework has the potential to leverage millions of distributed computing resources and data storage, surpassing the total FLOPS of a supercomputer.
In addition, our framework allows for the continuous updating of the global model through iterative model update processes, improving model maintainability.

In this work, we specifically address two challenges in a large-scale 3D reconstruction with federated learning:
(i) The global model should be scalable in terms of model size for representing a large-scale scene.
Previous work~\cite{fednerf} uses a voxel grid representation, and it requires significant storage even for relatively small scenes.
To improve its scalability in terms of model size, we adopt 3DGS~\cite{3dgs} to represent scenes and propose a distillation-based model update scheme tailored for 3DGS.
(ii) The appearance varies for each client because clients may collect data at different times or seasons, and it affects the quality of 3D reconstruction.
To mitigate this, we introduce an appearance modeling scheme into our federated learning framework.

We validate the effectiveness of Fed3DGS on large-scale datasets, such as Mill 19~\cite{Turki_2022_CVPR}, UrbanScene3D~\cite{UrbanScene3D}, and Quad 6k~\cite{quad6k}.
Our method demonstrates rendered image quality comparable to baseline methods, such as Mega-NeRF~\cite{turki2023suds} and Switch-NeRF~\cite{mi2023switchnerf}, in terms of SSIM~\cite{ssim} and LPIPS~\cite{lpips}.
Moreover, our method has a smaller global model size and takes shorter training time per client, compared to the federated learning baseline with NeRF~\cite{fednerf}.
In the context of 3D reconstruction in a continuous update scenario, we also evaluate our framework on 4Seasons~\cite{wenzel2020fourseasons}, which collects data across different seasons.
In this experiment, we show that Fed3DGS can reflect changes in the scene and effectively model appearance changes resulting from seasonal variations.

Our contribution is summarized as follows:
(i) We propose Fed3DGS, a federated learning framework with 3DGS.
We design a distillation-based model update protocol for 3DGS.
(ii) We propose an appearance modeling scheme for 3DGS.
We introduce an appearance model into 3DGS, and it is also updated in a federated learning manner.
(iii) We assess the effectiveness of our framework on several datasets.
We show that Fed3DGS surpasses the federated learning baseline~\cite{fednerf} in terms of the scalability and rendered image quality.
Furthermore, we show that our method can reflect changes in the scene and effectively model appearance changes resulting from seasonal variations.

\section{Related Work}
\label{sec:related-work}
\subsection{Large-Scale 3D Reconstruction}
3D reconstruction is a fundamental problem in computer vision~\cite{agarwal2011building,snavely2006photo,hartley2003multiple,schoenberger2016sfm,schoenberger2016mvs,lindenberger2021pixsfm,zhang2023revisiting,kato2020differentiable}.
It is basically performed using the multiple view geometry to reconstruct scenes as 3D point clouds.

Although point clouds serve as a common representation for 3D scenes,
the emergence of differentiable rendering~\cite{kato2018neural,nerf,niemeyer2020differentiable,kopanas2021point} has popularized the use of neural fields~\cite{xie2022neural} as 3D representations.
Neural fields have gained prominence due to their ability to render high-quality images~\cite{barron2021mip,barron2022mip} and represent smooth surfaces~\cite{wang2021neus,yariv2023bakedsdf}.
For their advantages, many applications using neural fields have been explored~\cite{sucar2021imap,maggio2023loc,kwon2023renderable,adamkiewicz2022vision,li2022climatenerf,li20223d}.

Building city-scale 3D models poses a significant challenge and has garnered considerable attention due to its wide-ranging applications, such as urban planning, simulation for autonomous driving, and VR/AR systems.
Numerous studies have addressed city-scale reconstruction~\cite{fruh2004automated,snavely2006photo,agarwal2011building,rematas2022urban}.
Recent studies use NeRF~\cite{nerf} with a distributed training approach for photo-realistic scene reconstruction~\cite{tancik2022block,Turki_2022_CVPR,jia2024drone,turki2023suds}.
They divide scenes into small regions and then model the regions by multiple NeRF models.
This strategy efficiently models a large scene by training multiple models in parallel.
However, since the computational costs and data storage increase with the size of the scene, the centralized approaches, including distributed approaches~\cite{tancik2022block,Turki_2022_CVPR,jia2024drone,turki2023suds}, limit their scalability due to the cost of enlarging computational resources and data storage of the central server.

To develop a more scalable framework, Suzuki~\cite{fednerf} proposed a federated learning framework for 3D reconstruction with NeRF (we refer to it as FedNeRF in this work), which can leverage computational resources of millions of clients.
FedNeRF uses a voxel grid representation, similar to Plenoxels~\cite{plenoxels}, to approach problems specific to 3D reconstruction with federated learning described in Sec. \ref{sec:problem}.
However, since the voxel grid requires large disk space, it is unacceptable for large-scale scene reconstruction beyond the city-scale.
In addition, FedNeRF is not designed for non-IID data, signifying that the appearance of training images varies among clients capturing images at different times and/or seasons.

\subsection{Federated Learning}
Federated learning (FL)~\cite{mcmahan2017communication,kairouz2021advances} is a distributed training pipeline where many clients (\textit{e.g.}, mobile devices or whole organizations) collaboratively train a model, while keeping the training data decentralized.
Since FL can leverage computational resources and data of mullions of clients with preserving their privacy, FL is now being applied in many applications~\cite{hard2018federated,yang2018applied,ramaswamy2019federated}.

FL roughly consists of five steps:
(1) \textit{client selection} that selects clients who will participate in a model update round,
(2) \textit{broadcast} that distributes the current global model weights to the selected clients,
(3) \textit{local model training} that trains local models on the clients,
(4) \textit{aggregation} that collects the computational results of clients, and
(5) \textit{model update} that updates the global model with aggregated local results.
Our framework basically follows these steps except for ignoring step 2 because clients train local models from scratch in our framework.

FL has many challenging problems.
For example, the local data are often non-IID, and a simple FL algorithm~\cite{mcmahan2017communication} does not work for the non-IID data~\cite{zhao2018federated}.
Thus, the FL for the non-IID data is widely studied~\cite{zhao2018federated,li2021fedbn,karimireddy2020scaffold}.
Similar to the non-IID problem (\textit{i.e.}, data heterogeneity), heterogeneity in computing capacities of clients is also a challenging problem, which delays the aggregation step due to waiting for the clients with poor resources.
Some studies have addressed this problem~\cite{lin2020ensemble,zhu2021data}; for example, Nishio and Yonetani~\cite{nishio2019client} proposed client selection strategy to manage the resources of heterogeneous clients, and Xie \textit{et al.}~\cite{xie2019asynchronous} introduced an asynchronous federated learning protocol, which updates the global model as soon as the central server receives a local model without waiting for other clients.
In terms of the costs associated with training a global model, the bottleneck often lies in transmitting local results, such as model parameters or gradients, to the central server.
This challenge is more pronounced than the costs associated with training local models for federated learning, especially in the context of cross-device federated learning.
Some studies have proposed compression techniques to improve communication efficiency~\cite{konevcny2016federated,lin2017deep}.

In addition to the above problems, there are some problems specific to 3D reconstruction with FL, as pointed out in \cite{fednerf}.
We discuss it in the next section.

\section{Problem Definition}
\label{sec:problem}
We show our federated learning pipeline in Fig. \ref{fig:overview}.
Each client reconstructs a part of scenes from its local data $\mathcal{D}_l=\{I^{(j)}, \mathbf{K}^{(j)}, \mathbf{E}^{(j)}\}_j$, where $I^{(j)}$ denotes a $j$-th image, and $\mathbf{K}^{(j)}$ and $\mathbf{E}^{(j)}$ denote intrinsic and extrinsic matrices corresponding to a $j$-th image.
Clients transmit their local 3DGS model, $\mathcal{G}_l$, and camera parameters, $\mathcal{C}_l=\{\mathbf{K}^{(j)}, \mathbf{E}^{(j)}\}_j$, to the central server.
Then, the central server updates the global model based on data transmitted from clients.
As in FedNeRF~\cite{fednerf}, we assume an asynchronous setting~\cite{xu2023asynchronous}; namely, the global model is updated as soon as the central server receives a local model without waiting for other clients' computation.
In addition, the global model is not distributed to clients; instead, clients reconstruct scenes from scratch.
That is because it simplifies the pipeline, and initialization is relatively not important for 3D reconstruction, compared to the common problems addressed in FL, such as classification.
Note that we can distribute the global model to clients, although this would increase communication costs while decreasing computational costs on clients.

We assume that clients are vehicles, such as cars and drones, in this work.
Each client is equipped with a monocular camera to capture scenes and sensors to determine its global pose, such as a GPS, an IMU, and a compass.
We further assume that the camera is calibrated, meaning that the camera intrinsic matrix is known.
Also, clients have the capability to determine relative poses between images either through SfM techniques or information obtained from sensors.

In federated learning, there are many challenges, such as privacy, communication efficiency, and heterogeneity in clients' computing capacities and data~\cite{kairouz2021advances}.
In addition, there are several challenges specific to 3D reconstruction with federated learning:
(i) \textbf{Global pose ambiguity.} The local coordinates and scale between clients are not aligned even with sensors, such as a GPS, due to their noise.
Thus, the central server needs to align them before the model update step.
(ii) \textbf{Appearance diversity.}
The appearance of images may vary for each client though it would be the same in the local data because each client may collect data at different times or different seasons.
This is a common non-IID setting in federated learning, but 3D reconstruction with various appearance is a challenging problem~\cite{nerfw,meshry2019neural,radenovic2016dusk}.
In particular, unlike the centralized approaches, each client does not access other clients' data in federated learning to obtain images with different appearance, making it more difficult to handle differences in appearance.
(iii) \textbf{Locality of model update.} Since a standard 3D reconstruction task models only observed scenes, the global model should only be updated for areas observed by clients.

FedNeRF~\cite{fednerf} addresses challenges related to global pose ambiguity and locality of updates.
The global pose alignment proposed by FedNeRF also works well for our method, as shown in the Appendix \ref{sec:add-ablation}.
However, the appearance diversity problem still remains and FedNeRF has a limitation in terms of scalability of the global model size;
it represents the global model with the voxel-based representation~\cite{plenoxels} for realizing local updates, but it requires large disk space even if the scene is not very large.

In this work, we focus on addressing the appearance diversity problem and improving scalability in terms of the global model size while satisfying the locality of model updates.
To enhance the scalability, we adopt 3DGS~\cite{3dgs}.
Furthermore, we incorporate an appearance modeling scheme into 3DGS and propose a model update framework for 3DGS in a federate learning manner.

It is important to note that, in this work, our focus is not on addressing other issues in federated learning.
Specifically, we assume that the computational resources and bandwidth of all clients are sufficient to train the model and transmit it to the central server.
Addressing more challenging settings and additional concerns will be future work.

\section{Methodology}
We adopt 3DGS ~\cite{3dgs} to represent scenes because it is point-based representation and does not need to allocate memory to empty regions, which is more efficient than the voxel-based representation used in FedNeRF~\cite{fednerf}.
Besides, 3DGS is an explicit representation, which allows us to locally update the model.
We briefly review 3DGS in Sec. \ref{sec:3dgs}.
To capture diverse appearance, we introduce an appearance model inspired by NeRF-W~\cite{nerfw} into 3DGS in Sec. \ref{sec:app}.
For federated learning, we propose a global model update scheme tailored to 3DGS in Sec. \ref{sec:merge}.

\subsection{3D Gaussian Splatting}
\label{sec:3dgs}
We briefly review 3DGS~\cite{3dgs}.
For more details, please refer to \cite{3dgs}.

Let $\mathcal{G}=\{G^{(i)}\}_i$ be a set of 3D Gaussians.
An $i$-th Gaussian, $G^{(i)}$, has four parameters: $G^{(i)}=(\mathbf{x}^{(i)},\mathbf{\Sigma}^{(i)},k^{(i)},o^{(i)})$, where $\mathbf{x}^{(i)}=(x,y,z)$ denotes a 3D position, $\mathbf{\Sigma}^{(i)}\in\mathbb{R}^{3\times3}$ denotes a covariance matrix, $k^{(i)}\in\mathbb{R}^{l}$ denotes coefficients of spherical harmonics, and $o^{(i)}\in\mathbb{R}$ denotes a opacity logit.
3DGS renders the novel view image, $\hat{I}$, through a differentiable renderer, $R$, as follows:
\begin{align}
    \hat{I}=R(\mathbf{K}, \mathbf{E}, \mathcal{G}),
\end{align}
where $\mathbf{K}\in\mathbb{R}^{3\times3}$ and $\mathbf{E}\in\mathbb{R}^{3\times4}$ are camera intrinsic and extrinsic matrices.
The rendering function, $R$, corresponds to the following eqs. \eqref{eq:pos_proj}--\eqref{eq:blend}.

To render images from 3D Gaussians, we first project a 3D Gaussian to 2D as follows:
\begin{align}
    \label{eq:pos_proj}
    \mathbf{x}^\text{2D}=\mathbf{K}((\mathbf{Ex})/(\mathbf{Ex})_z),\\
    \label{eq:cov_proj}
    \mathbf{\Sigma}^\text{2D}=\mathbf{J}\mathbf{W}\mathbf{\Sigma}\mathbf{W}^\top\mathbf{J}^\top,
\end{align}
where $(\mathbf{Ex})_z$ denotes the value of $\mathbf{Ex}$ along the z-axis (\ie, the view direction); $\mathbf{J}\in\mathbb{R}^{3\times3}$ is the Jacobian of the affine approximation of the projective transformation~\cite{zwicker2002ewa}; $\mathbf{W}\in\mathbb{R}^{3\times3}$ denotes a viewing transformation, which corresponds to a rotation matrix of the camera extrinsic in this case.
After 2D projection, the parameter of $\alpha$-blending is computed as follows:
\begin{align}
    \label{eq:coef}
    \alpha_\mathbf{p}=\sigma(o)\exp\left(-\frac{1}{2}(\mathbf{p}-\mathbf{x}^\text{2D})^\top\mathbf{\Sigma}^\text{2D}(\mathbf{p}-\mathbf{x}^\text{2D})\right),
\end{align}
where $\sigma(\cdot)$ denotes the sigmoid function and $\mathbf{p}\in\mathbb{R}^2$ is a pixel coordinate.
Then, the pixel color at $\mathbf{p}$ is rendered through the following point-based $\alpha$-blending with $\mathcal{N}$ ordered points overlapping the pixel as follows:
\begin{align}
    \label{eq:blend}
    C(\mathbf{p})=\sum_{i\in\mathcal{N}}c^{(i)}\alpha^{(i)}_\mathbf{p}\prod_{j=1}^{i-1}(1-\alpha^{(j)}_\mathbf{p}),
\end{align}
where $\alpha^{(i)}_\mathbf{p}$ is computed with the $i$-th Gaussian, $G^{(i)}$, through eqs. \eqref{eq:pos_proj}--\eqref{eq:coef}, and $c^{(i)}$ denotes the RGB color computed from the spherical harmonics with $k^{(i)}$.

The parameters of Gaussians are optimized by solving the following problem using the stochastic gradient descent:
\begin{align}
    \label{eq:3dgs-obj}
    \underset{\mathcal{G}}{\arg\min}\ \mathbb{E}_{(I^{(j)}, \mathbf{K}^{(j)}, \mathbf{E}^{(j)})\sim\mathcal{D}}[\mathcal{L}_\text{3dgs}(I^{(j)},R(\mathbf{K}^{(j)}, \mathbf{E}^{(j)}, \mathcal{G}))],
\end{align}
where $\mathcal{L}_\text{3dgs}$ consists of L1 loss and a D-SSIM term, $\mathcal{L}_\text{3dgs}(I,\hat{I})=(1-\lambda)\mathcal{L}_1(I,\hat{I})+\lambda\mathcal{L}_\text{D-SSIM}(I,\hat{I})$, where $\lambda$ is a hyperparameter, which is set to 0.2, following \cite{3dgs}.

\subsection{Appearance Modeling with 3DGS}
\label{sec:app}
As in NeRF-W~\cite{nerfw}, we model multiple appearances using a multi-layer perceptron (MLP) with an appearance vector.
Specifically, the coefficients of spherical harmonics are modified by the MLP with the appearance vector as follows:
\begin{align}
    \label{eq:app}
    \hat{k}^{(i)}=k^{(i)}+\phi(\ell,\mathbf{x}^{(i)}),
\end{align}
where $\ell\in\mathbb{R}^d$ is a $d$-dimensional appearance vector, which controls colors of Gaussians, and $\phi:\mathbb{R}^{d+3}\rightarrow\mathbb{R}^l$ denotes an MLP with hash encoding~\cite{mueller2022instant}.
The detailed structure of the MLP is available in the Appendix \ref{sec:detailed-app}.

Each client trains 3DGS with $\phi$ and $\{\ell^{(j)}\}_{j=1}^{N}$ that is a set of appearance vectors for $N$ training images.
Let $\hat{R}(\mathbf{K},\mathbf{E},\mathcal{G},\phi,\ell)$ be a rendering function with the appearance model and vector, which is the same as $R(\mathbf{K},\mathbf{E},\mathcal{G})$ except for modifying $k^{(i)}$ in $\mathcal{G}$ by eq. \eqref{eq:app}.
Then, the objective is as follows:
\begin{align}
    \label{eq:client-obj}
    \underset{\mathcal{G},\phi,\{\ell^{(j)}\}_j}{\arg\min}\ \mathbb{E}_{(I^{(j)}, \mathbf{K}^{(j)}, \mathbf{E}^{(j)})\sim\mathcal{D}_l}[\mathcal{L}_\text{3dgs}(I^{(j)},\hat{R}(\mathbf{K}^{(j)}, \mathbf{E}^{(j)}, \mathcal{G},\phi,\ell^{(j)}))].
\end{align}

\subsection{Global Model Update}
\label{sec:merge}
To update the global model, we merge Gaussians of a local model and those of the global model.
The naive strategy is to append Gaussians of the client model to those of the global model.
However, this strategy monotonically increases the number of Gaussians and the global model size will explode.
Therefore, we require a more efficient way to merge two 3DGS models.

The reasonable approach would be voxel grid filtering used to merge point clouds, which averages parameters of points inside the pre-defined voxel grid.
We can apply it to 3DGS by regarding Gaussians as 3D points based on their 3D position.
However, as demonstrated in the experiments, voxel grid filtering cannot work for 3DGS because each Gaussian has a different role.
For example, some are large and represent the base color of a covered area, while others are much smaller and capture detailed texture.
Since voxel grid filtering averages Gaussians having different roles based on their position, it disrupts both appearance and geometry.

To prevent monotonic growth in the number of Gaussians while preserving the rendered image quality, we propose a distillation-based approach.
In the federated learning scenario, distillation is sometimes used for model update~\cite{lin2020ensemble,zhu2021data}, and it is performed by distilling local models' outputs by the global model.

We first show a simple procedure and explain its problem.
After the central server receives the local model $\mathcal{G}_l$ and a set of cameras $\mathcal{C}_l=\{(\mathbf{K}^{(j)},\mathbf{E}^{(j)})\}_j$, we first merge the local model and the current global model, $\mathcal{G}_g$, as $\hat{\mathcal{G}}_g=\mathcal{G}_g\cup\mathcal{G}_l$.
Then, we optimize opacity logits in $\hat{\mathcal{G}}_g$ and the appearance model $\phi_g$ as follows:
\begin{align}
    \label{eq:simple-dist-obj}
    \underset{\{\hat{o}^{(i)}\}_i,\phi_g,\{\hat{\ell}^{(j)}\}_{j=1}^{|\mathcal{C}_l|}}{\arg\min}\mathbb{E}_{(\mathbf{K}^{(j)}, \mathbf{E}^{(j)})\sim\mathcal{C}_l}[\mathcal{L}_\text{3dgs}(\hat{I}^{(j)}_l, \hat{R}(\mathbf{K}^{(j)}, \mathbf{E}^{(j)}, \hat{\mathcal{G}_g}, \phi_g, \hat{\ell}^{(j)}))],
\end{align}
where $\{\hat{\ell}^{(j)}\}_{j=1}^{|\mathcal{C}_l|}$ is a set of appearance vectors corresponding to cameras in $\mathcal{C}_l$, and each vector is initialized as $\mathbf{0}$; $\hat{I}^{(j)}_l$ is an image rendered by $R(\mathbf{K}^{(j)}, \mathbf{E}^{(j)},\mathcal{G}_l)$.
After optimization, we prune redundant Gaussians based on their opacity and obtain a new global model.
Following \cite{3dgs}, we prune Gaussians that have lower opacity than threshold and set the opacity threshold to 0.05.
Note that since the aim of our distillation is to prune redundant Gaussians from $\hat{\mathcal{G}}_g$ while keeping the rendered image quality, we optimize only the opacity parameter for Gaussians.

\begin{figure}[t]
    \centering
    \includegraphics[clip,width=0.95\hsize]{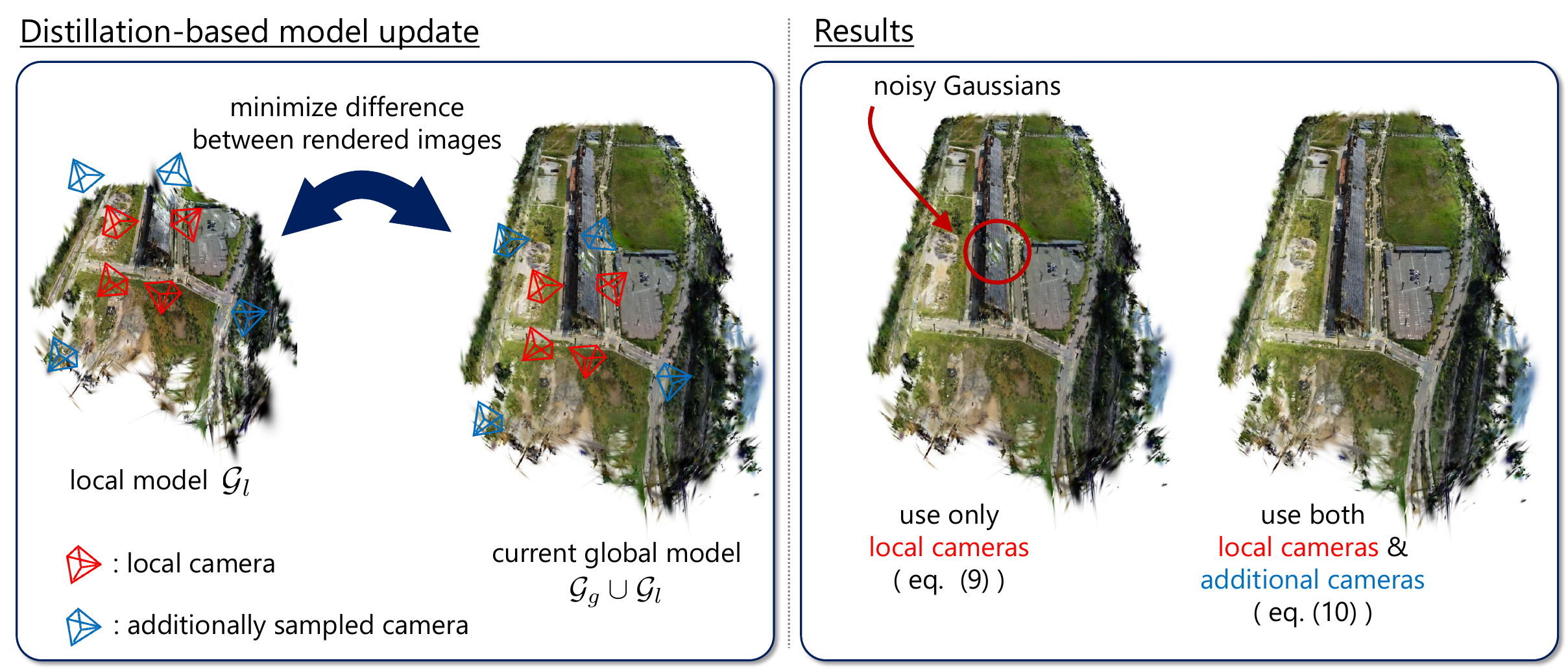}
    \caption{Illustration of the distillation-based model update. The result of distillation only with local cameras (\textit{i.e.}, cameras used to train a local model) has noisy Gaussians.}
    \label{fig:distillation}
\end{figure}
As shown in Fig. \ref{fig:distillation}, Gaussians near the scene boundary are too noisy, and such Gaussians remain after the above distillation procedure because they are not entirely visible from cameras in $\mathcal{C}_l$ and their opacity is rarely updated.

To alleviate this problem, we introduce a global model-based distillation term with additional cameras, $\mathcal{C}_r$, into eq. \eqref{eq:simple-dist-obj}.
We expect Gaussians of $\mathcal{G}_l$ near the scene boundary are visible from $\mathcal{C}_r$ (we describe later how to sample $\mathcal{C}_r$).
Since images rendered through $\mathcal{G}_l$ from cameras in $\mathcal{C}_r$ are not reliable due to the noisy Gaussians, we use $\mathcal{G}_g$ as the teacher model for the view from cameras in $\mathcal{C}_r$.
Then, eq. \eqref{eq:simple-dist-obj} is modified as follows:
\begin{align}
    \nonumber
    \underset{\{\hat{o}^{(i)}\}_i,\phi_g,\{\hat{\ell}^{(j)}\}_{j=1}^{|\mathcal{C}_l|+|\mathcal{C}_r|}}{\arg\min}\mathbb{E}_{(\mathbf{K}^{(j)}, \mathbf{E}^{(j)})\sim\mathcal{C}_l}[\mathcal{L}_\text{3dgs}(\hat{I}^{(j)}_l, \hat{R}(\mathbf{K}^{(j)}, \mathbf{E}^{(j)}, \hat{\mathcal{G}}_g, \phi_g, \hat{\ell}^{(j)}))]\\
    \label{eq:dist-obj}
    +\mathbb{E}_{(\mathbf{K}^{(k)}, \mathbf{E}^{(k)})\sim\mathcal{C}_r}[\mathcal{L}_\text{3dgs}(\hat{I}^{(k)}_g, \hat{R}(\mathbf{K}^{(k)}, \mathbf{E}^{(k)}, \hat{\mathcal{G}}_g, \phi_g, \hat{\ell}^{(|\mathcal{C}_l|+k)}))],
\end{align}
where $\mathbf{K}^{(k)}$ and $\mathbf{E}^{(k)}$ denote the camera intrinsic and extrinsic matrices of the $k$-th camera in $\mathcal{C}_r$, and $\hat{I}^{(k)}_g$ denotes an image rendered by $R(\mathbf{K}^{(k)}, \mathbf{E}^{(k)},\mathcal{G}_g)$.
We also solve this problem using stochastic gradient descent.
The detailed procedure is provided in the Appendix \ref{sec:detailed-pro}.
Since the noisy Gaussians in $\mathcal{G}_l$ increase the second term, we expect that their opacity decreases, and they are pruned.

To sample cameras $\mathcal{C}_r$, we collect camera parameters sent from clients to the central server.
Let $\mathcal{C}_g$ be a set of the collected camera parameters, which is updated after every model update step, as $\mathcal{C}_g\leftarrow\mathcal{C}_g\cup\mathcal{C}_l$.
For sampling $\mathcal{C}_r$, we first filter out cameras from $\mathcal{C}_g$, which have similar extrinsic parameters to cameras in $\mathcal{C}_l$.
We expect this filtering process to remove cameras inside an area modeled by the local model, while the cameras capturing the scene boundary remain.
Then, to selectively sample cameras viewing the local model, we compute sampling probabilities of the remaining cameras as $N^{(j)}/\sum_iN^{(i)}$, where $N^{(j)}$ denotes the number of Gaussians in $\mathcal{G}_l$ visible from $j$-th camera.
Finally, we sample cameras based on the probabilities.
We set the number of sampled cameras to $|\mathcal{C}_l|$ in the experiments.

Although eq. \eqref{eq:dist-obj} can merge two 3DGS without degrading the rendered image quality and reduce several redundant Gaussians, many redundant Gaussians still remain.
This is due to the gradient vanishing problem.
In eq. \eqref{eq:blend}, if the accumulated term, $\prod_{j=1}^{i-1}(1-\alpha_\mathbf{p}^{j})$, is saturated, the gradients with respect to the $i$-th Gaussian will be zero.
Thus, Gaussians located behind those with large opacity such that $\prod_{j=1}^{i-1}(1-\alpha_\mathbf{p}^{j})\approx0$ are not updated in the model update step.

To reduce the redundant Gaussians, we introduce the following techniques:

(i) \textbf{Reset opacity.}
To propagate the gradient to the Gaussians situated behind those with large opacity, we reset the opacity of Gaussians in the local model and the Gaussians of the global model around those of the local model before distillation.
We obtain Gaussians around the local models' Gaussians through range search.
More details are available in the Appendix \ref{sec:detailed-pro}.
The reset opacity is also used to train 3DGS~\cite{3dgs}, and following it, we reset the opacity logit so that its opacity, $\sigma(o)$, is 0.05.

(ii) \textbf{Entropy minimization.}
The reset opacity effectively reduces redundant Gaussians, but they still exist because some Gaussians have small but larger opacity than the pruning threshold after distillation.
To reduce such Gaussians, we enforce opacity to be 0 or 1 by introducing the following entropy minimization term into the distillation objective:
\begin{align}
    \mathcal{L}_\text{entropy}(\mathbf{K},\mathbf{E},\mathcal{G})=\frac{1}{|\mathcal{G}|}\sum_i\mathbb{I}_\text{vis}(\mathbf{x}^{(i)},\mathbf{K},\mathbf{E})H(\sigma(o^{(i)})),
\end{align}
where $\mathbb{I}_\text{vis}(\mathbf{x}^{(i)},\mathbf{K},\mathbf{E})$ is an indicator function, which is 1 if 2D projection of $\mathbf{x}^{(i)}$ is on the image plane of the camera $(\mathbf{K},\mathbf{E})$ and $\prod_{j=1}^{i-1}(1-\alpha_\mathbf{p}^{j})$ is not zero, and 0 otherwise;
$H(x)$ denotes the entropy, $H(x)=-x\log(x)-(1-x)\log(1-x)$.
Then, the objective function, $\mathcal{L}_\text{3dgs}$, is modified as $\mathcal{L}_\text{3dgs}+\eta\mathcal{L}_\text{entropy}$, where $\eta$ is a hyperparameter, which is set to 0.01 in the experiments.

The distillation-based model update reconstructs local data through local models and optimizes the model on the central server; however, it offers several advantages, compared to the centralized approaches.
The size of local models is smaller than the local data, alleviating bandwidth consumption (we show the model size and local data size in the Appendix \ref{sec:add-ablation}).
Furthermore, recent exploration into the compression of 3DGS~\cite{fan2023lightgaussian,lee2023compact} suggests additional opportunities to further reduce the local model size.
In terms of computational costs, each model update step takes only a few minutes, which is significantly shorter than training from scratch, and the peak load on the central server for our method is lower than that of centralized approaches.

\section{Experiments}
\subsection{Evaluation Protocols}
\textbf{Datasets.}
Following \cite{Turki_2022_CVPR}, we use six scenes for evaluation: Mill 19 Building and Rubble scenes~\cite{Turki_2022_CVPR}, Quad 6k~\cite{quad6k}, and UrbanScene3D Residence, Sci-Art, and Campus scenes~\cite{UrbanScene3D}.
For Quad 6k, we use a pretrained segmentation model~\cite{suzuki2022clustering} to remove common movable objects, as in \cite{Turki_2022_CVPR}.

\textbf{Federated learning setup.}
We simulate our method in the federated learning scenario using existing datasets.
Basically, we follow the FedNeRF setup~\cite{fednerf}; namely, we generate sets of local data by randomly sampling 100 to 200 images per client from the original data.
We describe the detailed sampling procedure in the Appendix \ref{sec:detailed-exp}.
We generate 400 sets (\textit{i.e.}, 400 clients) for Quad 6k and the UrbanScene3D Campus scene and 200 sets for the other scenes because Quad 6k and Campus have about twice as many images as the other scenes.
The local models are trained using the local data, and then we build the global model from the local models by repeating the model update step, eq. \eqref{eq:dist-obj}, for a number of clients.
Hyperparameters, such as learning rate, are described in the Appendix \ref{sec:detailed-exp}.
As global model initialization, we randomly select one local model as the global model.
As client selection in every model update step, we choose a client whose data $\mathcal{C}_l$ (partially) overlaps the global model's data $\mathcal{C}_g$.
More detailed protocols are described in the Appendix \ref{sec:detailed-exp}.
Note that since our main focus in this work is the scalability and appearance modeling, we assume that the global pose of clients obtained from their sensors is accurate in the experiments.
We evaluate global pose alignment using the FedNeRF's procedure~\cite{fednerf} in the Appendix \ref{sec:add-ablation}.

\textbf{Metrics.}
We report PSNR, SSIM~\cite{ssim}, and the VGG~\cite{simonyan2014very} implementation of LPIPS~\cite{lpips} for comparison.

\subsection{Main Results}
We compare our method with centralized approaches~\cite{nerf,mi2023switchnerf,zhang2023efficient}, including distributed training~\cite{Turki_2022_CVPR,jia2024drone}, and a federated learning baseline~\cite{fednerf}.
We show the results on the benchmarks in Tab. \ref{tab:main-res}.

\begin{figure}
    \centering
    \begin{tabular}{cccc}
        \includegraphics[clip,width=0.24\hsize]{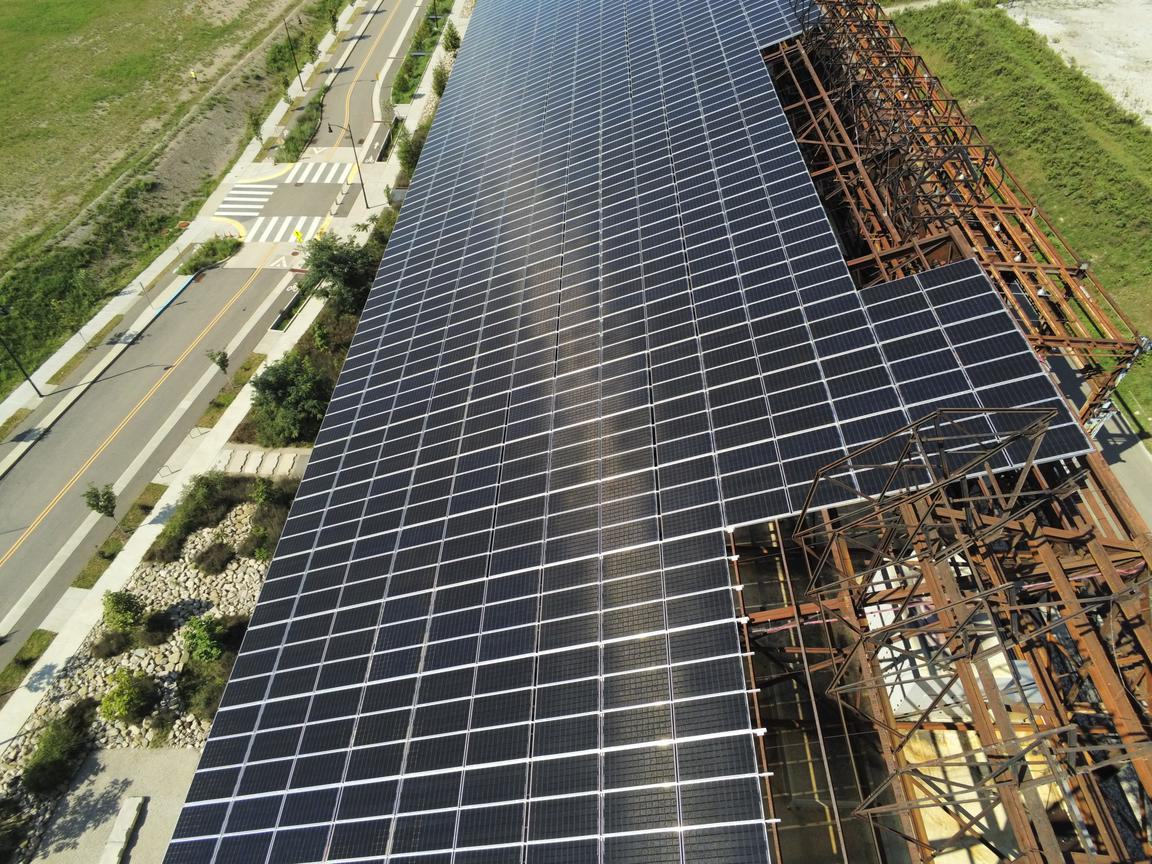} & 
        \includegraphics[clip,width=0.24\hsize]{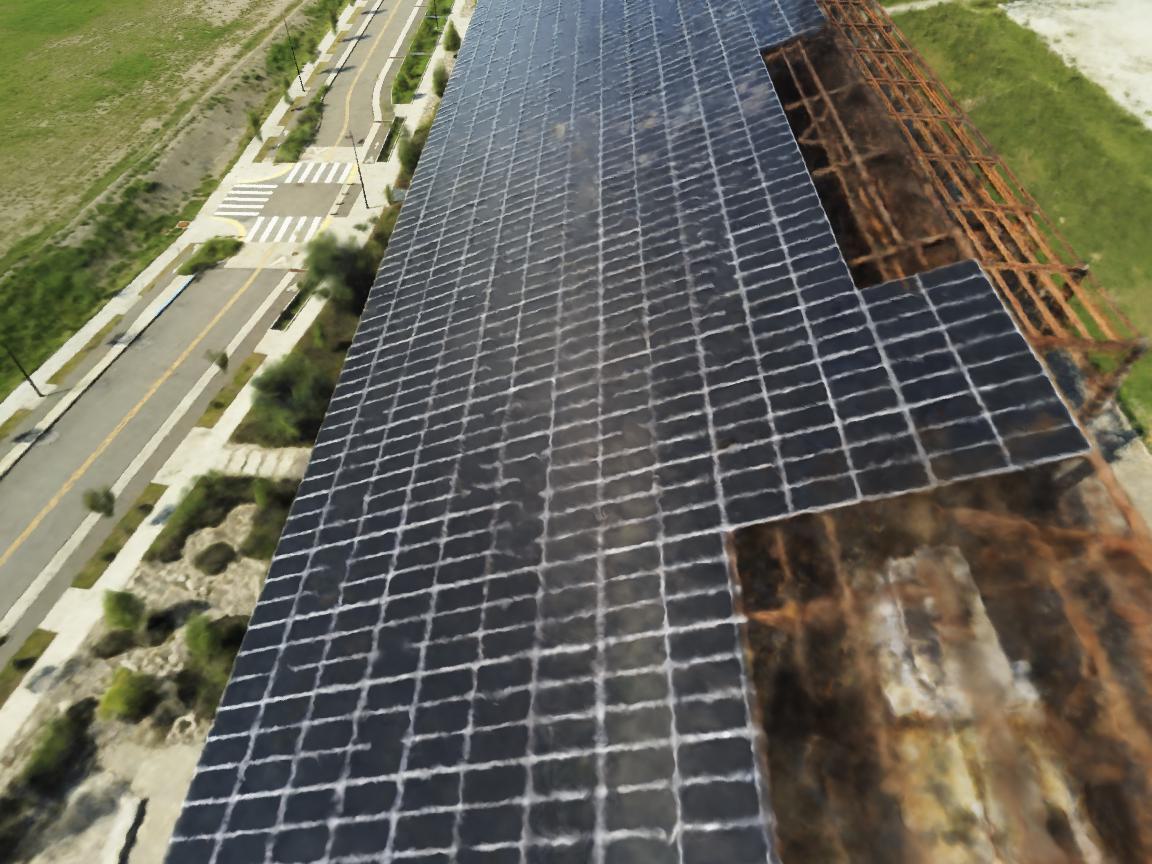} & 
        \includegraphics[clip,width=0.24\hsize]{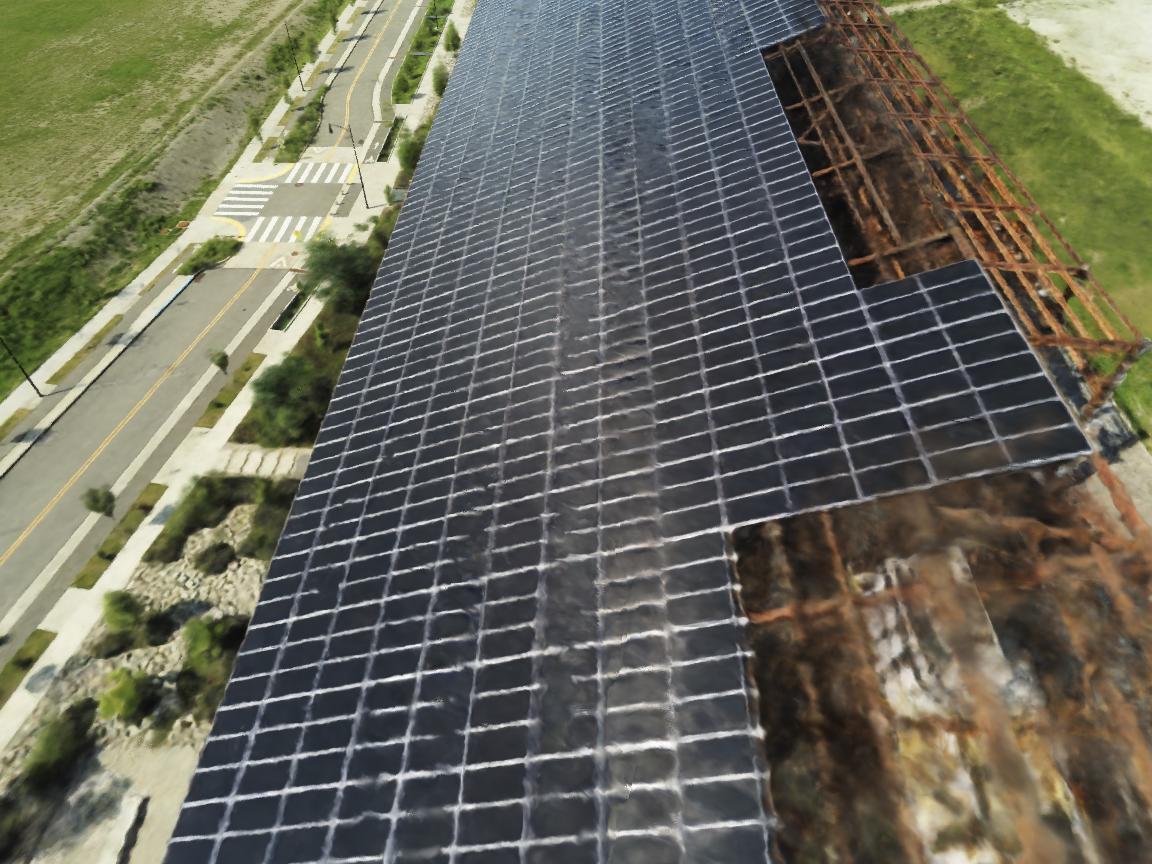} & 
        \includegraphics[clip,width=0.24\hsize]{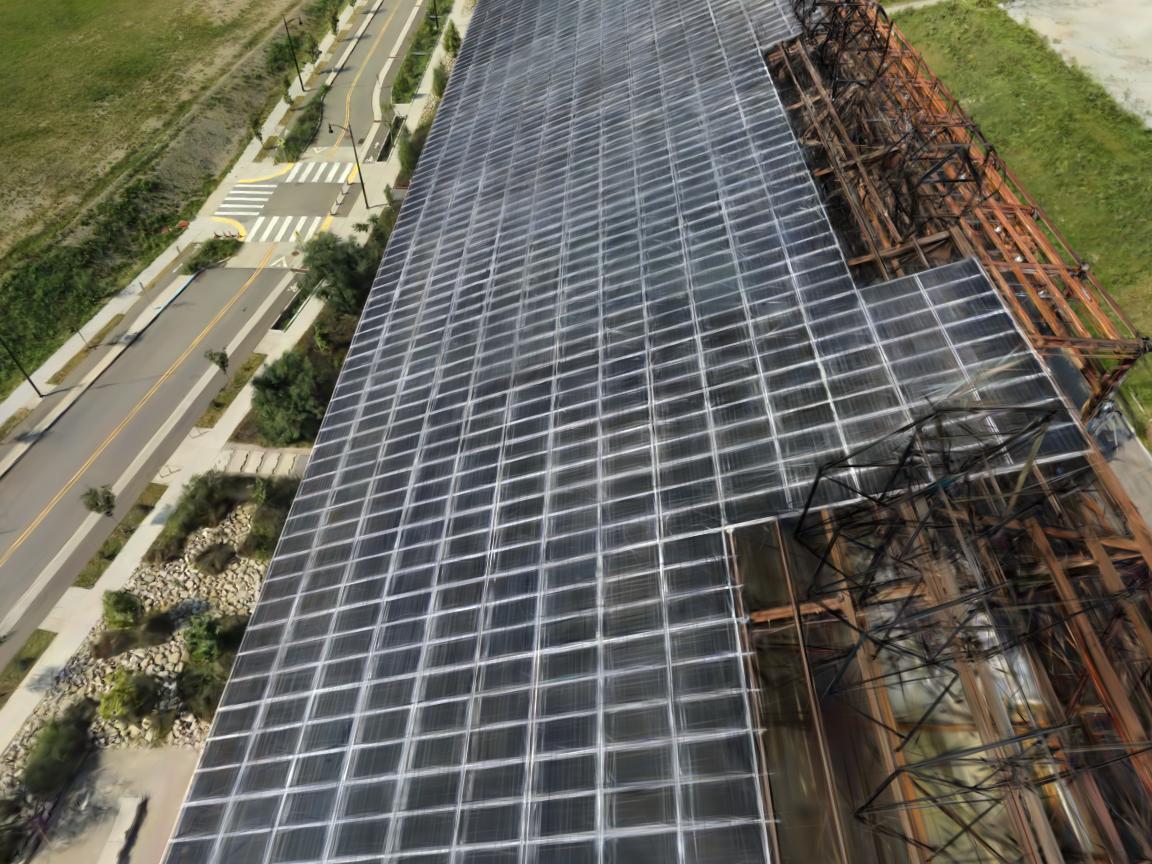} \\
        & 
        \includegraphics[clip,width=0.24\hsize]{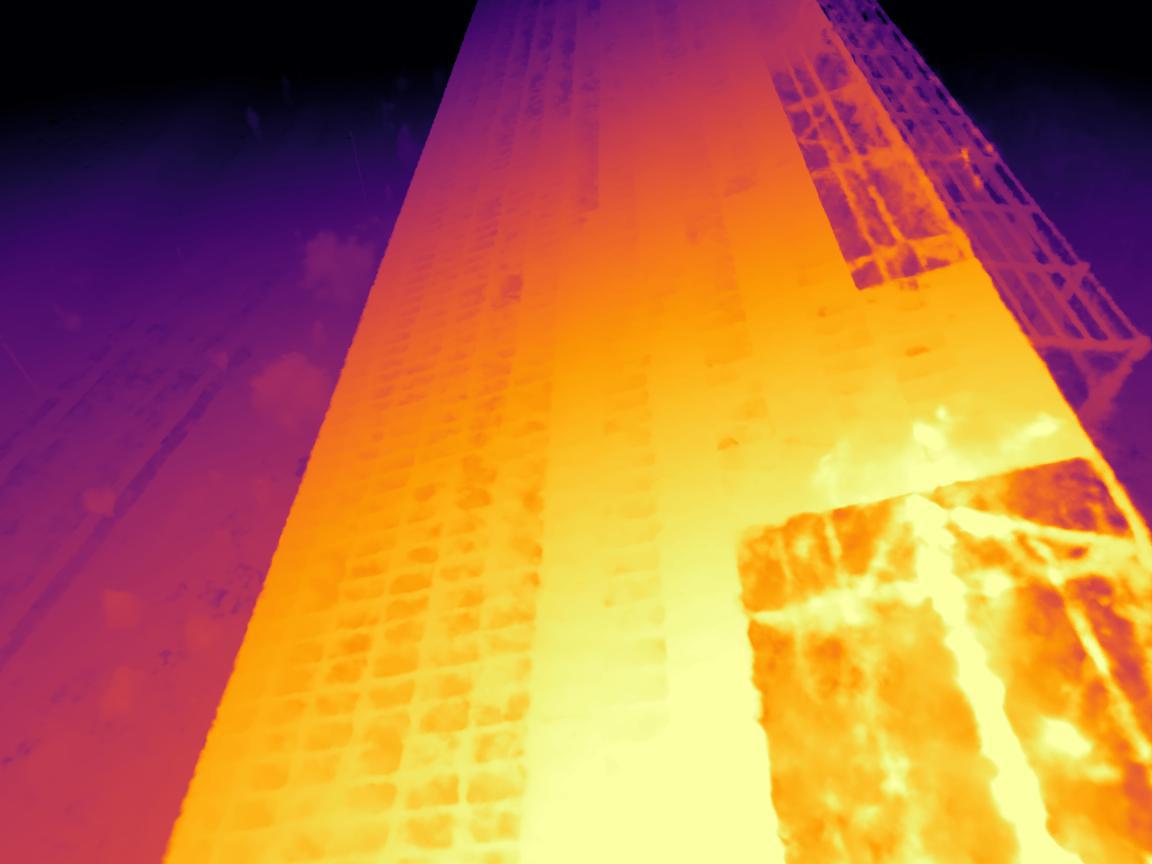} & 
        \includegraphics[clip,width=0.24\hsize]{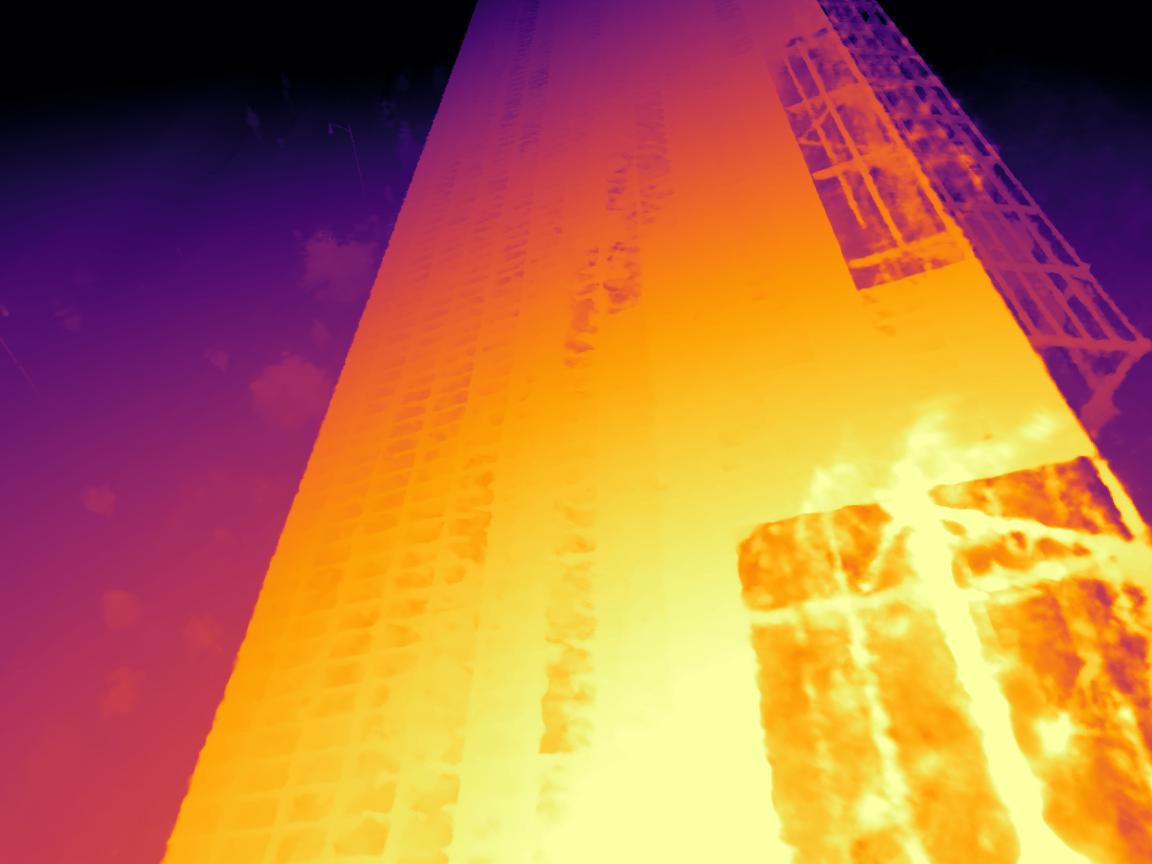} & 
        \includegraphics[clip,width=0.24\hsize]{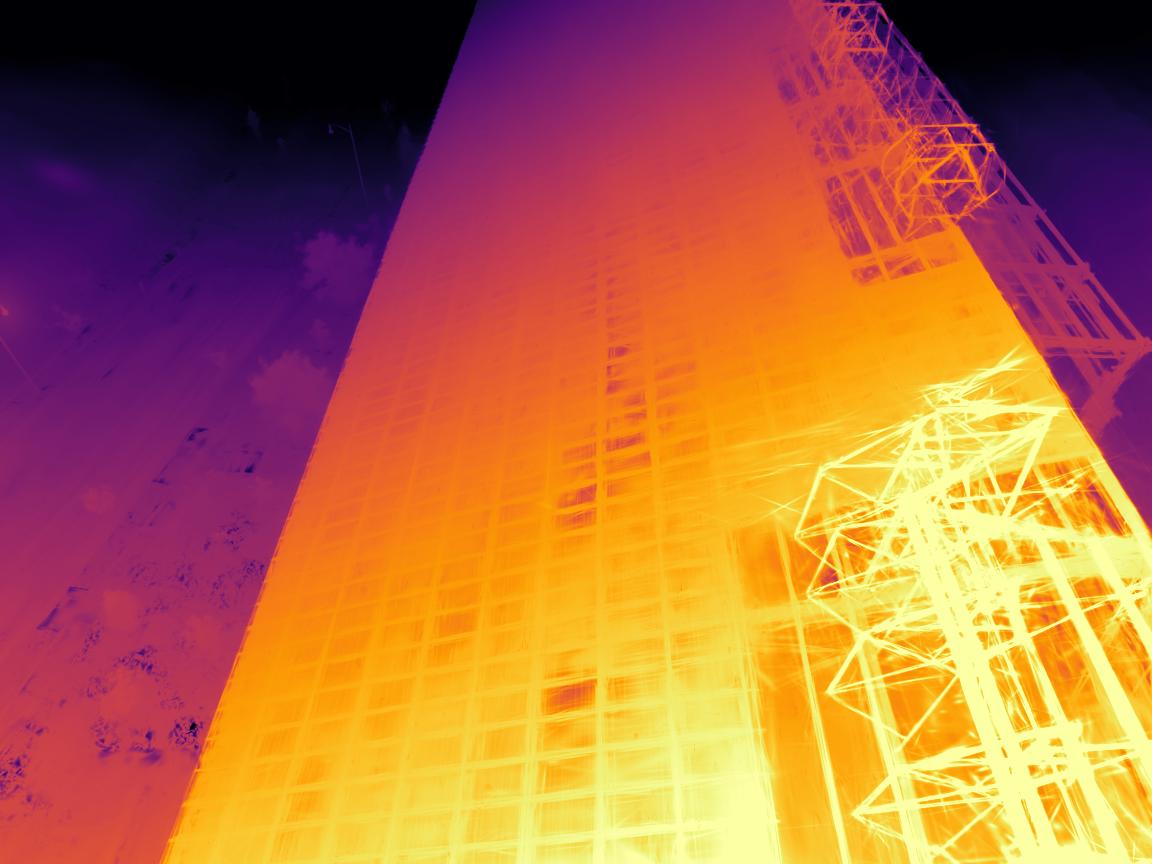} \\
        \includegraphics[clip,width=0.24\hsize]{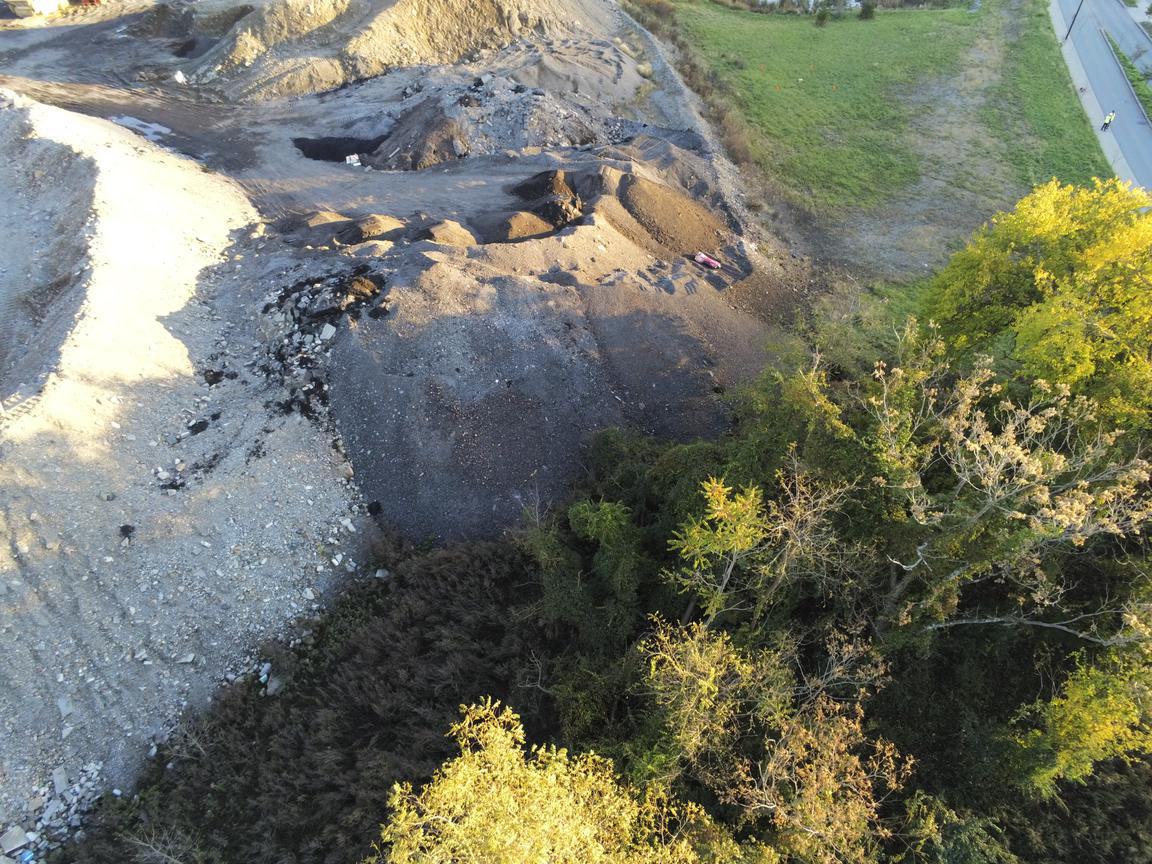} & 
        \includegraphics[clip,width=0.24\hsize]{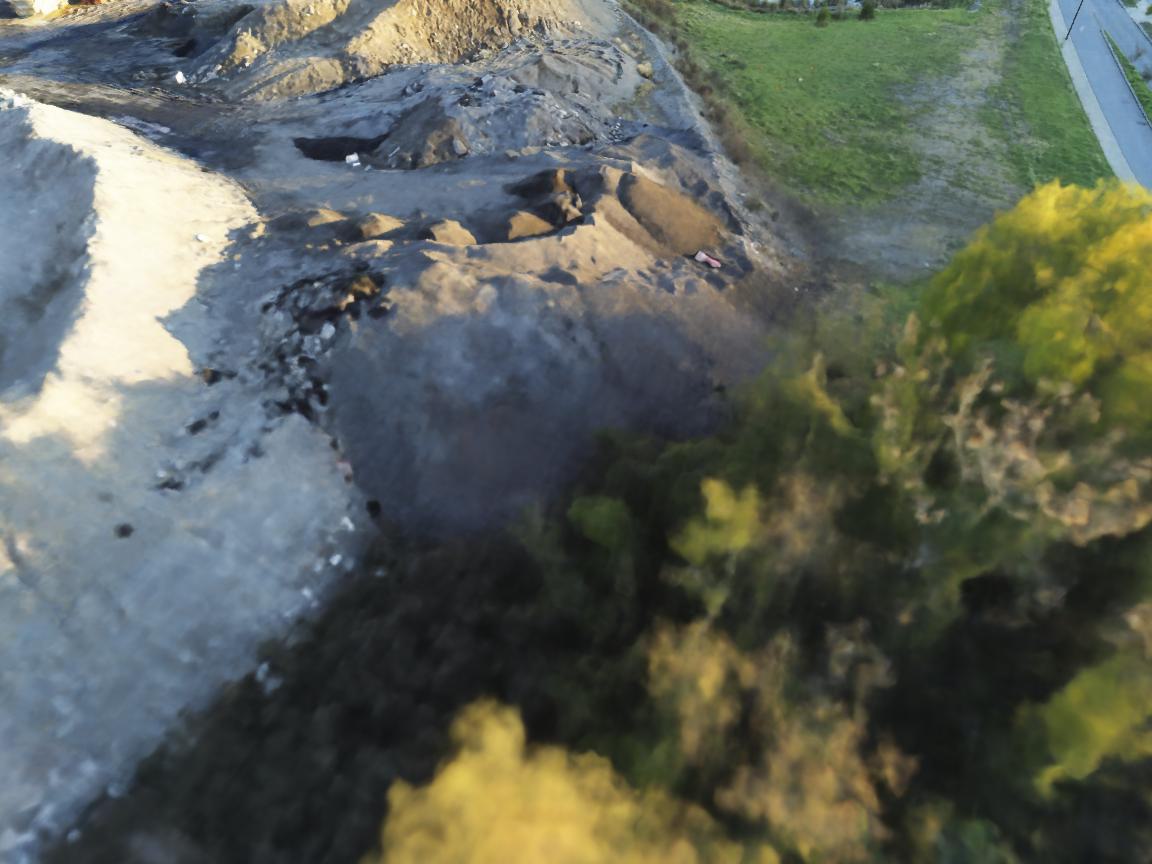} & 
        \includegraphics[clip,width=0.24\hsize]{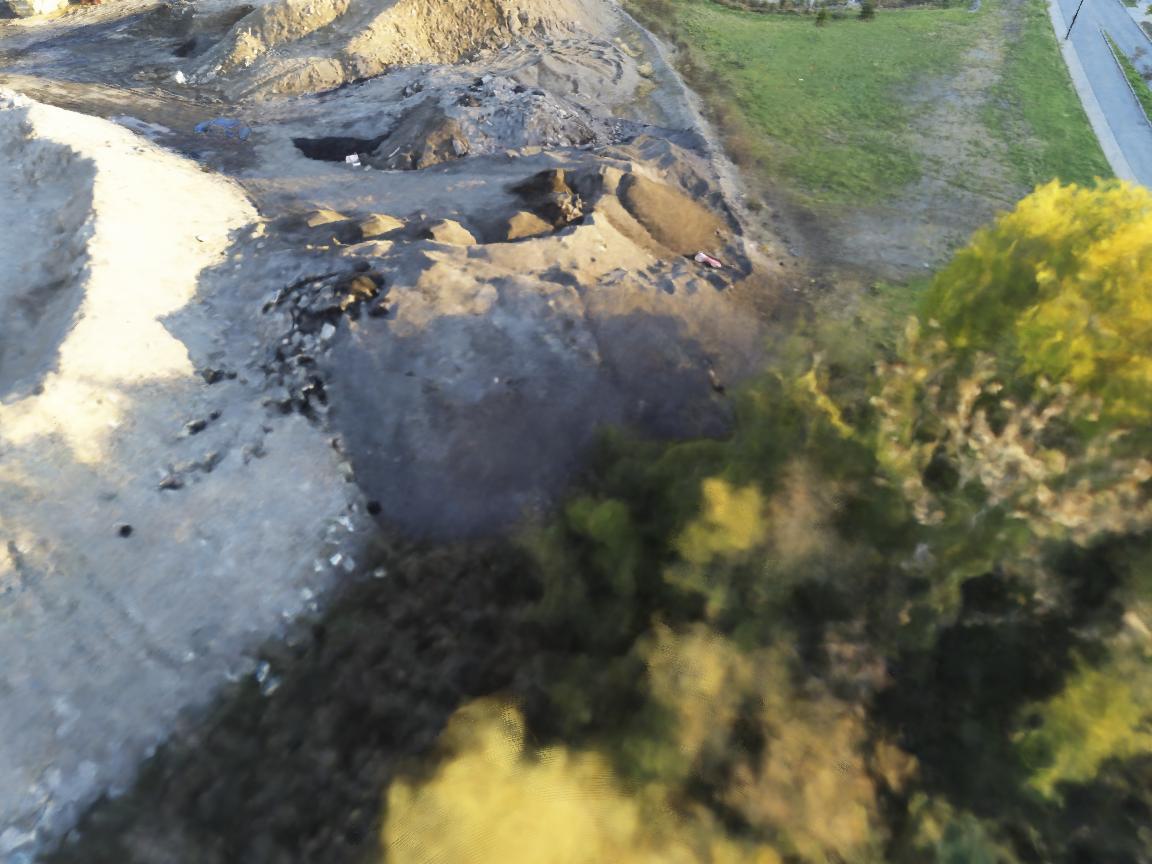} & 
        \includegraphics[clip,width=0.24\hsize]{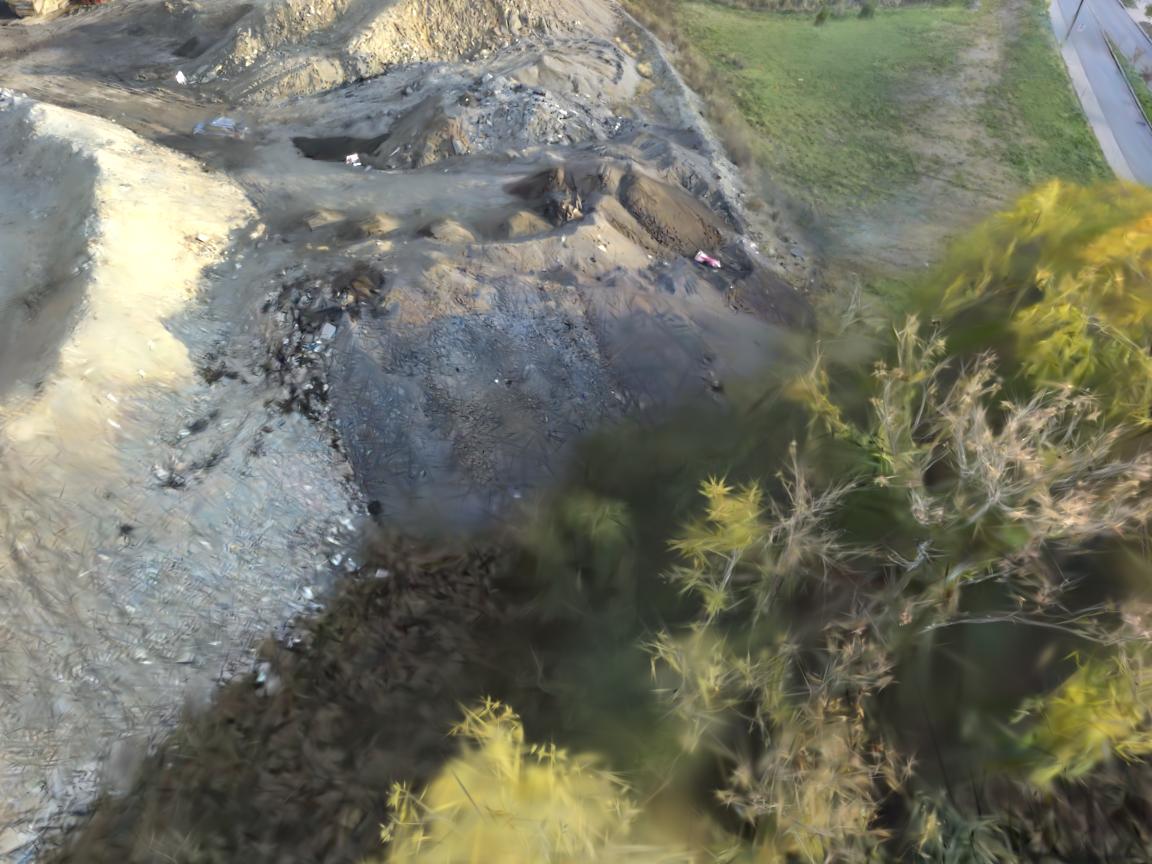} \\
        & 
        \includegraphics[clip,width=0.24\hsize]{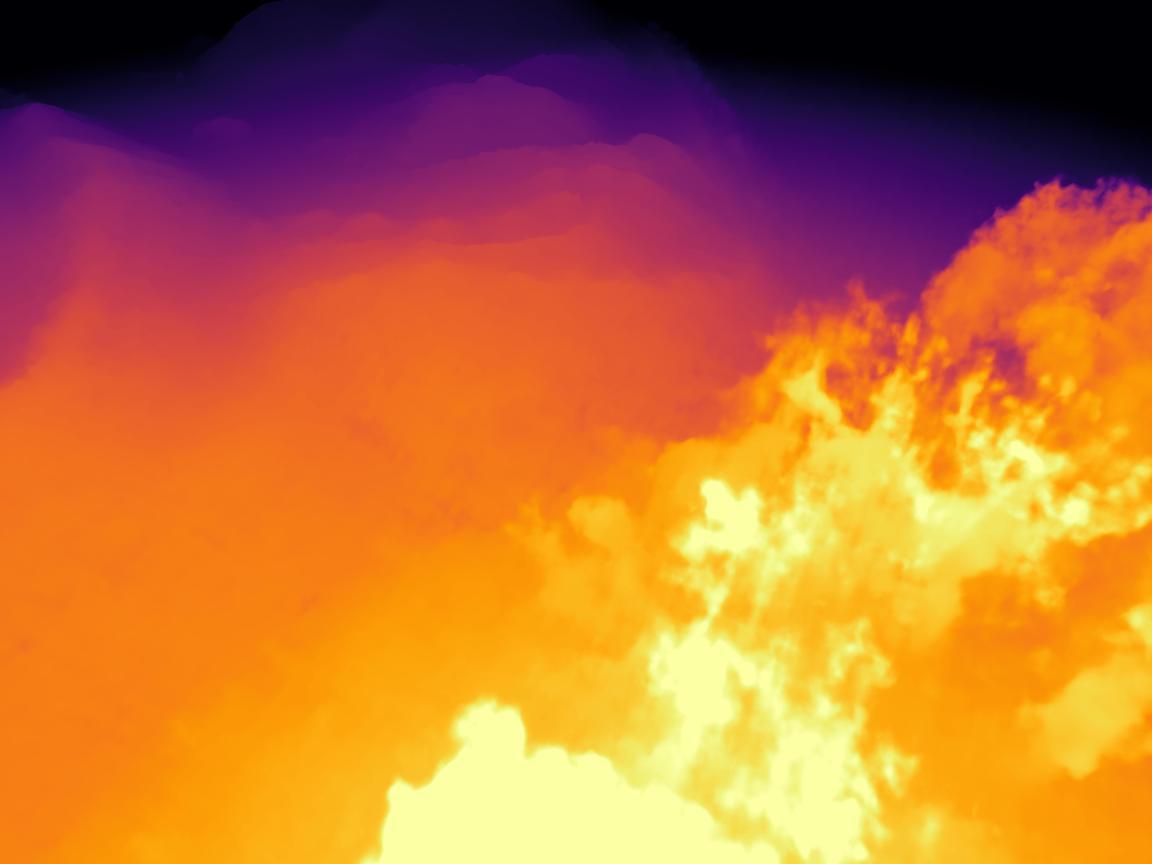} & 
        \includegraphics[clip,width=0.24\hsize]{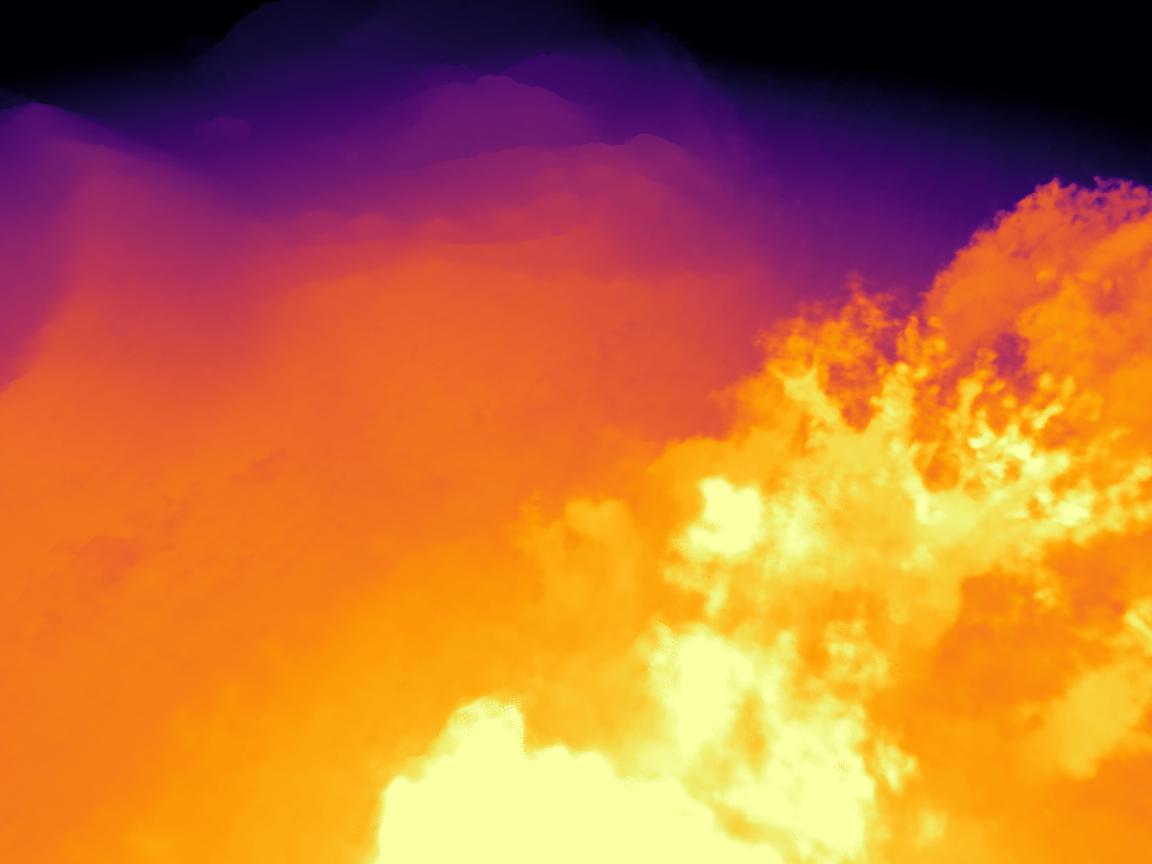} & 
        \includegraphics[clip,width=0.24\hsize]{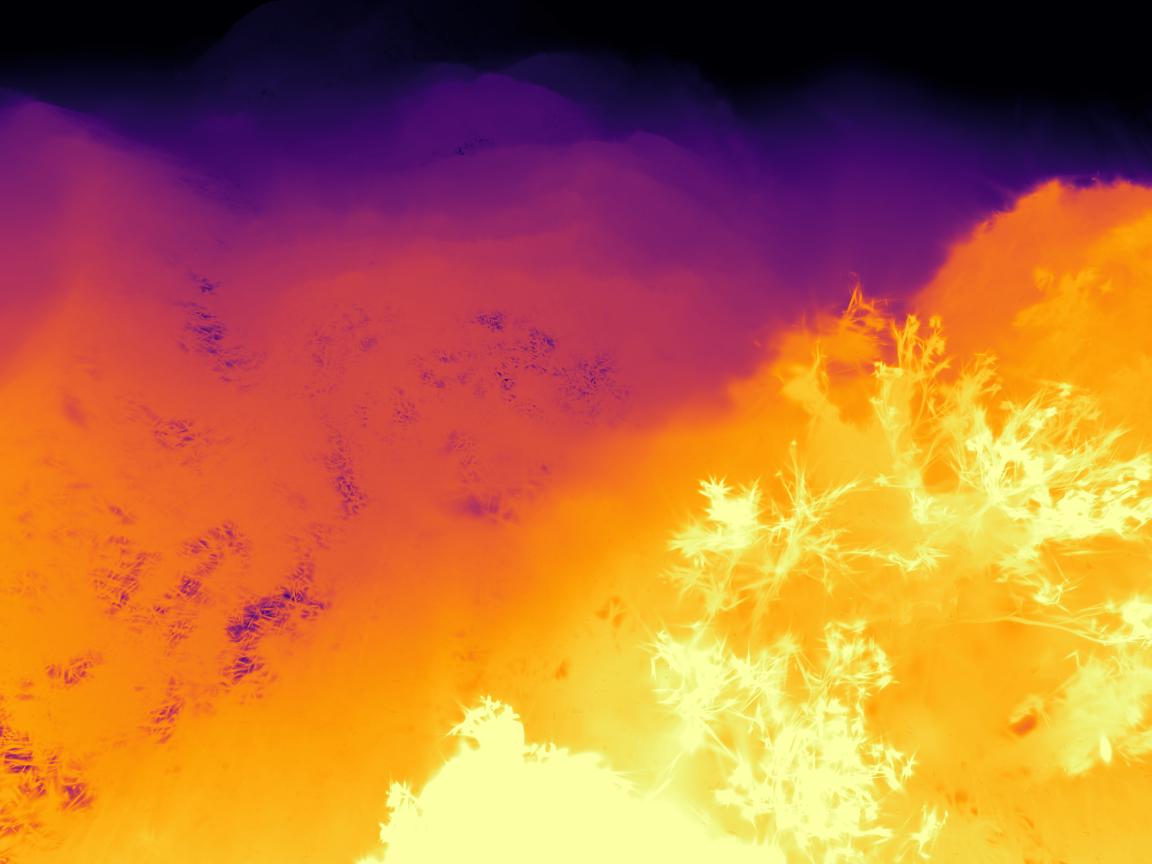} \\
        \includegraphics[clip,width=0.24\hsize]{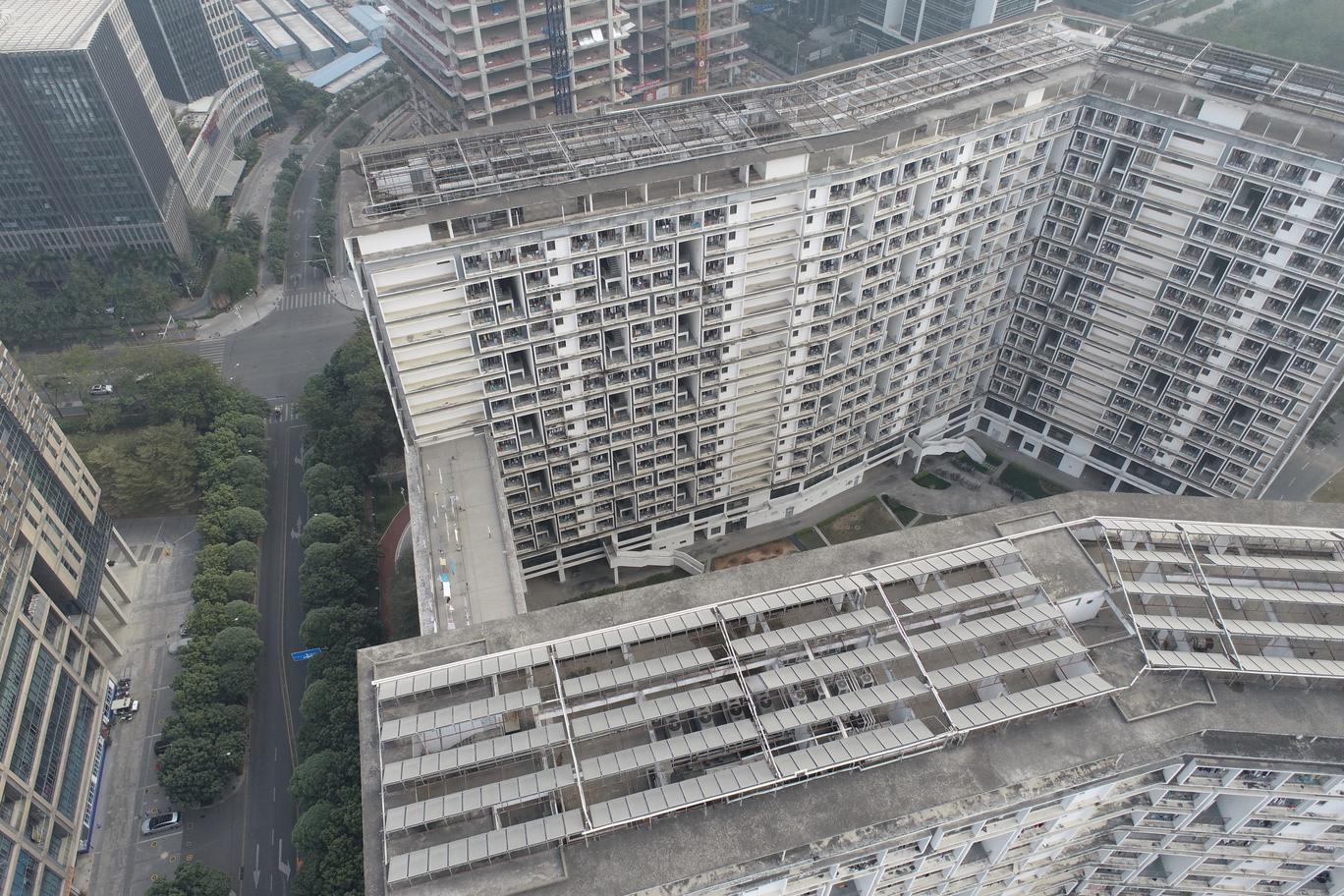} & 
        \includegraphics[clip,width=0.24\hsize]{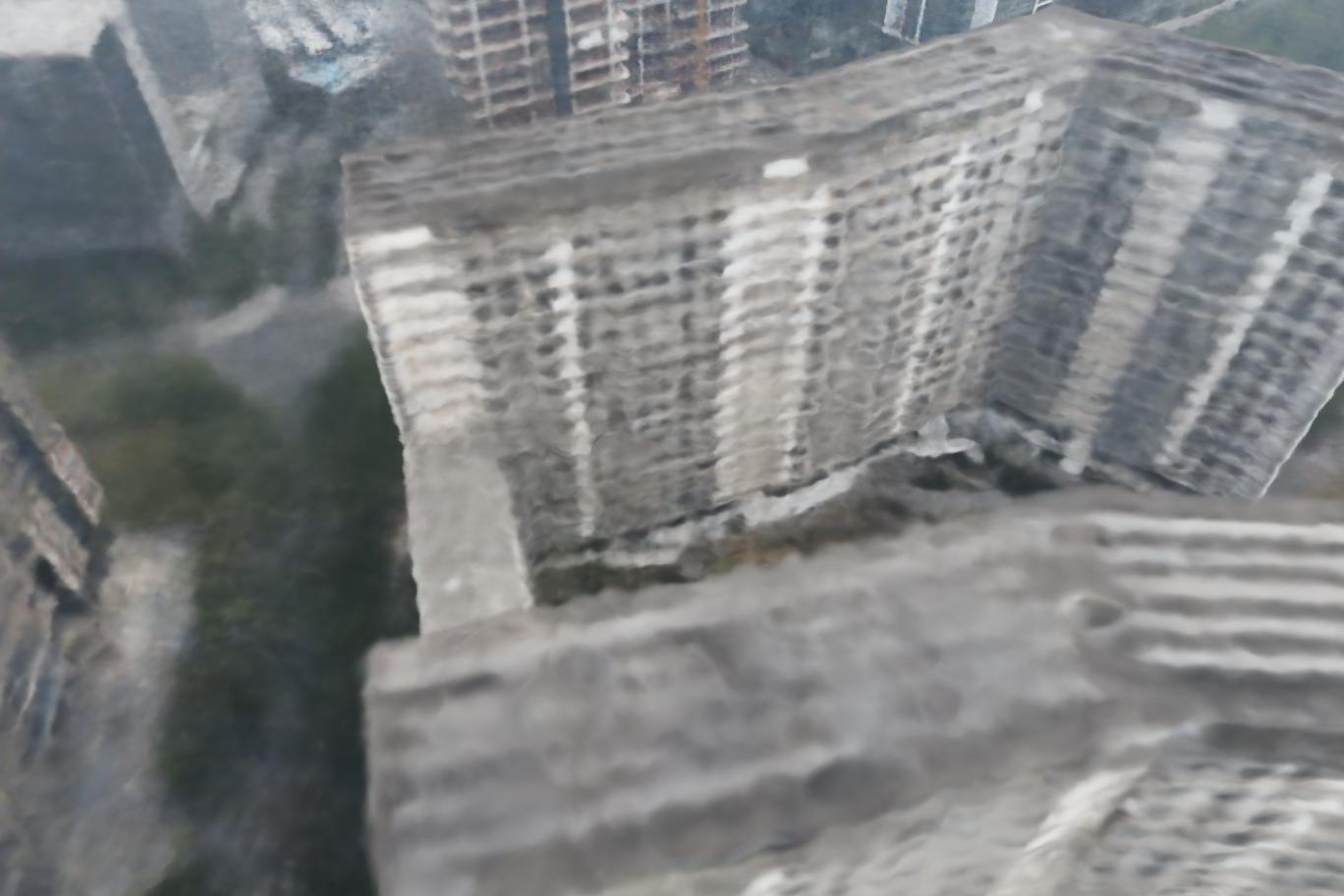} & 
        \includegraphics[clip,width=0.24\hsize]{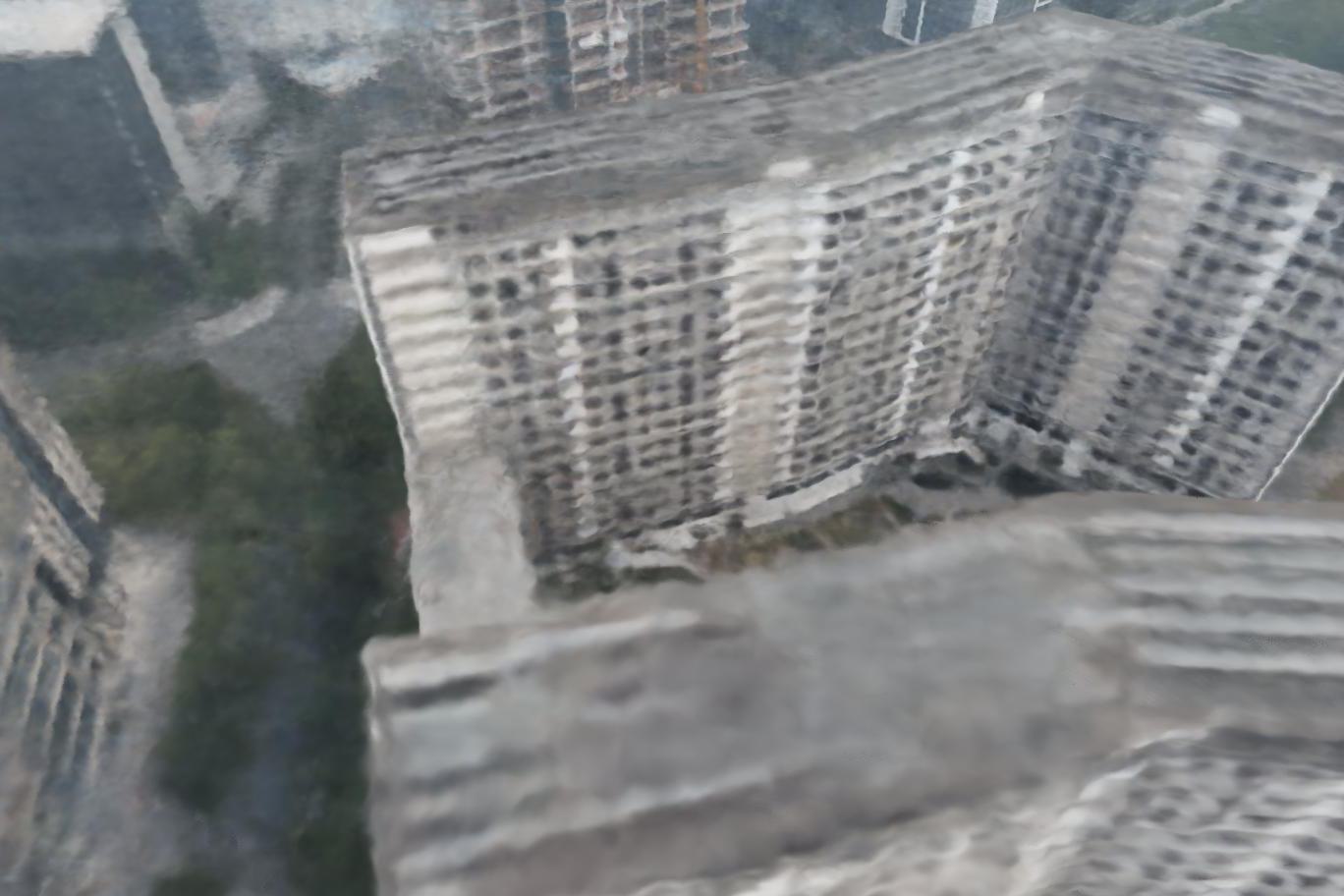} & 
        \includegraphics[clip,width=0.24\hsize]{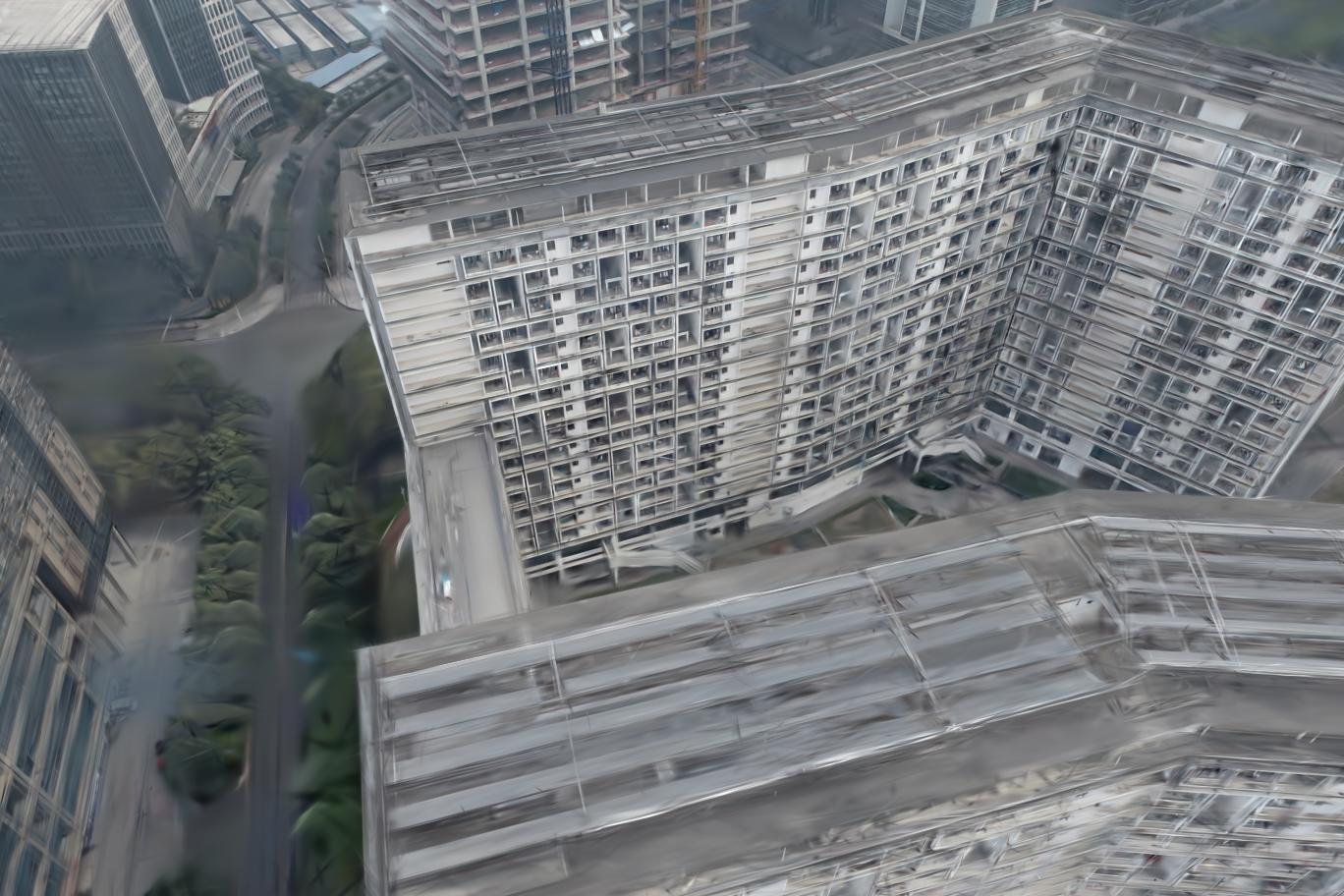} \\
        & 
        \includegraphics[clip,width=0.24\hsize]{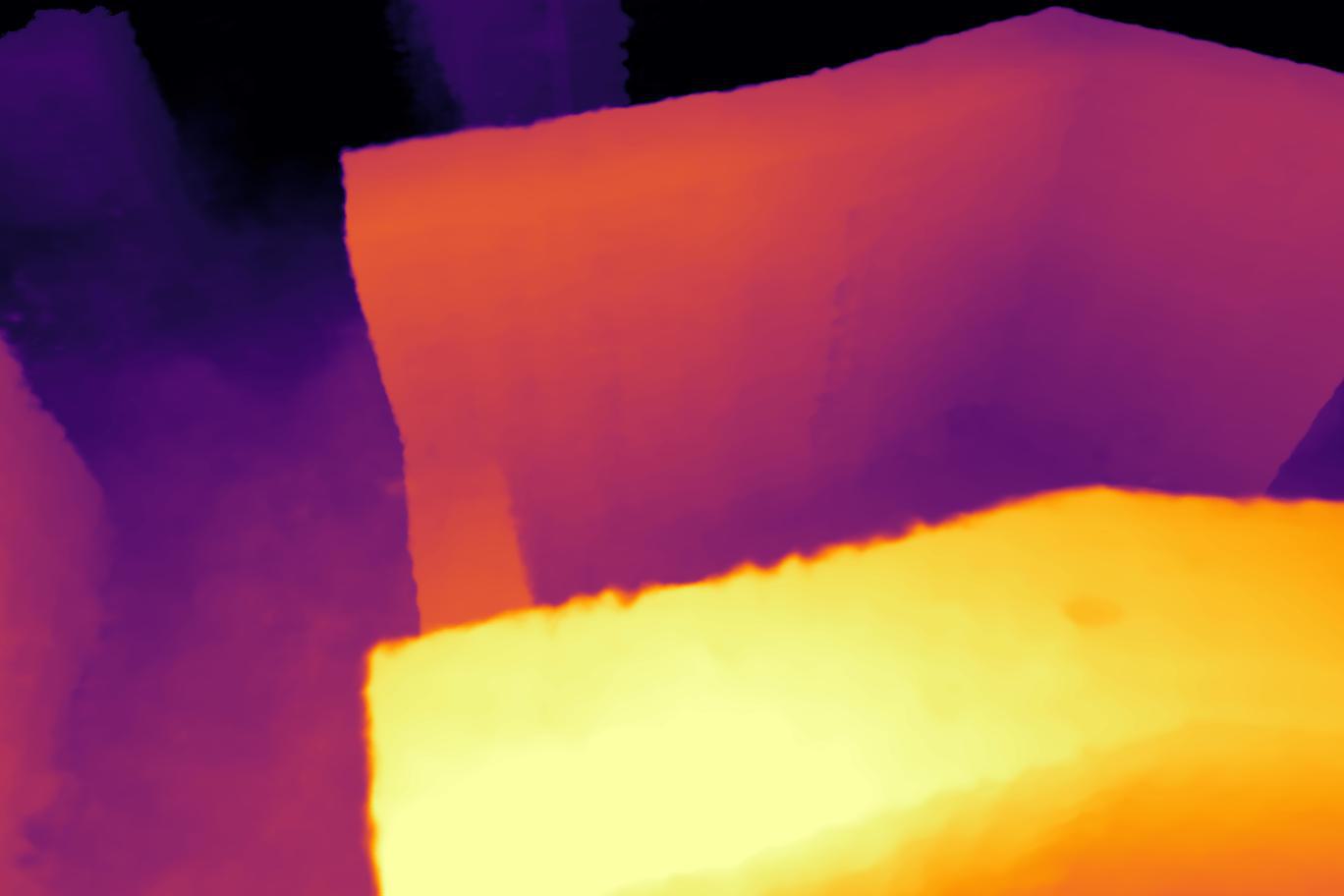} & 
        \includegraphics[clip,width=0.24\hsize]{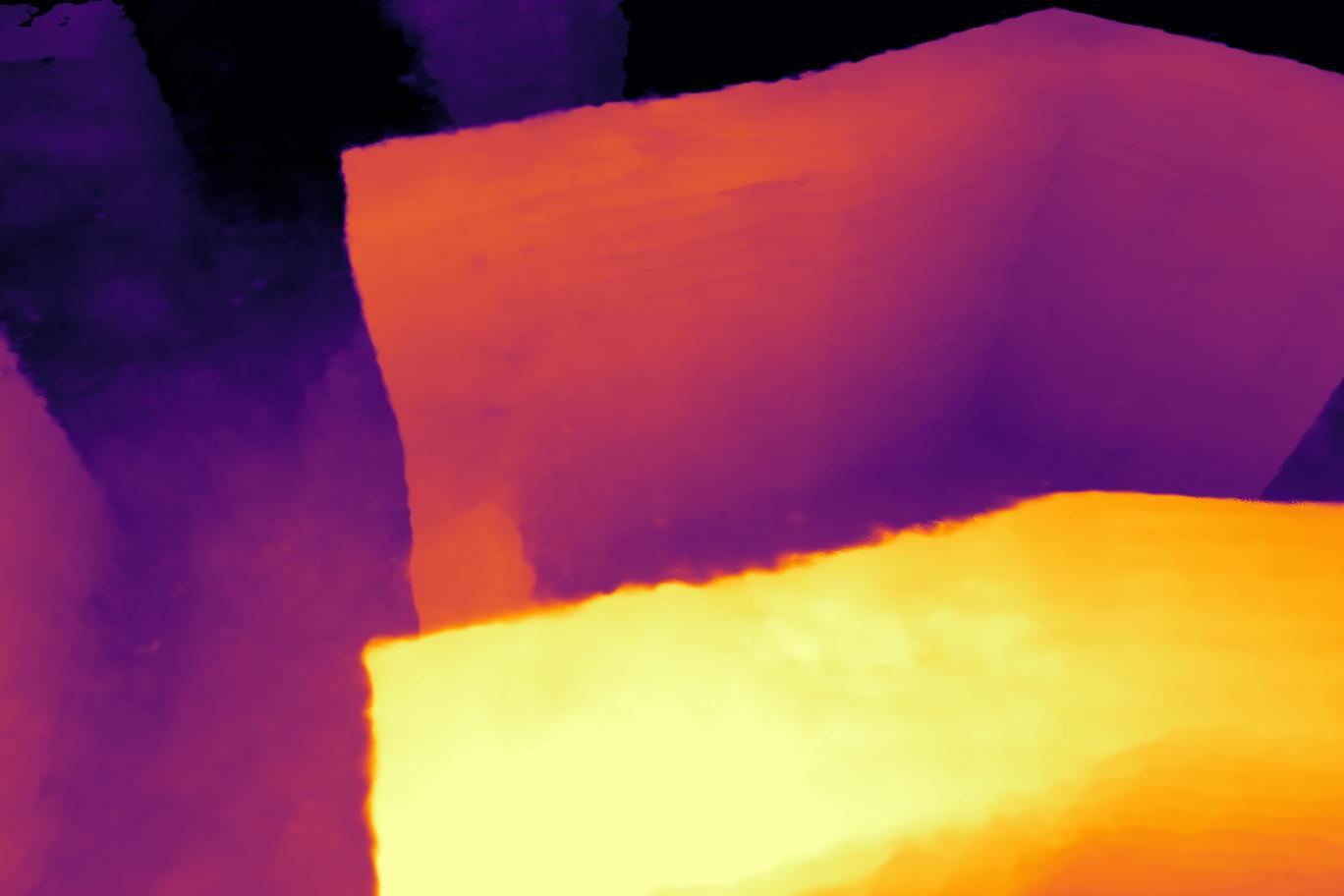} & 
        \includegraphics[clip,width=0.24\hsize]{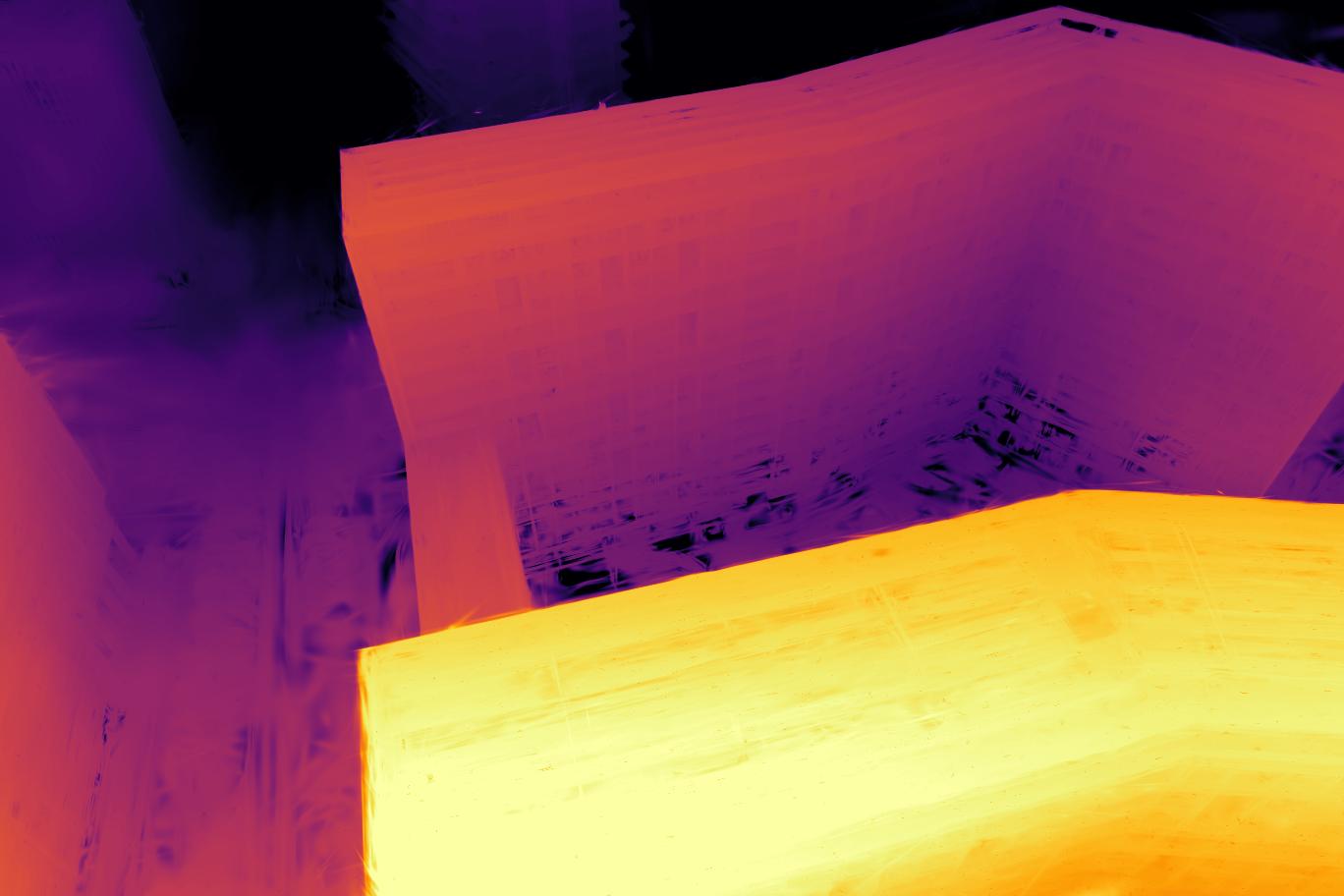} \\
        \includegraphics[clip,width=0.24\hsize]{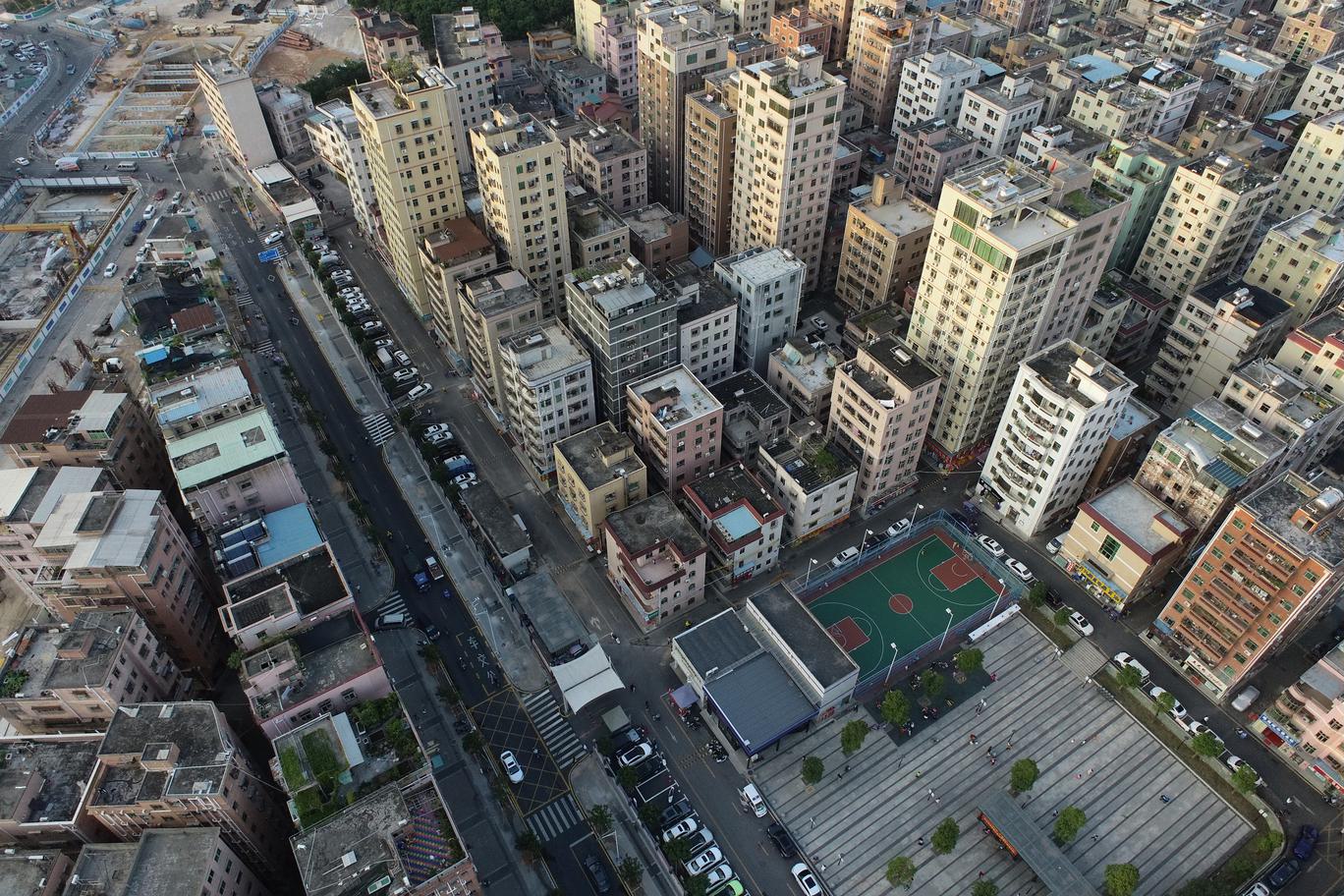} & 
        \includegraphics[clip,width=0.24\hsize]{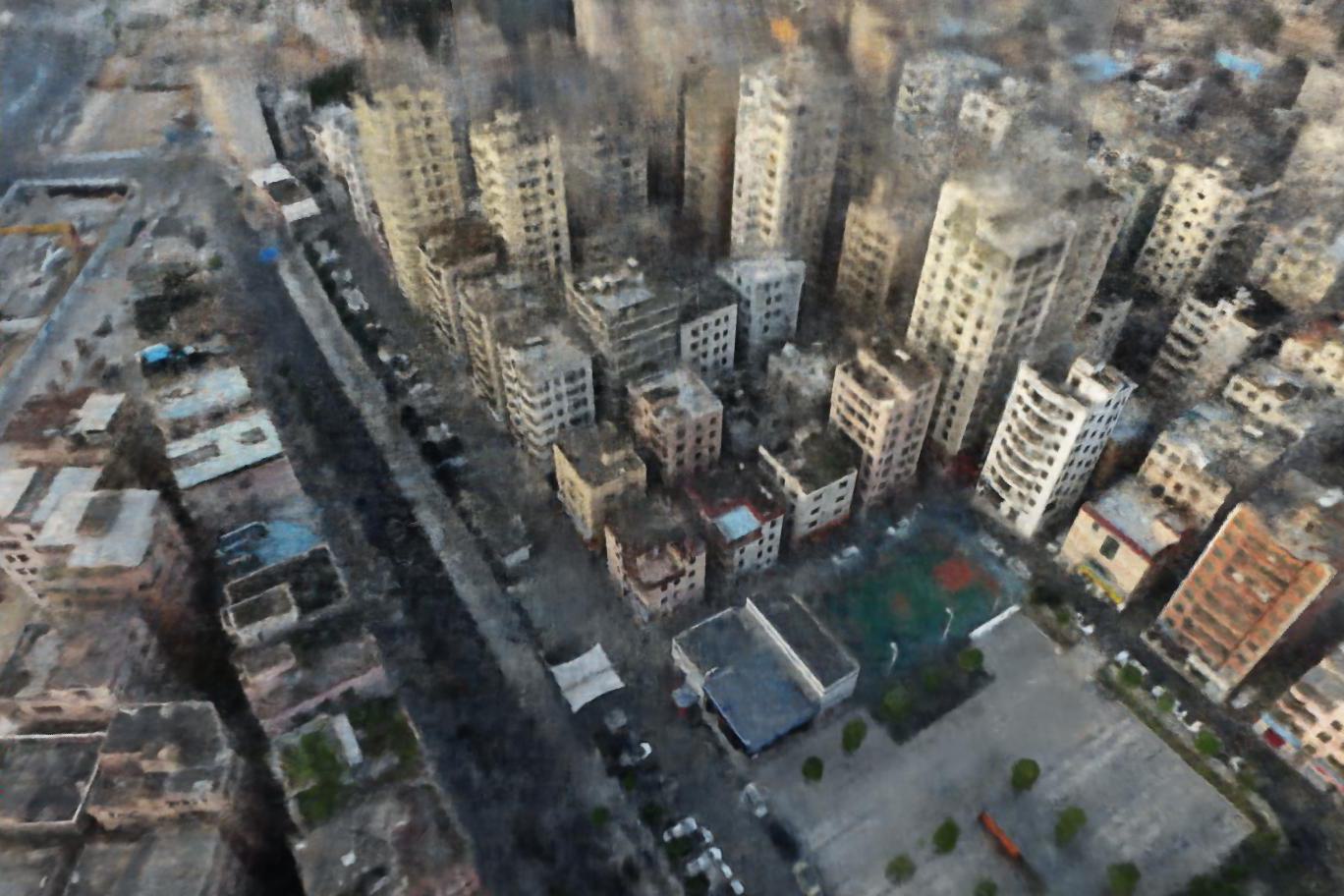} & 
        \includegraphics[clip,width=0.24\hsize]{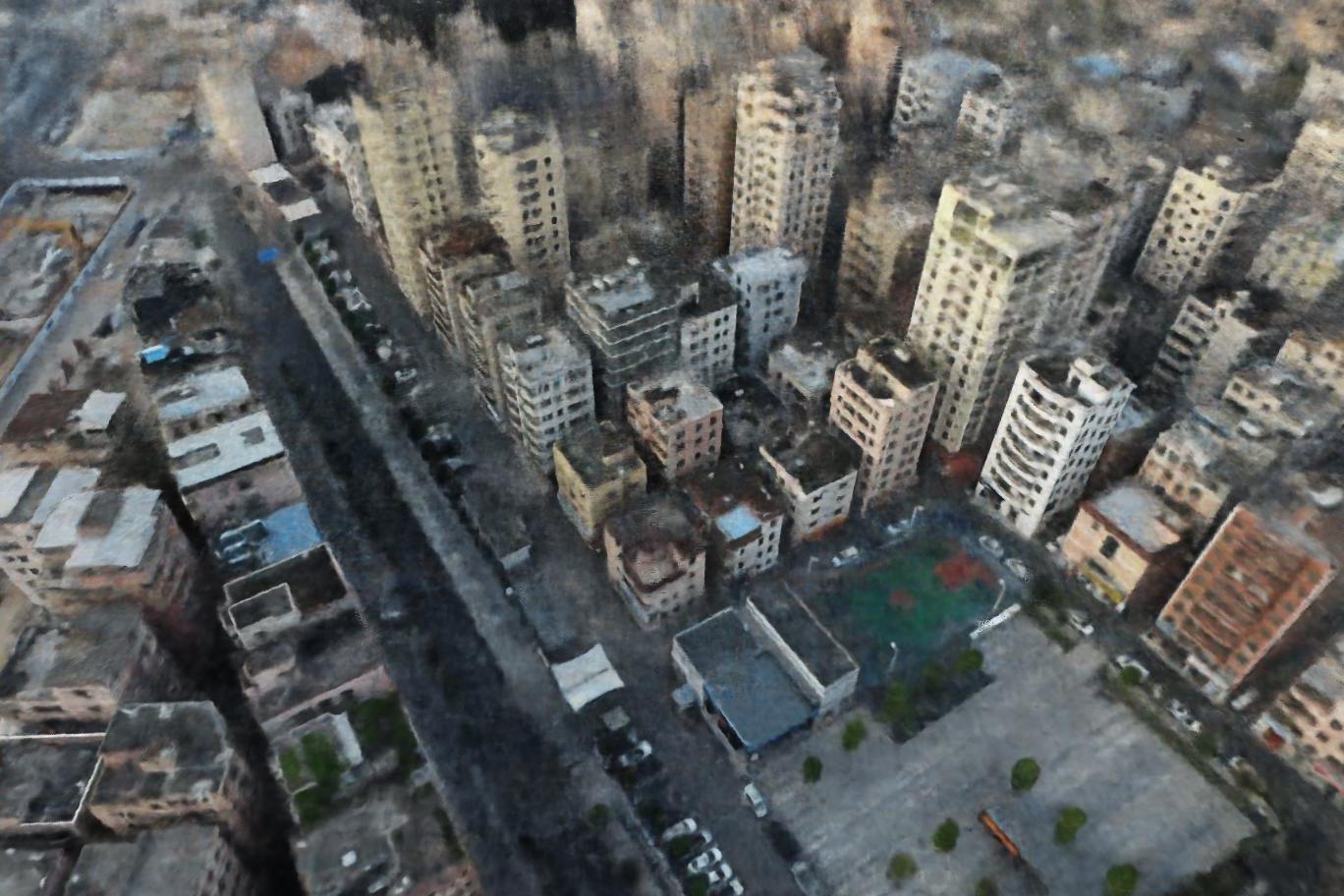} & 
        \includegraphics[clip,width=0.24\hsize]{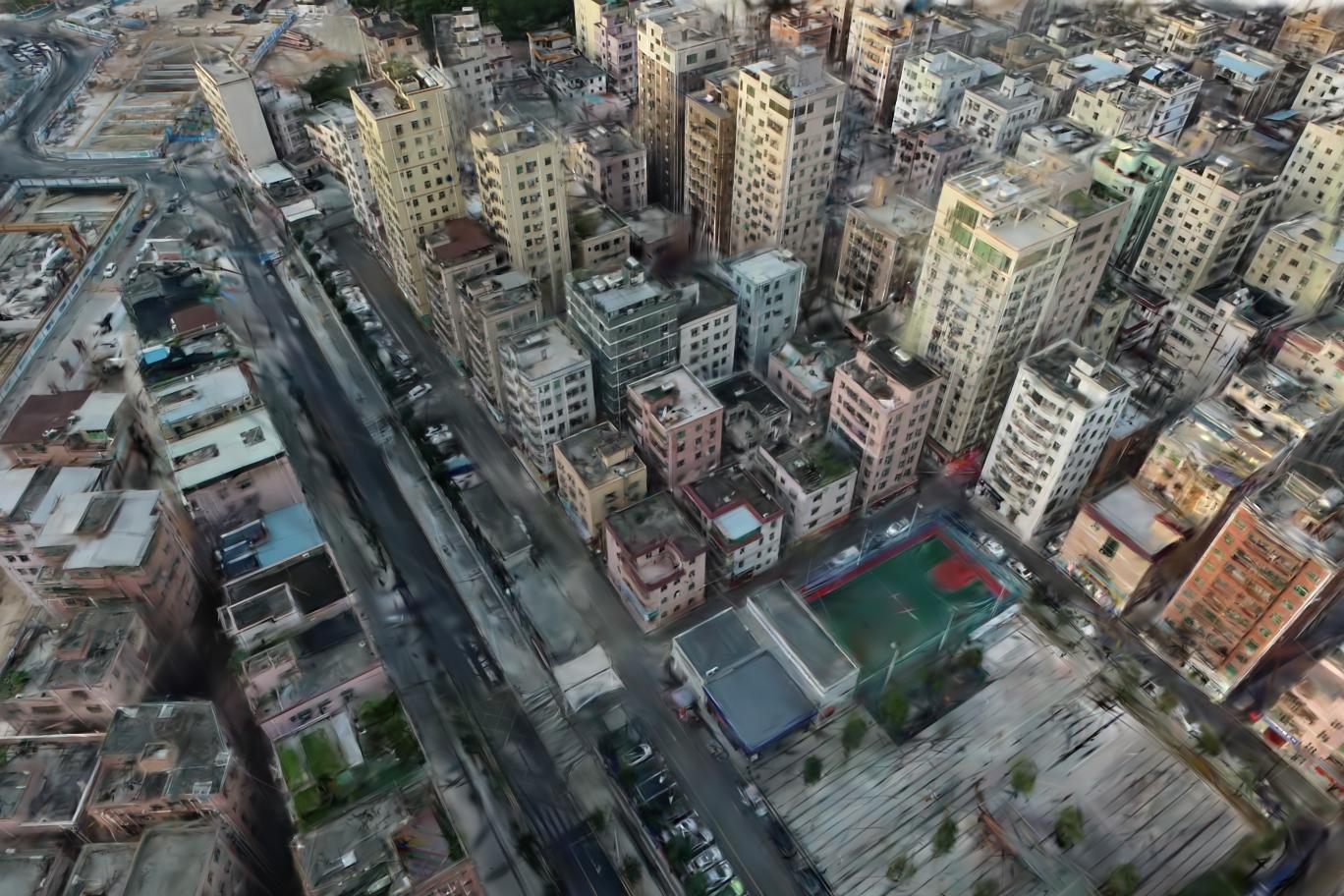} \\
        & 
        \includegraphics[clip,width=0.24\hsize]{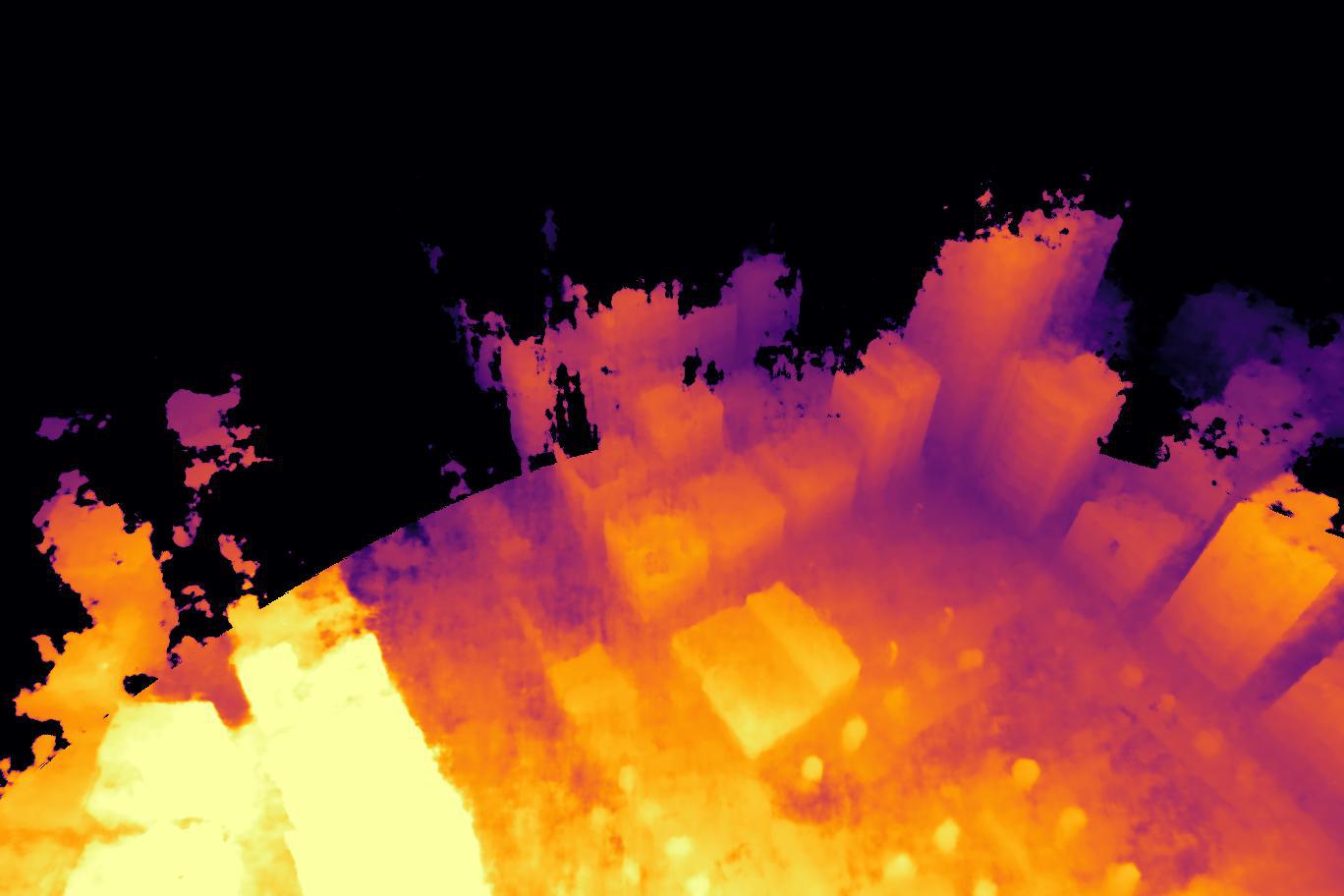} & 
        \includegraphics[clip,width=0.24\hsize]{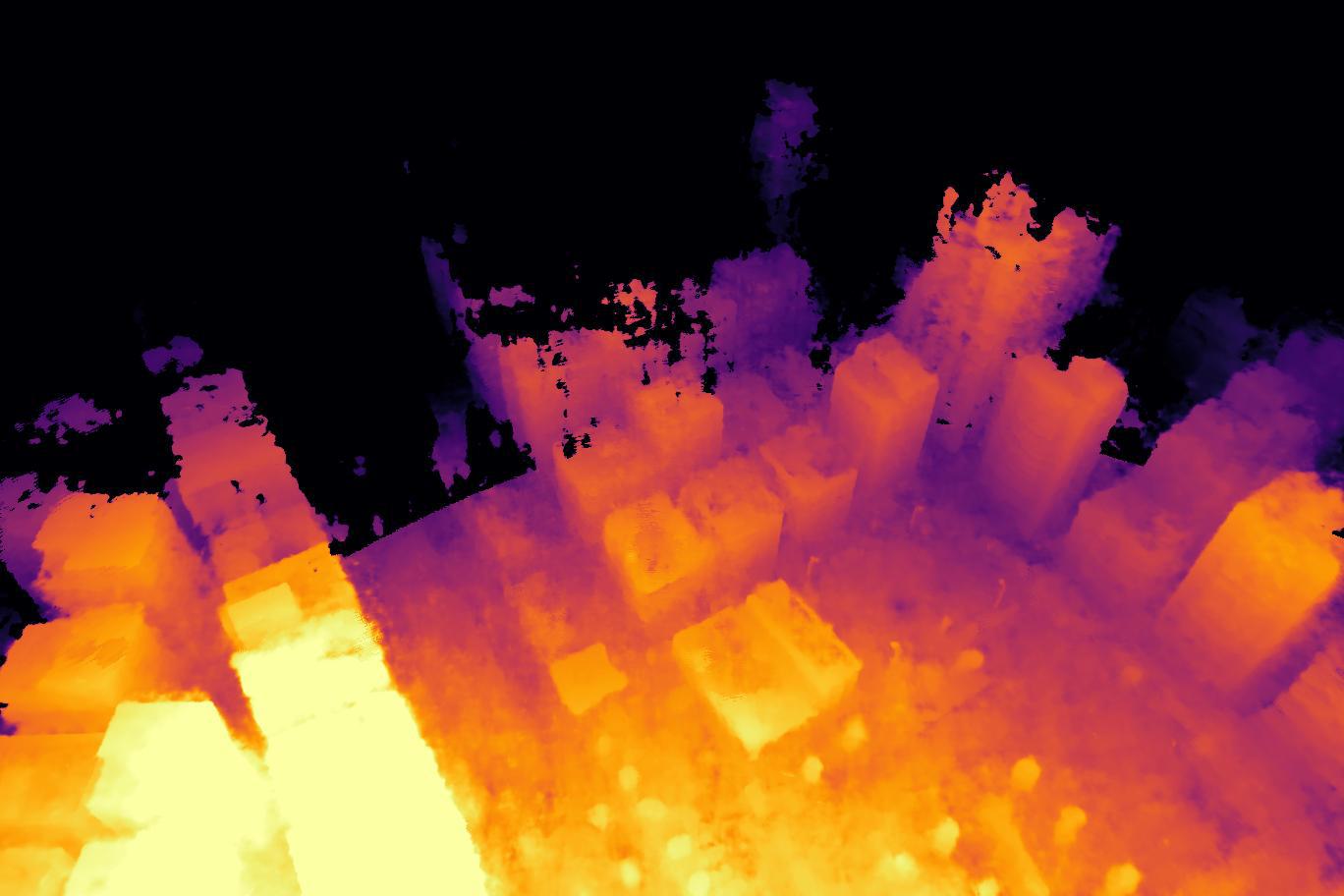} & 
        \includegraphics[clip,width=0.24\hsize]{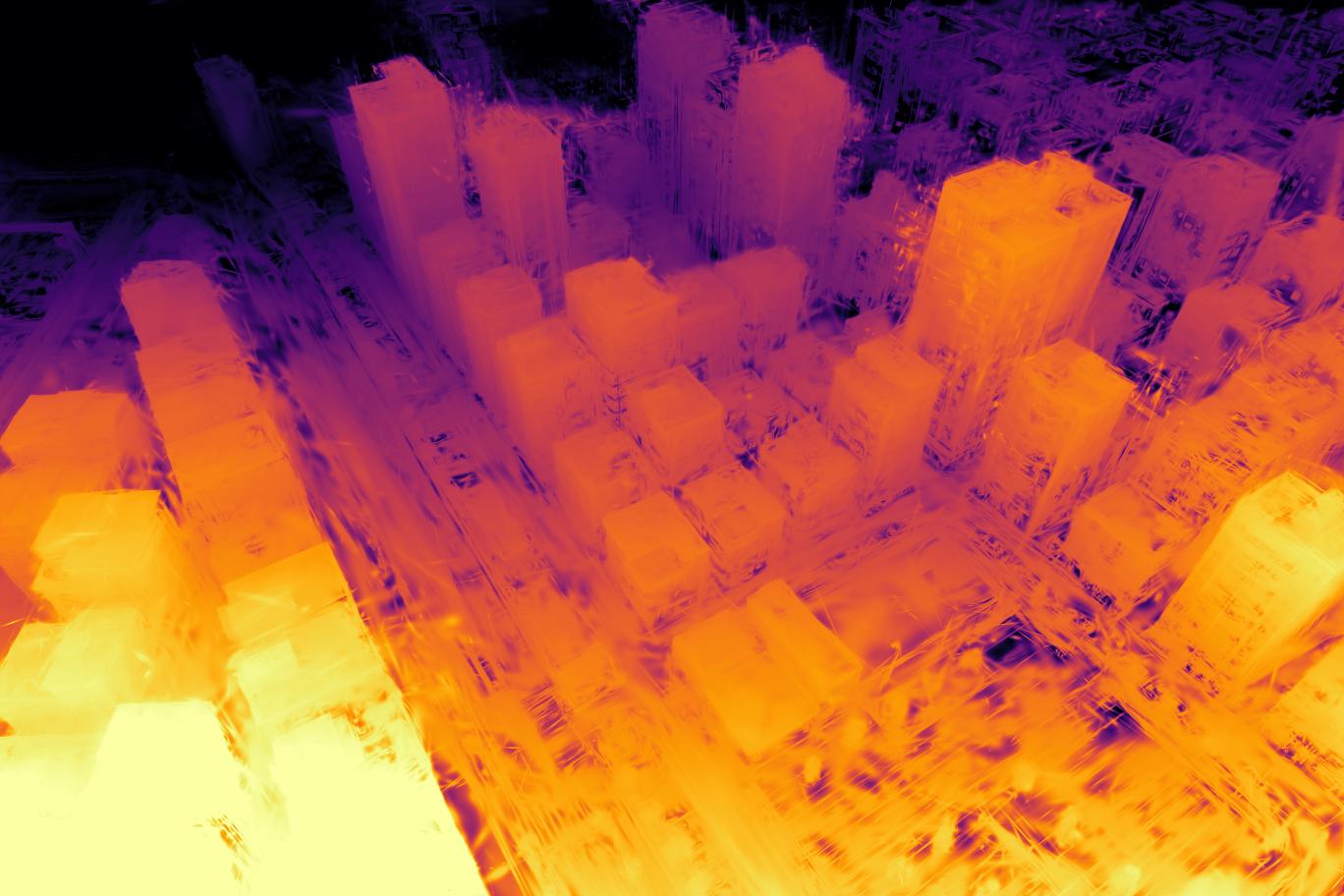} \\

        (a) Ground-Truth & (b) Mega-NeRF~\cite{Turki_2022_CVPR} & (c) Switch-NeRF~\cite{mi2023switchnerf} & (d) Fed3DGS
    \end{tabular}
    \caption{Rendered RGB images and depth images from test views. Fed3DGS captures the detailed structures (\textit{e.g.}, the steel frames in the top row and the distant buildings in the bottom row).}
    \label{fig:qualitative-comparison}
\end{figure}

Our method demonstrates the highest SSIM and LPIPS across most scenes, except for the Sci-Art and Quad 6k scenes.
This suggests that our approach effectively models the structure of the scenes.
As shown in Fig. \ref{fig:qualitative-comparison}, Fed3DGS indeed produces sharper images than Mega-NeRF and Switch-NeRF, which is consistent with the high SSIM of our method.
Unlike the other scenes, the Sci-Art scene has a building standing in the distance, and the Quad 6k scene has many moving, transient objects.
In addition, these scenes contain sky pixels, and 3DGS cannot handle such pixels having infinite depth.
We believe these factors degrade the quality of the initial points obtained from SfM and make training difficult, resulting in somewhat poor performance.

PSNR of our method is worse than that of centralized approaches.
In fact, our method sometimes fails to correctly represent appearance while structures are modeled well, resulting in somewhat poor PSNR (\eg, the color of the solar panels and grass in the top row of Fig. \ref{fig:qualitative-comparison}).
It indicates that our method does not model target appearance as well as the centralized approaches since PSNR is more sensitive to difference between pixel colors (\textit{i.e.}, appearance) than SSIM and LPIPS.
In federated learning, modeling appearance is more difficult than centralized approaches, including distributed training, because the number of images per client (\textit{i.e.}, variety of appearance) is much smaller than centralized approaches.
We analysis it in depth in the Appendix \ref{sec:fed3dgs-vs-3dgs}.
Accurately modeling appearance in federated learning is a challenging problem, and it will be future work.

Compared to the federated learning baseline, FedNeRF~\cite{fednerf}, our method shows better PSNR.
Subsequent ablation studies will illustrate that the improvement is attributed to our appearance model.

\begin{table*}[t]
\caption{Results on several benchmarks. For Fed3DGS, we report results averaged over five trials.}
\label{tab:main-res}
\resizebox{\textwidth}{!}{
\begin{tabular}{lccccccccc}
\toprule 
&\multicolumn{3}{c}{Building} & \multicolumn{3}{c}{Rubble} & \multicolumn{3}{c}{Quad 6k} \\
& $\uparrow$PSNR & $\uparrow$SSIM & $\downarrow$LPIPS & $\uparrow$PSNR & $\uparrow$SSIM & $\downarrow$LPIPS & $\uparrow$PSNR & $\uparrow$SSIM & $\downarrow$LPIPS \\ \hline

\multicolumn{4}{l}{\small\textit{- Centralized Training}}\\
\hspace{1em}NeRF~\cite{nerf}\xspace & 19.54 & 0.525 & 0.512
& 21.14 & 0.522 & 0.546
& 16.75 & 0.559 & 0.616 \\
\hspace{1em}Switch-NeRF~\cite{mi2023switchnerf}\xspace & \textbf{21.54} & 0.579 & 0.474
& \textbf{24.31} & 0.562 & 0.496
& - & - & - \\
\hspace{1em}GP-NeRF~\cite{zhang2023efficient}\xspace & 20.99 & 0.565 & 0.490  
& 24.08 & 0.563 & 0.497 
& 17.67 & 0.521 & 0.623 \\
\hline

\multicolumn{4}{l}{\small\textit{- Distributed Training}} \\
\hspace{1em}Mega-NeRF~\cite{Turki_2022_CVPR}\xspace  & 20.93 & 0.547 & 0.504
& 24.06 & 0.553 & 0.516
& \textbf{18.13} & \textbf{0.568} & 0.602 \\
\hspace{1em}Drone-NeRF~\cite{jia2024drone}\xspace & 18.46 & 0.490 & 0.469  
& 19.51 & 0.528 & 0.489
& - & - & - \\
\hline

\multicolumn{4}{l}{\small\textit{- Federated Learning}} \\
\hspace{1em}FedNeRF~\cite{fednerf}\xspace & 17.51 & - & -  
& 20.12 & - & -
& - & - & - \\
\hspace{1em}Fed3DGS\xspace & 18.66 & \textbf{0.602} & \textbf{0.362}
& 20.62 & \textbf{0.588} & \textbf{0.437}
& 15.41 & 0.528 & \textbf{0.485} \\

\toprule
&\multicolumn{3}{c}{Residence} & \multicolumn{3}{c}{Sci-Art} & \multicolumn{3}{c}{Campus} \\
& $\uparrow$PSNR & $\uparrow$SSIM & $\downarrow$LPIPS & $\uparrow$PSNR & $\uparrow$SSIM & $\downarrow$LPIPS & $\uparrow$PSNR & $\uparrow$SSIM & $\downarrow$LPIPS \\ \hline

\multicolumn{4}{l}{\small\textit{- Centralized Training}} \\
\hspace{1em}NeRF~\cite{nerf}\xspace & 19.01 & 0.593 & 0.488 
& 20.70 & 0.727 & 0.418 
& 21.83 & 0.521 & 0.630  \\
\hspace{1em}Switch-NeRF~\cite{mi2023switchnerf}\xspace & \textbf{22.57} & 0.654 & 0.457  
& \textbf{26.52} & \textbf{0.795} & 0.360
& \textbf{23.62} & 0.541 & 0.609 \\
\hspace{1em}GP-NeRF~\cite{zhang2023efficient}\xspace & 22.41 &  0.659 &  0.451 
& 25.56 & 0.783 & 0.373
& 23.46 & 0.544 & 0.611 \\
\hline

\multicolumn{4}{l}{\small\textit{- Distributed Training}} \\
\hspace{1em}Mega-NeRF~\cite{Turki_2022_CVPR}\xspace & 22.08 & 0.628 & 0.489
& 25.60 & 0.770 & 0.390
& 23.42 & 0.537 & 0.618 \\
\hspace{1em}Drone-NeRF~\cite{jia2024drone}\xspace & - & - & -  
& - & - & -
& - & - & - \\
\hline

\multicolumn{4}{l}{\small\textit{- Federated Learning}} \\
\hspace{1em}FedNeRF~\cite{fednerf}\xspace & - & - & -  
& - & - & -
& - & - & - \\
\hspace{1em}Fed3DGS\xspace & 20.00 & \textbf{0.665} & \textbf{0.344}
& 21.03 & 0.730 & \textbf{0.335}
& 21.64 & \textbf{0.635} & \textbf{0.436} \\

\bottomrule
\end{tabular}
}

\end{table*}

\subsection{Ablation Study}
We also verify the effectiveness of our method and its components.

\textbf{Scalability of the global model.}
We conduct a comparison between our method and FedNeRF~\cite{fednerf} based on training time and the global model size. 
The reported training time per client is measured on the NVIDIA V100 GPU, and the global model size is averaged across building and rubble scenes in Mill 19, as shown in Tab. \ref{tab:vs-fednerf}.
Our method exhibits a smaller model size than FedNeRF.
Furthermore, our approach incurs a shorter training time per client, thereby reducing computational costs on the client side.
\begin{table}[t]
\centering
\caption{Training time and a model size of ours and FedNeRF~\cite{fednerf} on Mill 19~\cite{Turki_2022_CVPR}.}
\label{tab:vs-fednerf}
\begin{tabular}{lcc} \toprule
                       &\hspace{1em} $\downarrow$Training Time/Client  & \hspace{1em} $\downarrow$Global Model Size\\ \hline
FedNeRF~\cite{fednerf} &   0.96 h   & 9.33 GB \\
Ours                   &   0.29 h   & 0.62 GB \\ \hline
\end{tabular}
\end{table}

\textbf{The effectiveness of distillation-based model updating.}
We compare the distillation-based model updating with two merge methods: the voxel grid filtering and replacement.
The former is widely used to merge point cloud, and the later just replaces Gaussians of the global model with the corresponding Gaussians of the local model.
The detailed procedure of the replacement is available in the Appendix \ref{sec:detailed-pro}.

We show the results in Tab. \ref{tab:merge-strategy}.
Note that since replacement and voxel grid filtering do not have the update scheme for appearance model, we do not apply appearance modeling for all methods in this comparison.
The performance of the replacement and voxel grid filtering is much worse than the proposed distillation-based update because they merge Gaussians based only on physical distance and ignore other factors, such as color and shape.
As a result, they compromise both geometry and appearance in the merged models.

\begin{table}[t]
\centering
\caption{Comparison between various merge strategies on Mill 19~\cite{Turki_2022_CVPR}.}
\label{tab:merge-strategy}
\begin{tabular}{lcccccc}
\toprule
                     & \multicolumn{3}{c}{Building}                               & \multicolumn{3}{c}{Rubble} \\
                     & $\uparrow$PSNR     & $\uparrow$SSIM    & $\downarrow$LPIPS & $\uparrow$PSNR    & $\uparrow$SSIM  & $\downarrow$LPIPS \\ \hline
Replacement          &       12.30        &        0.304      &        0.752      &       13.96       &       0.342     &        0.772      \\
Voxel Grid Filtering &       12.33        &        0.306      &        0.753      &       13.98       &       0.345     &        0.776      \\
Distillation         & \textbf{17.37}     &   \textbf{0.589}  &   \textbf{0.376}  &   \textbf{19.41}  & \textbf{0.580}  &   \textbf{0.431}  \\ \hline
\end{tabular}
\end{table}

\begin{figure}[t]
    \centering
    \includegraphics[clip,width=0.85\hsize]{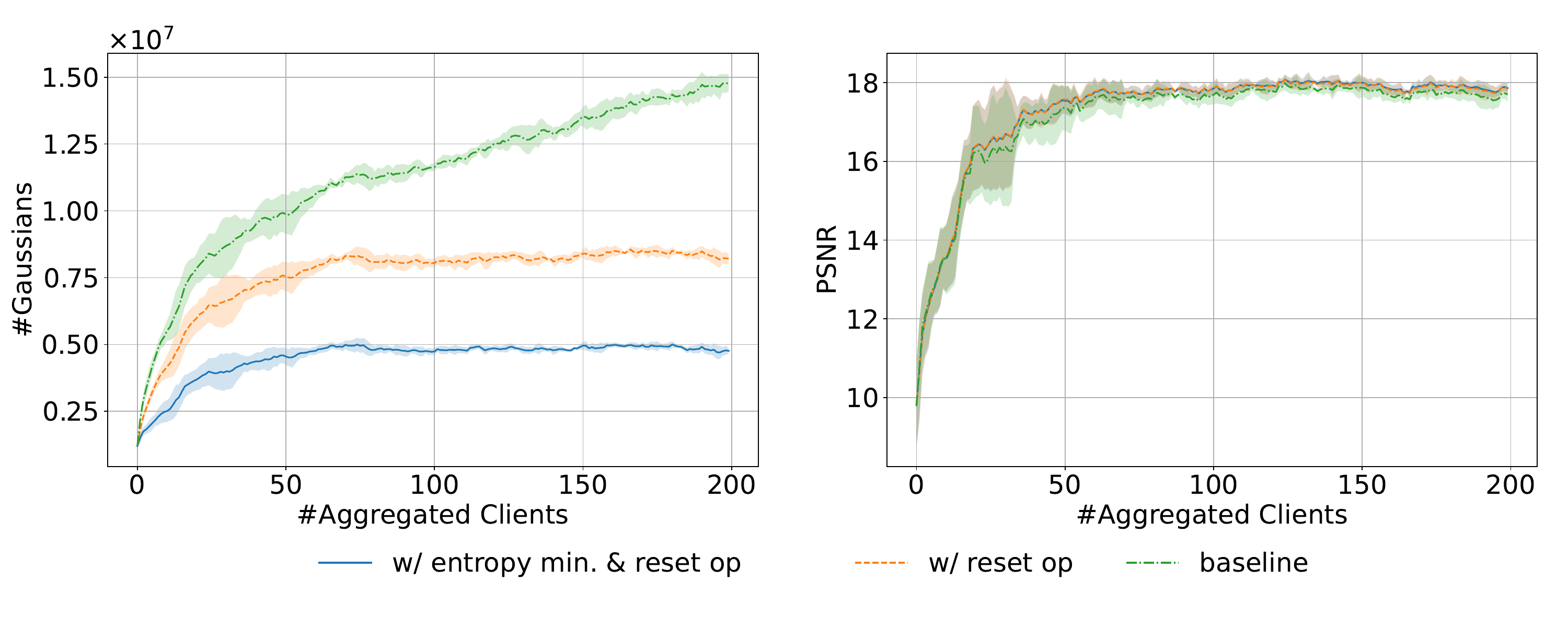}
    \caption{The number of Gaussians and PSNR with and without reset opacity (reset op) and entropy minimization (entropy min.) on the building scene. We show the mean and standard deviation over five trials. Baseline indicates a result without both reset opacity and entropy minimization.}
    \label{fig:transition}
\end{figure}

\textbf{The effectiveness of the randomly sampled cameras for distillation.}
We assess the effectiveness of the loss term for $\mathcal{C}_r$ in eq. \eqref{eq:dist-obj}.
As shown in Tab. \ref{tab:eff-cam}, the loss term significantly improves the quality of the rendered images because it can reduce the noisy Gaussians.
We visually assess it in the Appendix \ref{sec:qua-eval}.
\begin{table}[t]
\centering
\caption{Comparison between the models trained with eq. \eqref{eq:simple-dist-obj} and with eq. \eqref{eq:dist-obj}.}
\label{tab:eff-cam}
\begin{tabular}{lcccccc}
\toprule
                                & \multicolumn{3}{c}{Building}                             & \multicolumn{3}{c}{Rubble} \\
                                & $\uparrow$PSNR     & $\uparrow$SSIM  & $\downarrow$LPIPS & $\uparrow$PSNR    & $\uparrow$SSIM  & $\downarrow$LPIPS \\ \hline
Eq. \eqref{eq:simple-dist-obj}  &        15.70       &        0.489    &       0.464       &        17.13      &      0.496      &        0.526      \\
Eq. \eqref{eq:dist-obj}         &  \textbf{18.66}    &  \textbf{0.602} &  \textbf{0.362}   &  \textbf{20.62}   &  \textbf{0.588} &   \textbf{0.437}  \\ \hline
\end{tabular}
\end{table}

\begin{wrapfigure}[13]{r}[0pt]{0.4\textwidth}
    \centering
    \includegraphics[clip,width=1\hsize]{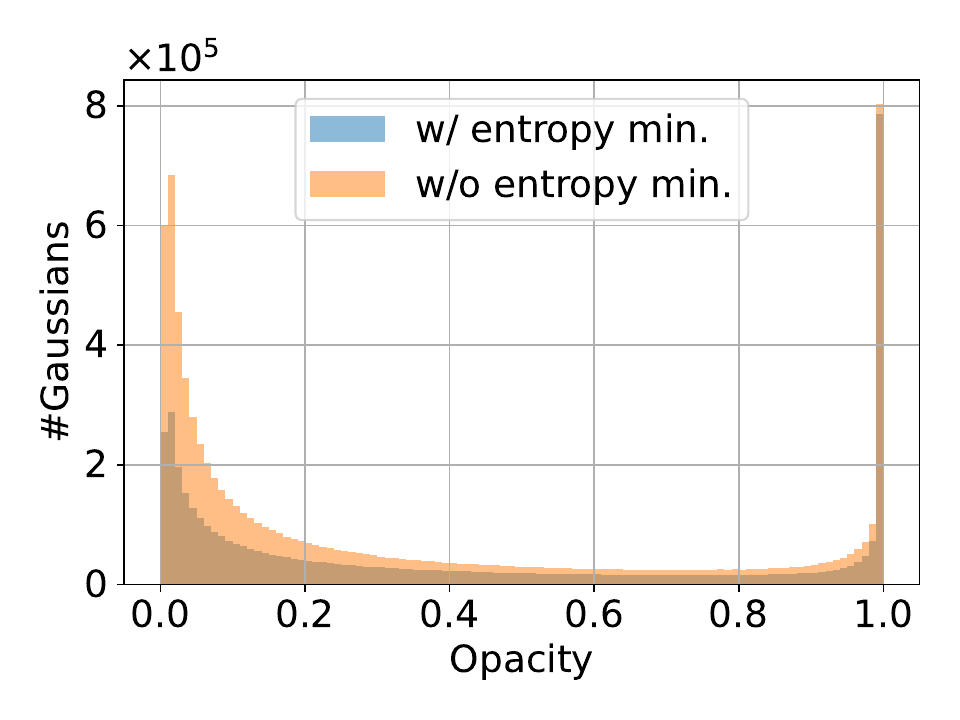}
    \caption{The histogram of the number of Gaussians.}
    \label{fig:histo-gauss}
\end{wrapfigure}

\textbf{The effectiveness of the reset opacity and the entropy minimization.}
We show the transition in the number of Gaussians and PSNR with and without reset opacity and the entropy minimization in Fig. \ref{fig:transition}.
In the baseline, which uses neither reset opacity nor entropy minimization, the number of Gaussians gradually increases with the number of aggregated clients.
The reset opacity prevents the increase of it.
The entropy minimization further reduces the number of Gaussians from the reset opacity while keeping PSNR.
We also show the histogram of the number of Gaussians for each opacity value in Fig. \ref{fig:histo-gauss}.
As expected, the entropy minimization prunes Gaussians of low contribution, \textit{i.e.}, low opacity.
We show the results for other datasets in the Appendix \ref{sec:add-ablation}.

\textbf{The effectiveness of the appearance modeling.}
We assess the effectiveness of the appearance model on the building and rubble scenes.
Note that although the images of these scenes are collected under the same lighting, appearance between images is a little different due to variations in camera exposure.
As shown in Tab. \ref{tab:app}, the appearance modeling improves the performance because the appearance model adjusts the camera exposure to the target image.
In fact, the relative improvement of PSNR is larger than SSIM and LPIPS.
We qualitatively compare them in the Appendix \ref{sec:qua-eval}.

\begin{table}[t]
\centering
\caption{The effect of appearance modeling. $\%$ denotes the relative improvement from the w/o appearance.}
\label{tab:app}
\begin{tabular}{lllllll}
\toprule
               & \multicolumn{3}{c}{Building}                               & \multicolumn{3}{c}{Rubble} \\
               & $\uparrow$PSNR     & $\uparrow$SSIM    & $\downarrow$LPIPS & $\uparrow$PSNR    & $\uparrow$SSIM  & $\downarrow$LPIPS \\ \hline
w/o appearance &        17.37       &        0.589      &       0.376       &       19.41       &       0.580     &  0.431   \\
w/ appearance  &   18.66{\tiny 7.4\%$\uparrow$}   &   0.602{\tiny 2.2\%$\uparrow$}  &  0.362{\tiny 1.6\%$\uparrow$}   &   20.62{\tiny 6.2\%$\uparrow$}  & 0.588{\tiny 1.4\%$\uparrow$}  &  0.437{\tiny 1.4\%$\downarrow$}   \\ \hline
\end{tabular}
\end{table}

\subsection{Appearance Modeling toward a Continuous Update Scenario}
In the scenario of continuously updating the global model, appearance would change due to seasonal changes.
Thus, we verify whether our framework can model appearance in such a case.

We simulate our method on a countryside scene in 4Seasons~\cite{wenzel2020fourseasons}, which contains images collected in different seasons.
We use images collected in summer and winter because there is a significant difference in appearance of them.
The detailed dataset properties are given in the Appendix \ref{sec:detailed-exp}.
The setup of federated learning is the same as other datasets.
We generate 400 clients for each season.
Let $\{^W\mathcal{G}_l^{(j)}\}_j$ and $\{^S\mathcal{G}_l^{(j)}\}_j$ be sets of local models trained with winter and summer data, respectively.
In this experiment, we compare four models:
$\mathcal{G}_g^W$ is updated with $\{^W\mathcal{G}_l^{(j)}\}_j$, and $\mathcal{G}_g^S$ is updated with $\{^S\mathcal{G}_l^{(j)}\}_j$.
Also, $\mathcal{G}_g^{S\rightarrow W}$ is updated with $\{^S\mathcal{G}_l^{(j)}\}_j$, and then further updated with $\{^W\mathcal{G}_l^{(j)}\}_j$, and $\mathcal{G}_g^{W\rightarrow S}$ is updated with $\{^W\mathcal{G}_l^{(j)}\}_j$, and then further updated with $\{^S\mathcal{G}_l^{(j)}\}_j$.

We show the results in Tab. \ref{tab:4seasons}.
Compared to $\mathcal{G}_g^S$, $\mathcal{G}_g^{S\rightarrow W}$ shows better performance in winter, which indicates that our framework can reflect changes in the scene.
Also, it shows better performance in summer than $\mathcal{G}_g^W$, which indicates that our appearance model represents appearance in summer.
Although there is a non-negligible gap between the performances of $\mathcal{G}_g^{S\rightarrow W}$ and $\mathcal{G}_g^S$ in summer, this is because the structure changes due to seasonal changes (\eg, trees leaf out in summer but drop their leaves in winter), and $\mathcal{G}_g^{S\rightarrow W}$ models the structure in winter.
There are the same underlying reasons for the performance of $\mathcal{G}_g^{W\rightarrow S}$.

\begin{table}[t]
\centering
\caption{Results on the countryside scene in 4Seasons~\cite{wenzel2020fourseasons}.}
\label{tab:4seasons}
\begin{tabular}{lcccccc}
\toprule
                               & \multicolumn{3}{c}{Summer}                                   & \multicolumn{3}{c}{Winter} \\
                               & $\uparrow$PSNR     & $\uparrow$SSIM    &  $\downarrow$LPIPS  & $\uparrow$PSNR    & $\uparrow$SSIM    & $\downarrow$LPIPS  \\ \hline
$\mathcal{G}_g^S$                &    18.26           &     0.659         &       0.486         &        11.03      &     0.440         &  0.563        \\
$\mathcal{G}_g^W$                &    12.31           &     0.497         &       0.563         &        15.82      &     0.594         &  0.512    
                               \\
$\mathcal{G}_g^{S\rightarrow W}$ &    14.00           &     0.532         &       0.529         &        14.78      &     0.534         &  0.534         \\
$\mathcal{G}_g^{W\rightarrow S}$ &    17.28           &     0.645         &       0.506         &        13.19      &     0.515         &  0.547         \\ \hline
\end{tabular}
\end{table}

\section{Conclusion}
We introduced the federated learning framework with 3DGS, which consists of a distillation-based model update scheme tailored for 3DGS with the appearance modeling.
Our framework demonstrated superior results to FedNeRF, which is a federated learning framework for NeRF, in terms of both the rendered image quality and the scalability, and results comparable to baselines, including a state-of-the-art approach, on several benchmarks.
In addition, we demonstrated that our framework can reflect changes in the scene and our appearance modeling is capable of handling changes in appearance due to seasonal variations.

\textbf{Limitation.}
In this work, we do not consider common problems in federated learning, such as clients with limited computational resources.
Specifically, since clients in our framework reconstruct scenes from scratch, they require relatively robust computational capabilities.
While this challenge might be mitigated by distributing a portion of the global model, potential issues with increased communication costs may arise instead.

%
%
\bibliographystyle{splncs04}
\bibliography{main}

\appendix
\section{Problems in Federated Learning for 3D Reconstruction}
\label{sec:minor-problem}
We describe some problems in federated learning for 3D reconstruction here.

Unlike distributed training such as Mega-NeRF, the area covered by each client is not uniform.
In other words, the distributed training can divide the scene into regular grid, but it would be difficult for federated learning to control the region covered by each client in realistic scenarios.
The non-uniform division would make the quality of the local models uneven.
In addition, appearance differs for each client in federated learning while the distributed training can control it, which makes the task of appearance modeling more complex than the distributed training.

It is often difficult for clients to collect data from multiple viewpoints, especially if clients are cars.
Specifically, the camera mounted on cars is often forward facing, and clients rarely drive the same road from different directions.
In this case, variations of the viewpoints are limited, and 3D reconstruction from such limited viewpoints can adversely impact quality.
We have not addressed this specific problem in this work, and exploring 3D reconstruction under limited viewpoints is left for future work.

In NeRF-based SLAM scenarios~\cite{sucar2021imap,zhu2022nice}, it is reported that when the training data cover only a part of the scene, MLP-based NeRF often forgets scenes that the data do not cover.
Common FL approaches, such as FedAVG~\cite{mcmahan2017communication}, can lead to the forgetting problem because each client only observes a part of scenes and trains a model from this partial observation.
Explicit representation realizing local updates, such as voxel grids and point clouds, is useful to prevent the forgetting problem because it does not affect the unobserved scene~\cite{zhu2022nice}.
Thus, the locality of model updates is also crucial to prevent a forgetting problem.

In global pose alignment, the difference in appearance between the local and global models can pose challenges.
Although the global pose alignment is performed by minimizing the difference between the rendered images in FedNeRF~\cite{fednerf}, the vision-based pose alignment may be trapped in the local minima if the appearance of the rendered images differs.

\section{Detailed Procedures}
\label{sec:detailed-pro}
\subsection{Optimization Procedure for Distillation}
We update the global model based on eq. \eqref{eq:dist-obj} using stochastic gradient descent.
Let $\{\mathcal{I}^{(j)}_d\}_j$ be the union of sets of images $\{\hat{I}^{(j)}_l\}_j$ and $\{\hat{I}^{(k)}_g\}_k$, which are the rendered images through the local and global models in eq. \eqref{eq:dist-obj}, respectively.
Also, let $\{\mathbf{K}^{(j)}_d,\mathbf{E}^{(j)}_d\}_j$ be the corresponding camera parameters.
Then, our distillation objective with the entropy minimization is represented as follows:
\begin{align}
    \nonumber
    \underset{\{\hat{o}^{(i)}\}_i,\phi_g,\{\hat{\ell}^{(j)}\}_j}{\arg\min}\ \mathbb{E}_{(\hat{I}_d^{(j)}, \mathbf{K}_d^{(j)}, \mathbf{E}_d^{(j)})\sim\mathcal{D}_d}[\mathcal{L}_\text{3dgs}(\hat{I}^{(j)}_d,\hat{R}(\mathbf{K}^{(j)}_d, \mathbf{E}^{(j)}_d, \hat{\mathcal{G}}_g,\phi_g,\hat{\ell}^{(j)})) \\
    + \eta\mathcal{L}_\text{entropy}(\mathbf{K}_d^{(j)}, \mathbf{E}_d^{(j)},\hat{\mathcal{G}}_g)],
\end{align}
where $\mathcal{D}_d$ denotes a set of triplets $\{(\hat{I}_d^{(j)}, \mathbf{K}_d^{(j)}, \mathbf{E}_d^{(j)})\}_j$.
We solve the above problem with the stochastic gradient descent, as in 3DGS~\cite{3dgs}.
Specifically, we randomly sample a triplet $(\hat{I}_d^{(j)}, \mathbf{K}_d^{(j)}, \mathbf{E}_d^{(j)})$ from $\mathcal{D}_d$ and compute gradients for each parameter and update them with the gradient descent.
For repeating this step, we optimize opacity, the appearance model, and the appearance vectors.
Note that $\{\hat{\ell}^{(j)}\}_{j}$ is dismissed after every model update step.

\subsection{Search Range of Reset Opacity}
\label{sec:serach-range}
We reset the opacity of the local model and those of the global model around Gaussians of the local model before the distillation-based model update.
We search for Gaussians in the global model around those of a local model using range search, which involves searching for Gaussians within a range of $\epsilon$ from the local model's Gaussians.

We determine the search range, $\epsilon$, based on the density of the local model's Gaussians.
Specifically, we compute the Euclidean distance between the 3D position of $G^{(i)}\in\mathcal{G}_l$ and that of its nearest neighbor Gaussian $G^\text{nearest}\in\mathcal{G}_l\setminus G^{(i)}$ as $D^{(i)}=D(\mathbf{x}^{(i)},\mathbf{x}^\text{nearest})$, and we use its median of $\{D^{(i)}\}_i$ as the search range.

\subsection{Procedure of Replacement to Merge 3DGS}
As a baseline to merge two 3DGS, we evaluate the replacement method in Tab. \ref{tab:merge-strategy}, which replaces Gaussians of the global model with the corresponding Gaussians of a local model.
The corresponding Gaussians are identified using the range search.
Specifically, we search for the Gaussians of the global model within the range of $\epsilon$ from the Gaussians of the local model.
The located Gaussians are then removed from the global model, and the local model's Gaussians are appended to the global model.
The search range $\epsilon$ is determined with the same procedure as in the reset opacity.

\section{Details of Experiment Setup}
\label{sec:detailed-exp}
\textbf{Resources.}
We train local models on NVIDIA V100 GPUs with 16 GB and update the global model on an NVIDIA A100 GPU with 40 GB.

\textbf{Hyperparameters.}
We train each local model for 20,000 iterations with each local data.
The training protocols and hyperparameters are the same as in 3DGS~\cite{3dgs}, except that the density gradient threshold is set to 0.0004, opacity reset interval is set to 1,000, and the degree of spherical harmonics is set to 2.
For the detailed training protocols and other hyperparameters, please refer to \cite{3dgs}.

For updating the global model, we use the AdamW optimizer~\cite{loshchilov2017decoupled} and set learning rates for opacity, MLP, hash encoding, and appearance vectors to 0.05, 0.0001, 0.0001, and 0.001, respectively.
We apply weight decay to MLP, and its coefficient is set to 0.0001.
The number of epochs for each update step is set to 5 in the experiments.

For client selection, we select a client where there are more than 20 duplicates in the local and global data, $\mathcal{C}_l$ and $\mathcal{C}_g$.
In the sampling step of $\mathcal{C}_r$, we filter out the duplicated cameras from $\mathcal{C}_g$.
Note that the selection and filtering procedures described can only be performed in a simulation setting, as duplicates do not exist in realistic situations.
However, by evaluating the similarity between the camera extrinsic matrices in $\mathcal{C}_l$ and $\mathcal{C}_g$, we can apply the same strategy.

\textbf{Local data sampling.}
We randomly sample one training sample and compute distance between the sampled data's camera origin and the other camera origins.
Then, we collect $k$-nearest neighbors, where $k$ is randomly selected in a range of 100 to 200.
We refer to the collected data as local data and generate local data sets by repeating this procedure, as in \cite{fednerf}.
We note that this sampling strategy is simple but does not fully reflect the realistic situation.
For example, as discussed in Sec. \ref{sec:minor-problem}, the viewpoints in the local data would be limited in realistic situations.
However, it is not obvious that there is a straightforward way to sample more realistic local data from existing datasets, and it is an open problem.

\textbf{Dataset properties.}
We show the number of data and image size for each dataset in Tab. \ref{tab:dataset}.
Note that the 4Seasons dataset provides stereo camera data, but to evaluate our method in the monocular camera setting, we use only the left camera images with the GNSS poses in the experiments.
\begin{table}[]
    \centering
    \caption{The number of images and the resolution for each dataset. The bottom of the table denotes the 4Seasons countryside scene.}
    \label{tab:dataset}
    \begin{tabular}{cccccccc}\toprule
           & Building & Rubble & Quad 6k & Residence & Sci-Art & Campus \\ \hline
\#Images   & 1,940 & 1,678 & 5,147 & 2,582 & 3,019 & 5,871\\
Image Size & 4,608$\times$3,456 & 4,608$\times$3,456 & 1,708$\times$1,329 & 5,472$\times$3,648 & 4,864$\times$3,648 & 5,472$\times$3,648\\ \toprule
           & \multicolumn{3}{c}{Countryside (Summer)} & \multicolumn{3}{c}{Countryside (Winter)} \\ \hline
\#Images   & \multicolumn{3}{c}{6,971} & \multicolumn{3}{c}{6,139}\\
Image Size & \multicolumn{3}{c}{800$\times$400} & \multicolumn{3}{c}{800$\times$400}\\ \hline
    \end{tabular}

\end{table}

\begin{figure}[t]
    \centering
    \includegraphics[clip,width=0.8\hsize]{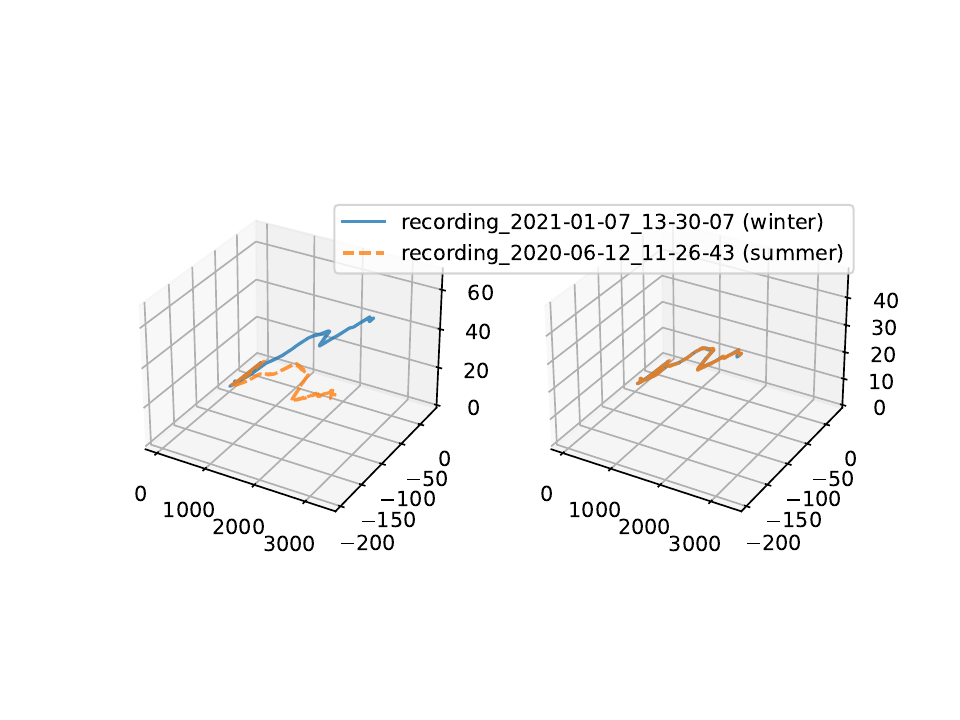}
    \caption{GNSS positions of the country side scene in 4Seasons. The left figure shows the given positions and the right figure shows them aligned by ICP.}
    \label{fig:gnss-poses}
\end{figure}

\textbf{Preprocess for 4Seasons.}
The GNSS poses of the 4Seasons dataset are not aligned between seasons (Fig. \ref{fig:gnss-poses}, left).
Thus, we align them by an iterative closest point (ICP) algorithm.
We compute rigid transformation parameters by ICP\footnote{We use ICP implemented in Open3D (\url{https://www.open3d.org/docs/release/tutorial/pipelines/icp_registration.html}).} and transform them.
The ICP result is shown in Fig. \ref{fig:gnss-poses}, right.

\begin{figure}[t]
    \centering
    \begin{tabular}{cc}
        \includegraphics[clip,width=0.4\hsize]{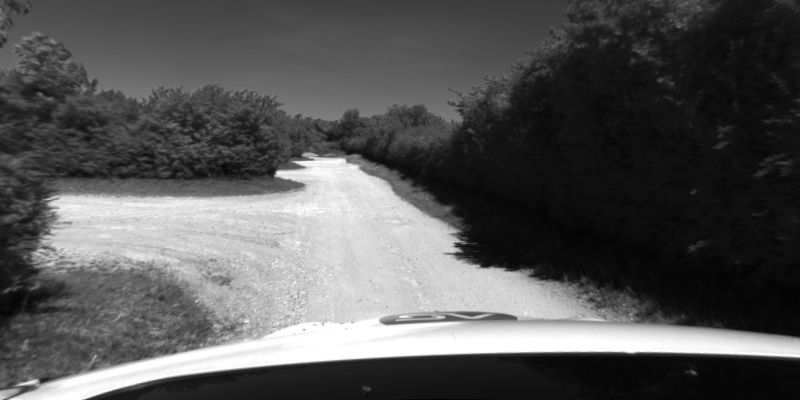} & 
        \includegraphics[clip,width=0.4\hsize]{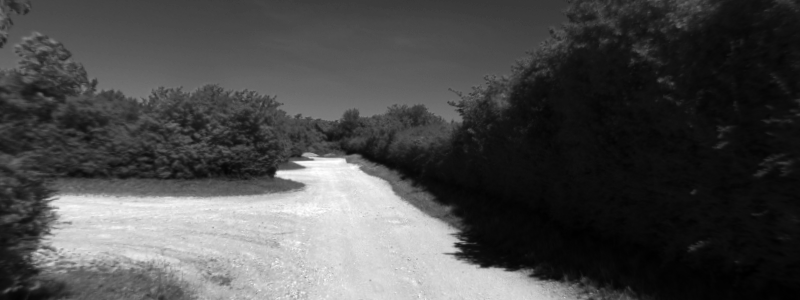} \\
        (a) & (b)
    \end{tabular}
    \caption{(a) An image in 4Seasons and (b) the image where the ego vehicle is trimmed.}
    \label{fig:4S-trim}
\end{figure}

Images in 4Seasons contain the ego vehicle, as shown in Fig. \ref{fig:4S-trim}~(a), and they make the evaluation inaccurate.
Thus, we remove 100 pixels from the bottom for training local models and evaluation, as shown in Fig. \ref{fig:4S-trim}~(b).

\textbf{Evaluation procedure.}
We follow Mega-NeRF's evaluation protocol~\cite{Turki_2022_CVPR}; namely, we use the left half of validation images for training to model appearance and evaluate models on the right half of them.
Since the distillation-based model update is performed with rendered images, and appearance vectors are dismissed after every update step, we do not have the appearance vectors for the validation images.
Thus, before the validation step, we optimize appearance vectors for validation images.
We initialize the appearance vectors as zero and optimized them by minimizing the mean squared error between the left half of the rendered and target images.
Optimization is performed with Adam~\cite{kingma2014adam} for 100 iterations for each image.
The learning rate is set to 0.05.

For the 4Seasons dataset, we sample validation data for every 100 training images from the training data ordered in a time series.

\section{Details of the Appearance Model}
\label{sec:detailed-app}
Our appearance model consists of three-layer perceptrons with hash encoding~\cite{mueller2022instant}.
For the hash encoding, we set the coarsest resolution to 32 and the finest resolution to 2048.
The number of scales is set to 16, and each scale has two-dimensional features.
Thus, the 3-dimensional coordinate is encoded to 32-dimensional features by the hash encoding.
We also set the dimension of the appearance vector, $d$, to 32.
The vector obtained by concatenating the 32-dimensional positional feature vector with the appearance vector is fed into the MLP.
The hidden layers of MLP have 64 units, and the ReLU activation is applied to the output of each hidden layer.
For initialization, the parameters of the output layer are set to zero so that the output vector is zero (\textit{i.e.}, the appearance model does not modify the coefficients of the spherical harmonics at the initial state).

\section{Qualitative Evaluation}
\label{sec:qua-eval}
\begin{figure}[t]
    \centering
    \begin{tabular}{cccc}
        \includegraphics[clip,width=0.24\hsize]{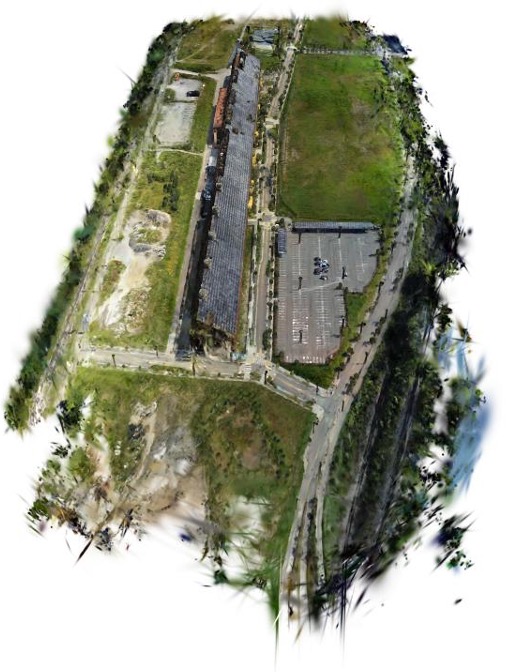} &
        \includegraphics[clip,width=0.24\hsize]{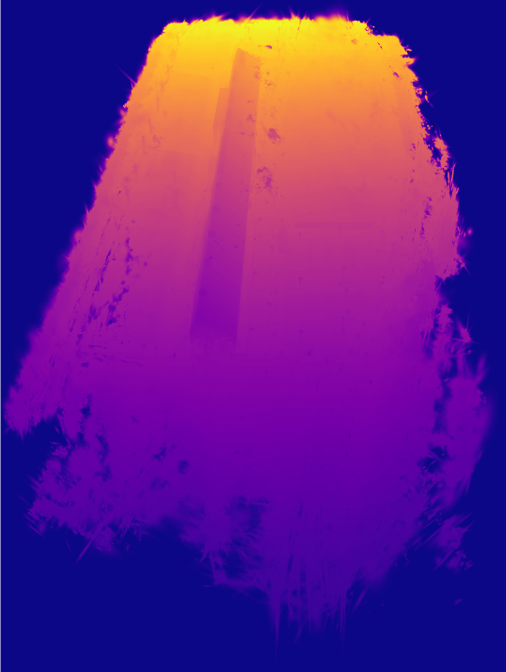} &
        \includegraphics[clip,width=0.24\hsize]{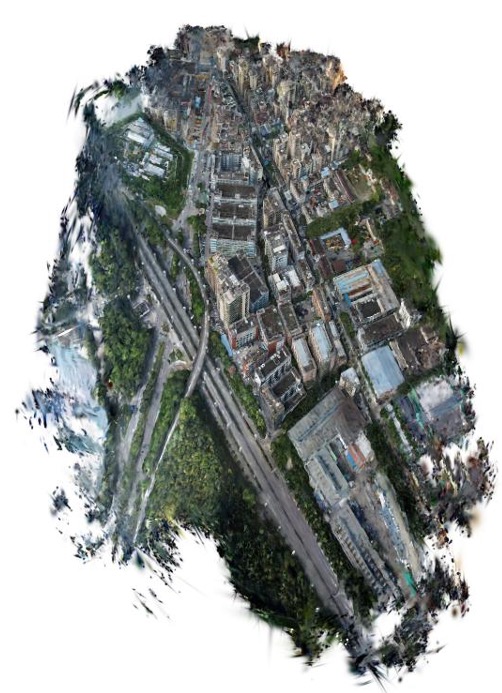} &
        \includegraphics[clip,width=0.24\hsize]{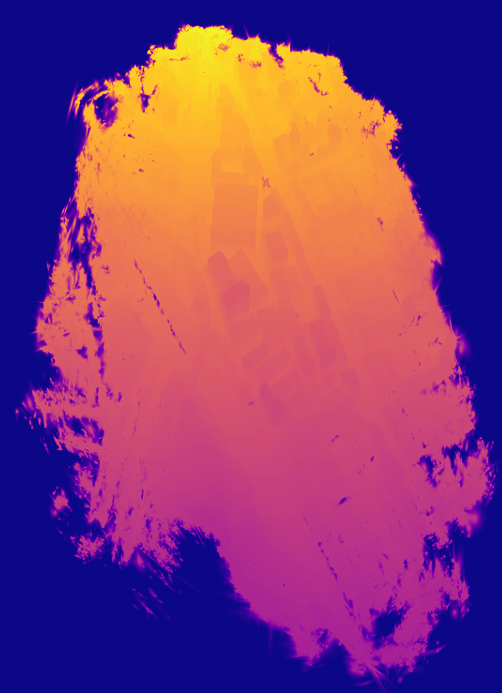}\\
        \multicolumn{2}{c}{(a) Building} & \multicolumn{2}{c}{(b) Residence} \\
        \includegraphics[clip,width=0.24\hsize]{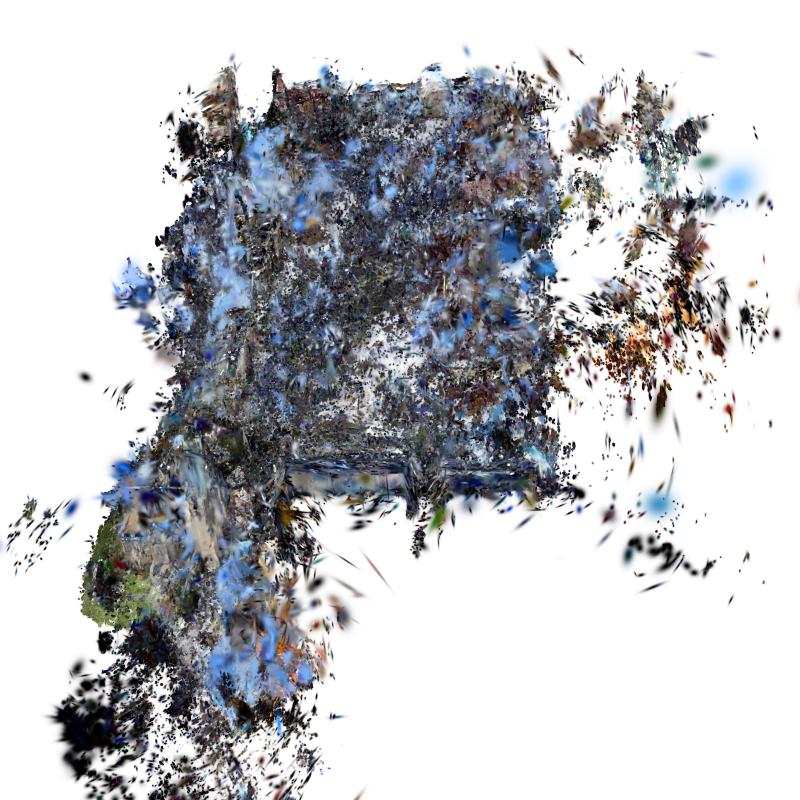} &
        \includegraphics[clip,width=0.24\hsize]{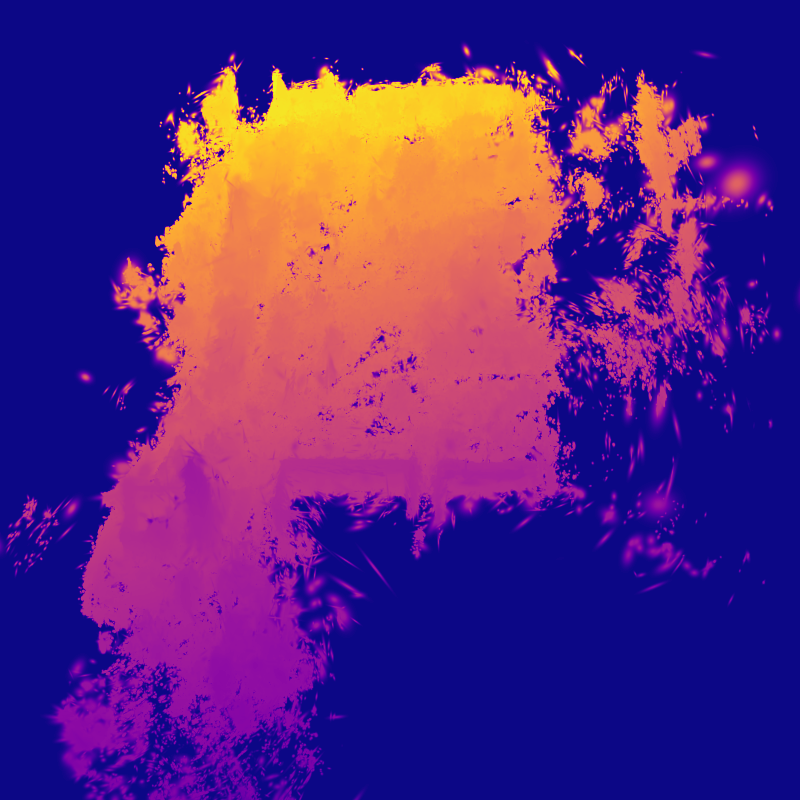} &
        \includegraphics[clip,width=0.24\hsize]{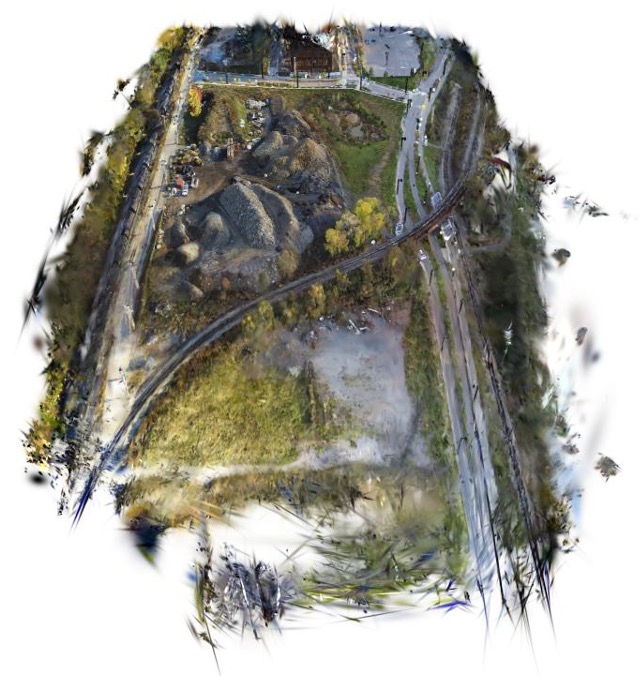} &
        \includegraphics[clip,width=0.24\hsize]{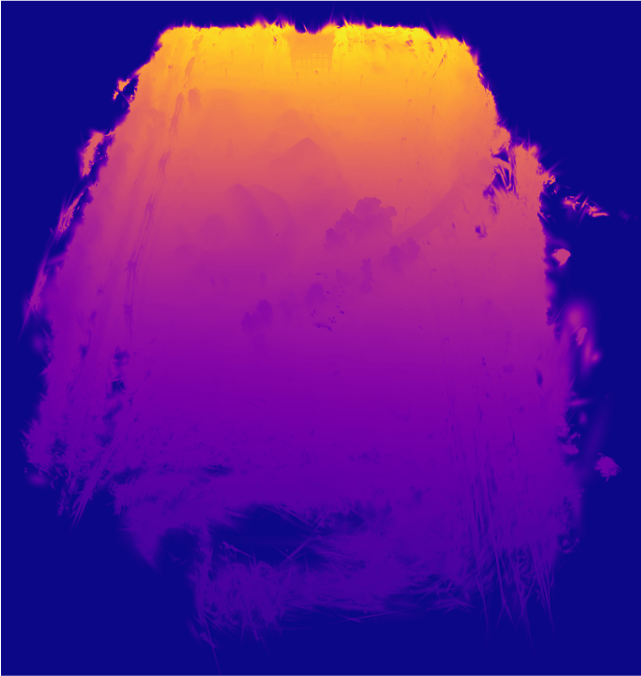}\\
        \multicolumn{2}{c}{(c) Quad 6k} & \multicolumn{2}{c}{(d) Rubble} \\
        \includegraphics[clip,width=0.24\hsize]{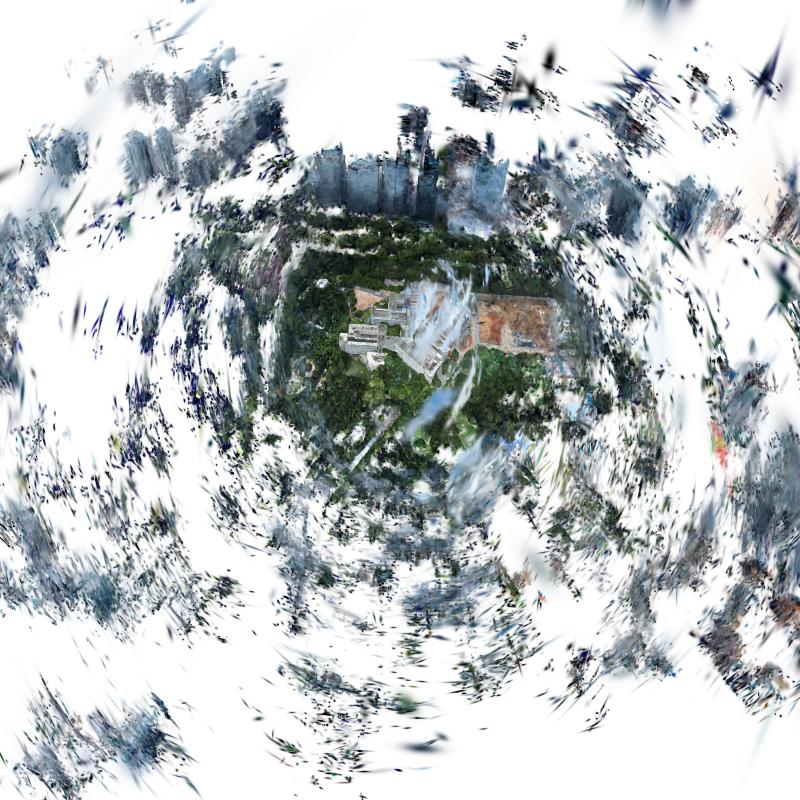} &
        \includegraphics[clip,width=0.24\hsize]{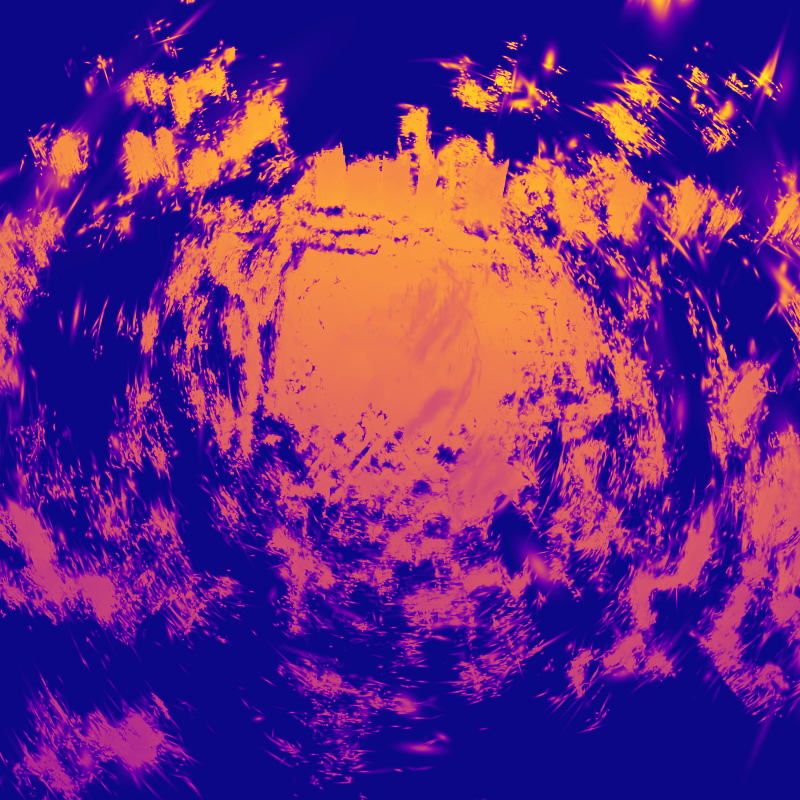} &
        \includegraphics[clip,width=0.24\hsize]{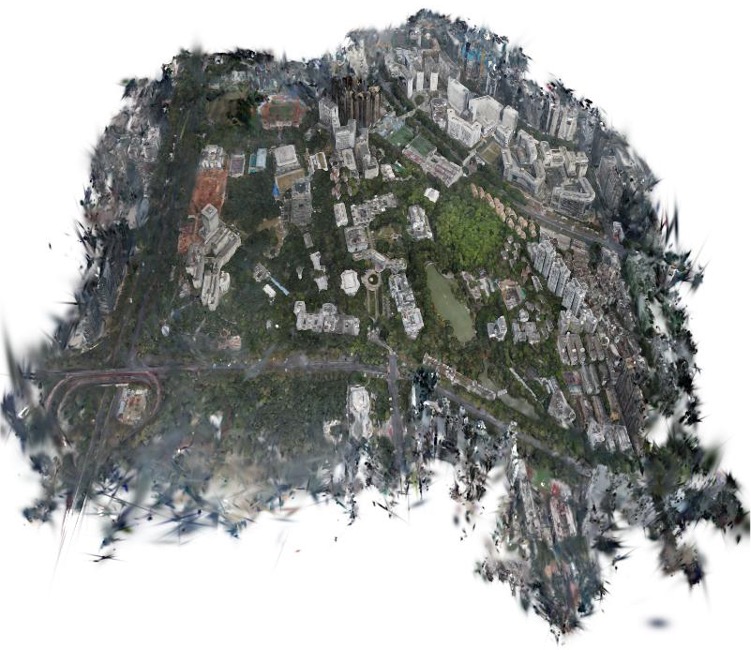} &
        \includegraphics[clip,width=0.24\hsize]{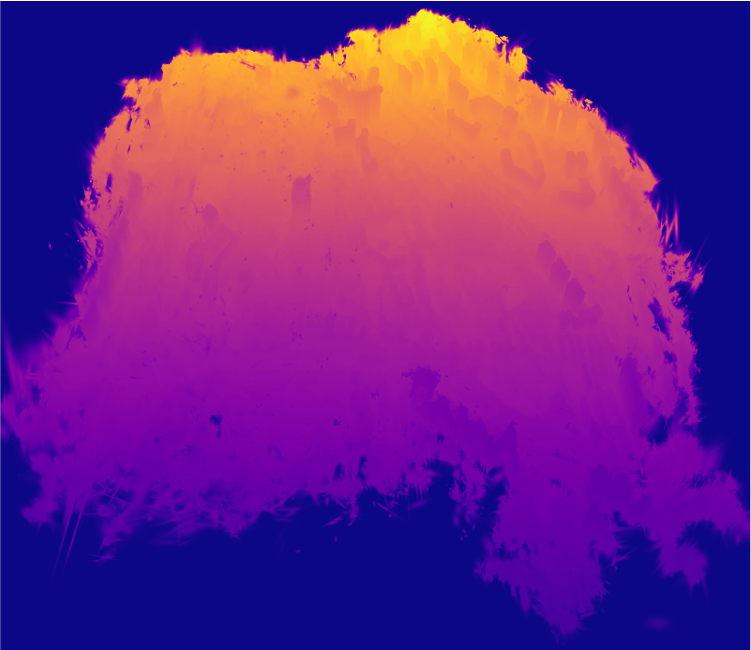}\\
        \multicolumn{2}{c}{(e) Sci-Art} & \multicolumn{2}{c}{(e) Campus}      
    \end{tabular}
    \caption{Top view of the global models with their depth.}
    \label{fig:topviews}
\end{figure}

\begin{figure}
    \centering
    \begin{tabular}{ccc}
        \includegraphics[clip,width=0.3\hsize]{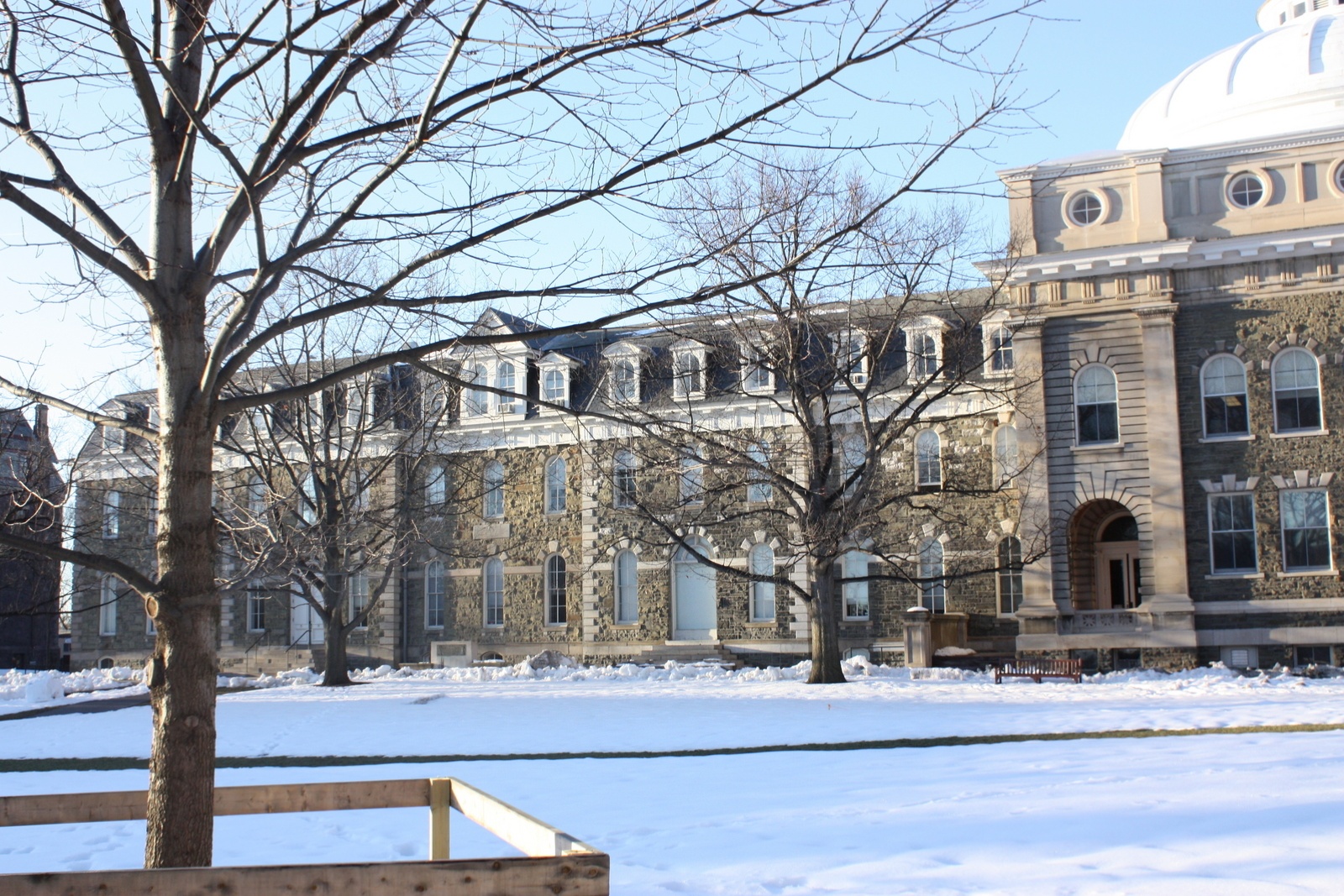} & 
        \includegraphics[clip,width=0.3\hsize]{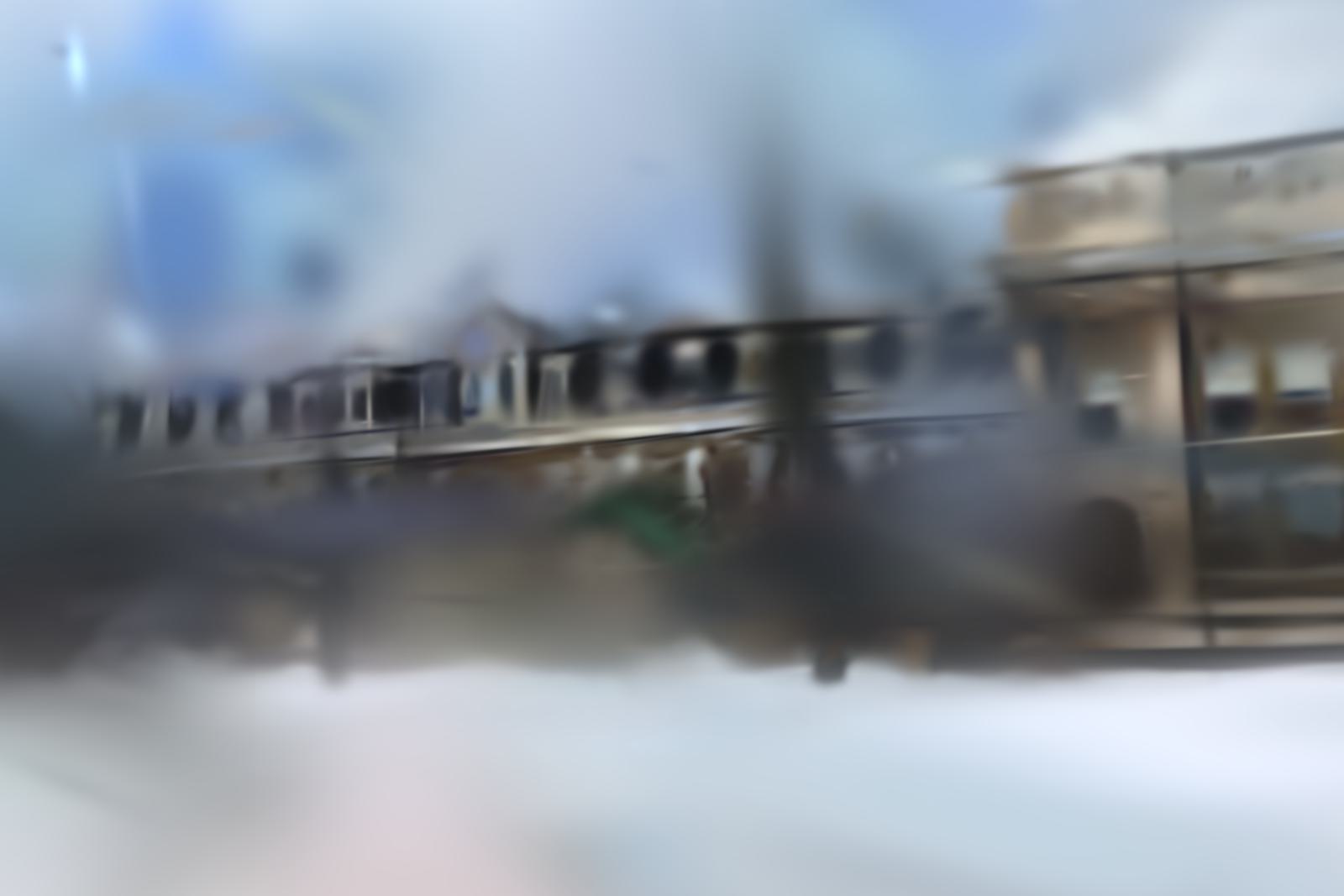} &
        \includegraphics[clip,width=0.3\hsize]{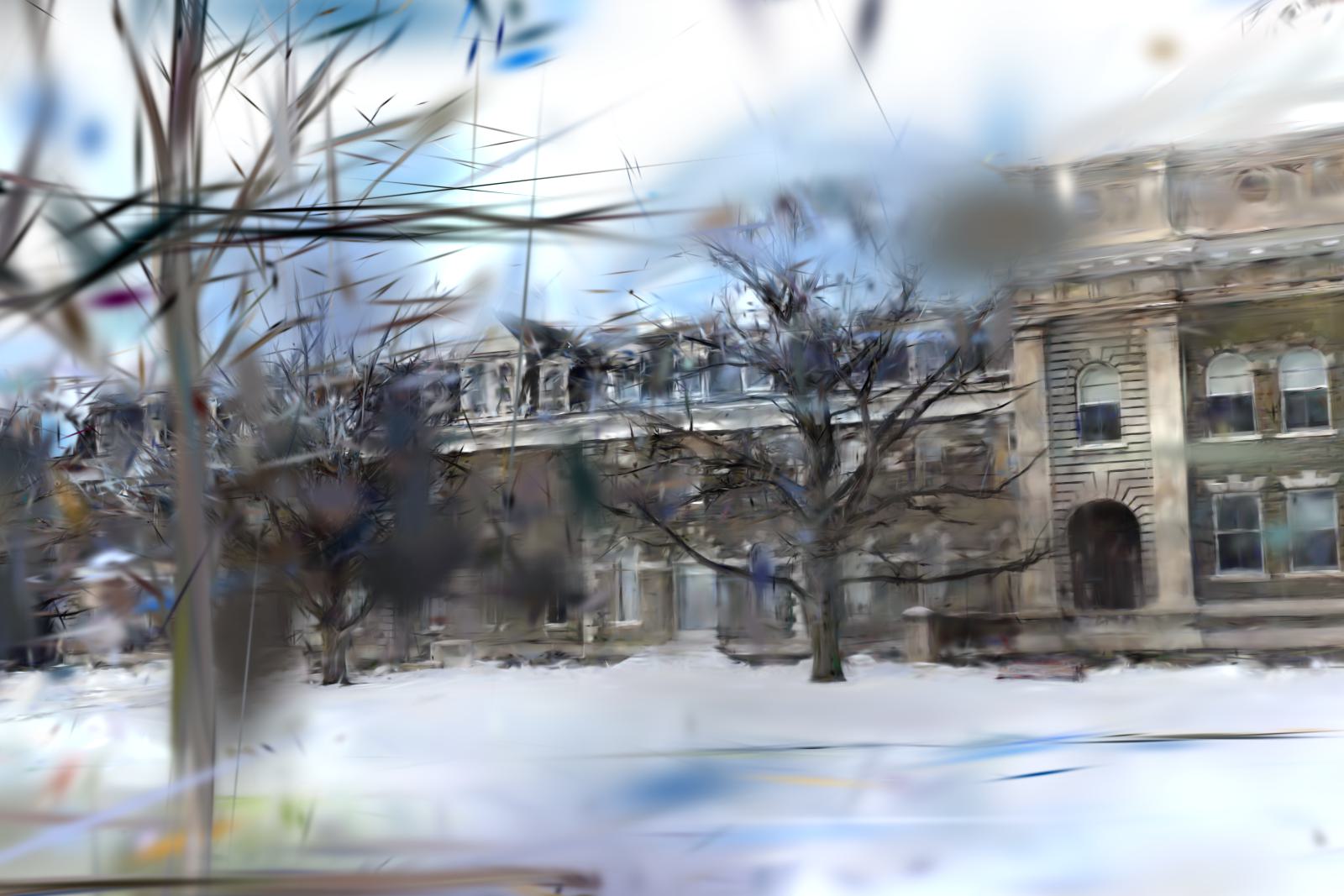} \\
        (a) Ground-Truth & (b) 3DGS & (c) Fed3DGS
    \end{tabular}
    \caption{Rendered images through 3DGS~\cite{3dgs} and Fed3DGS trained on Quad 6k.}
    \label{fig:quad}
\end{figure}

\begin{figure}
    \centering
    \begin{tabular}{cccc}
        \includegraphics[clip,width=0.24\hsize]{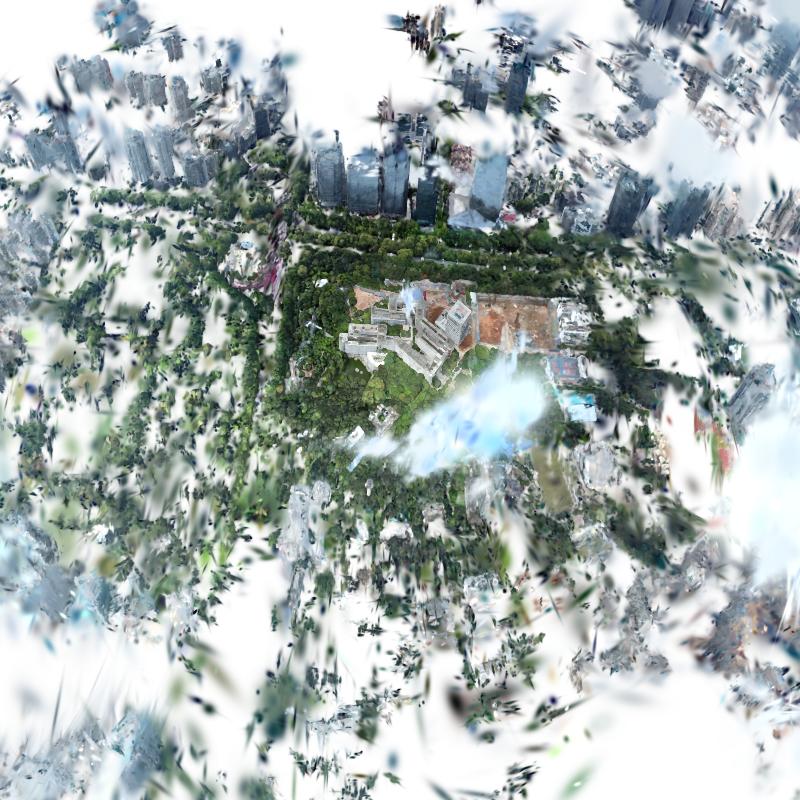} &  
        \includegraphics[clip,width=0.24\hsize]{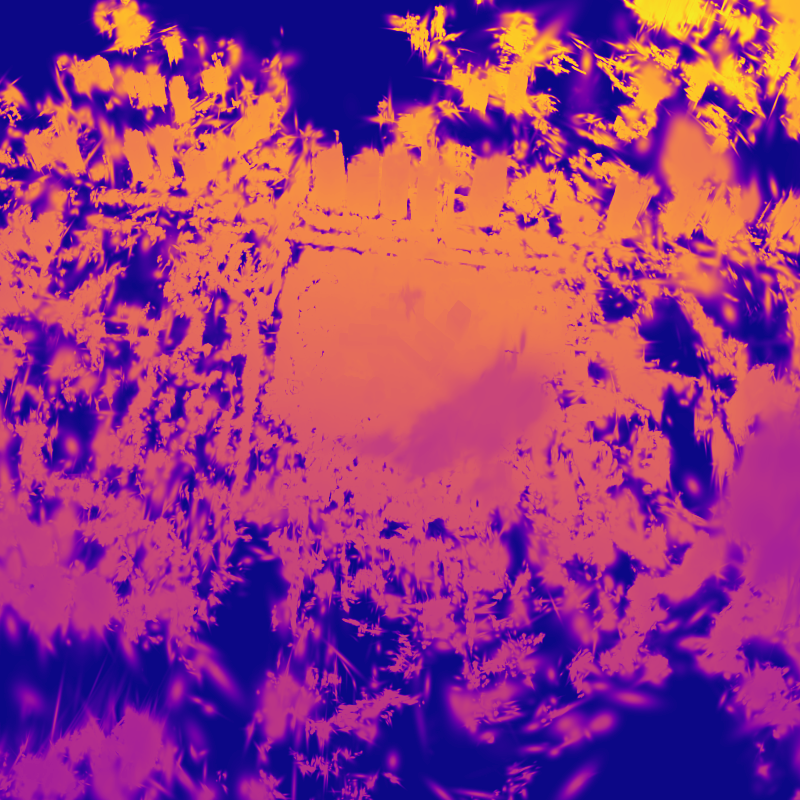} & 
        \includegraphics[clip,width=0.24\hsize]{images/topviews/fed/sci-art.jpg} &
        \includegraphics[clip,width=0.24\hsize]{images/topviews/fed/sci-art-depth.png}\\
        \includegraphics[clip,width=0.24\hsize]{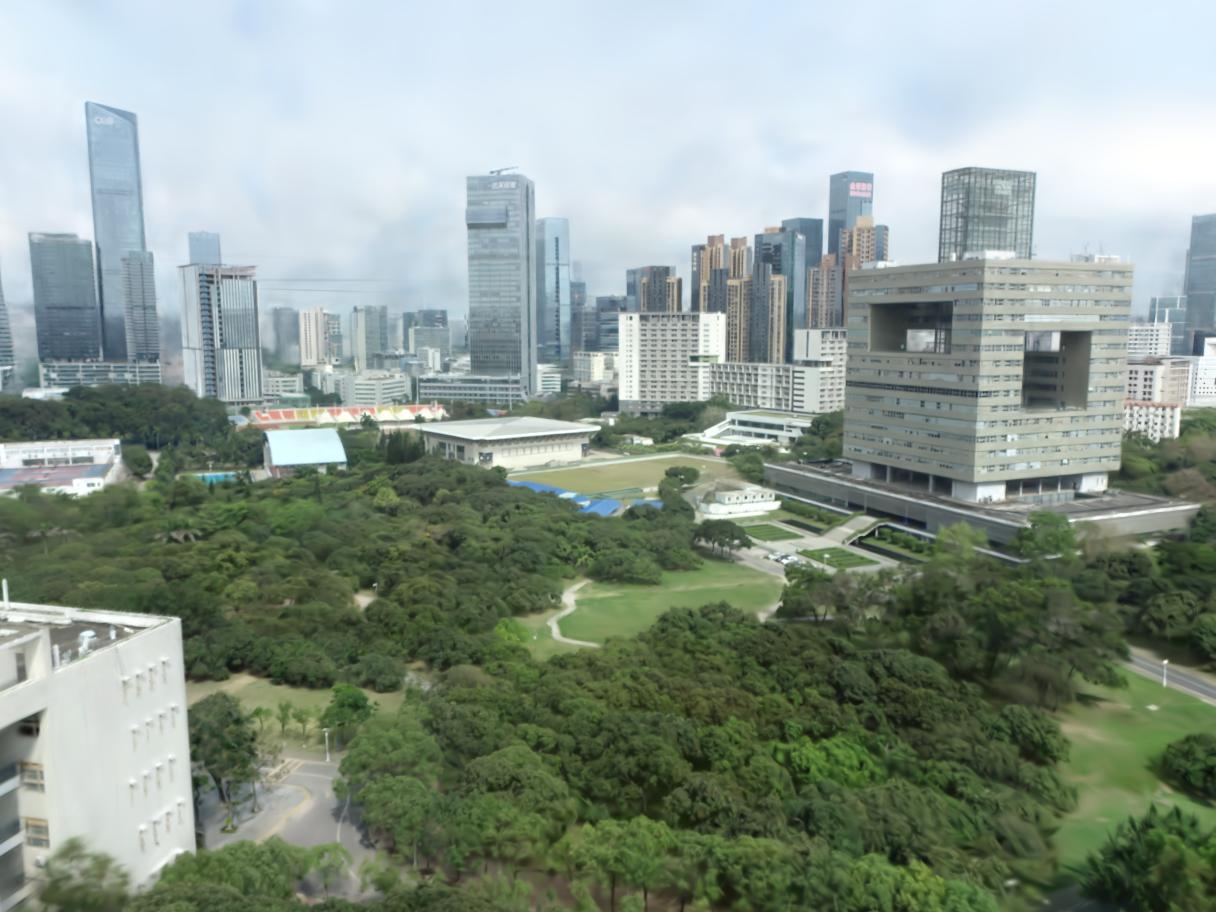} &  
        \includegraphics[clip,width=0.24\hsize]{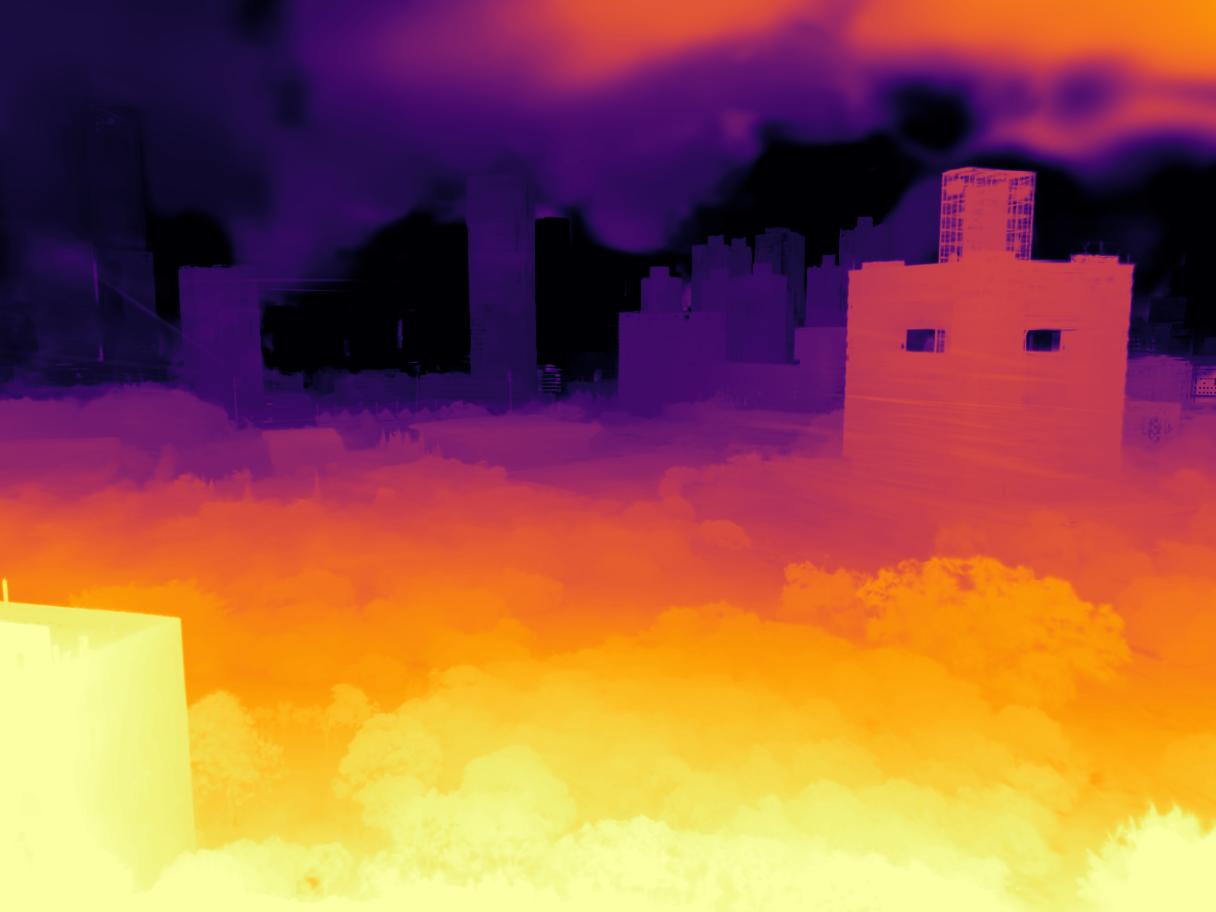} & 
        \includegraphics[clip,width=0.24\hsize]{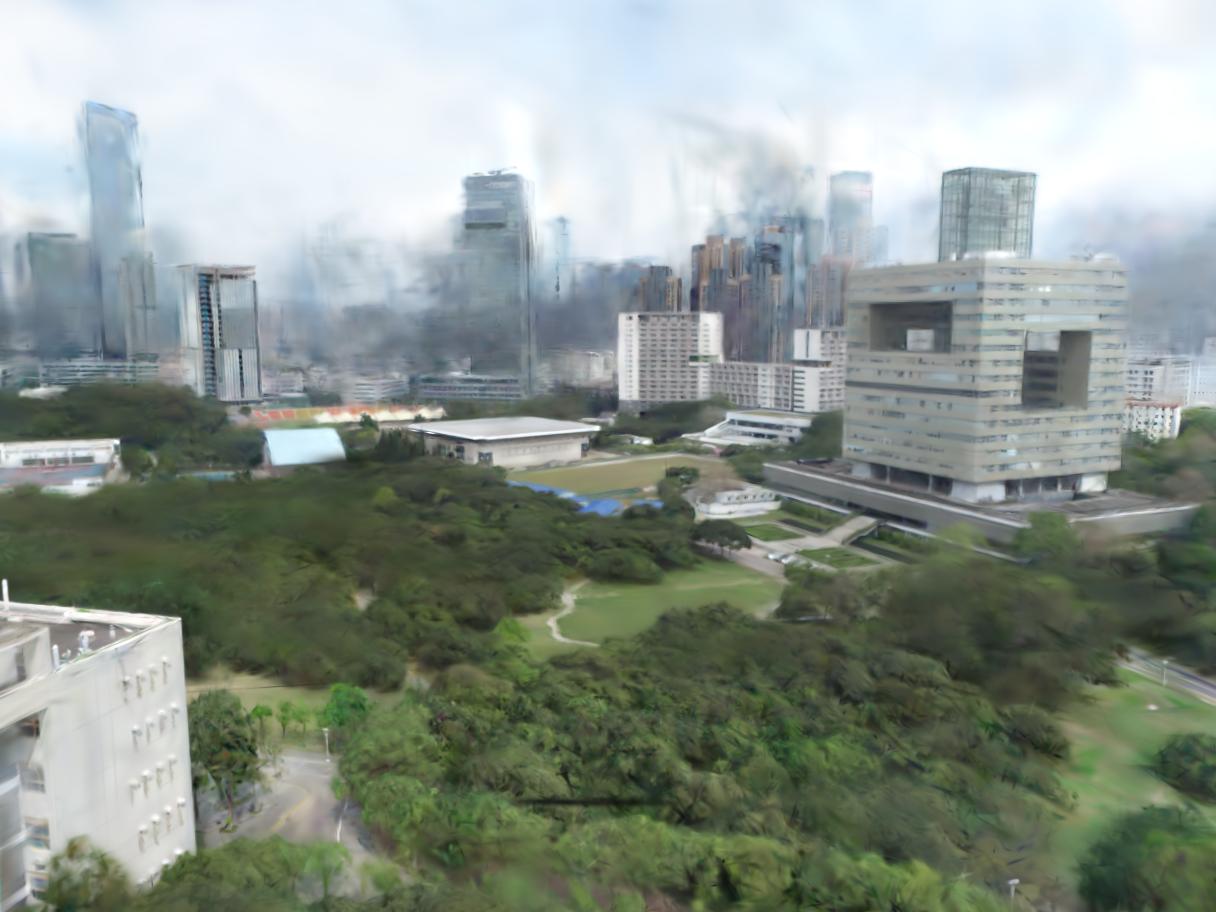} &
        \includegraphics[clip,width=0.24\hsize]{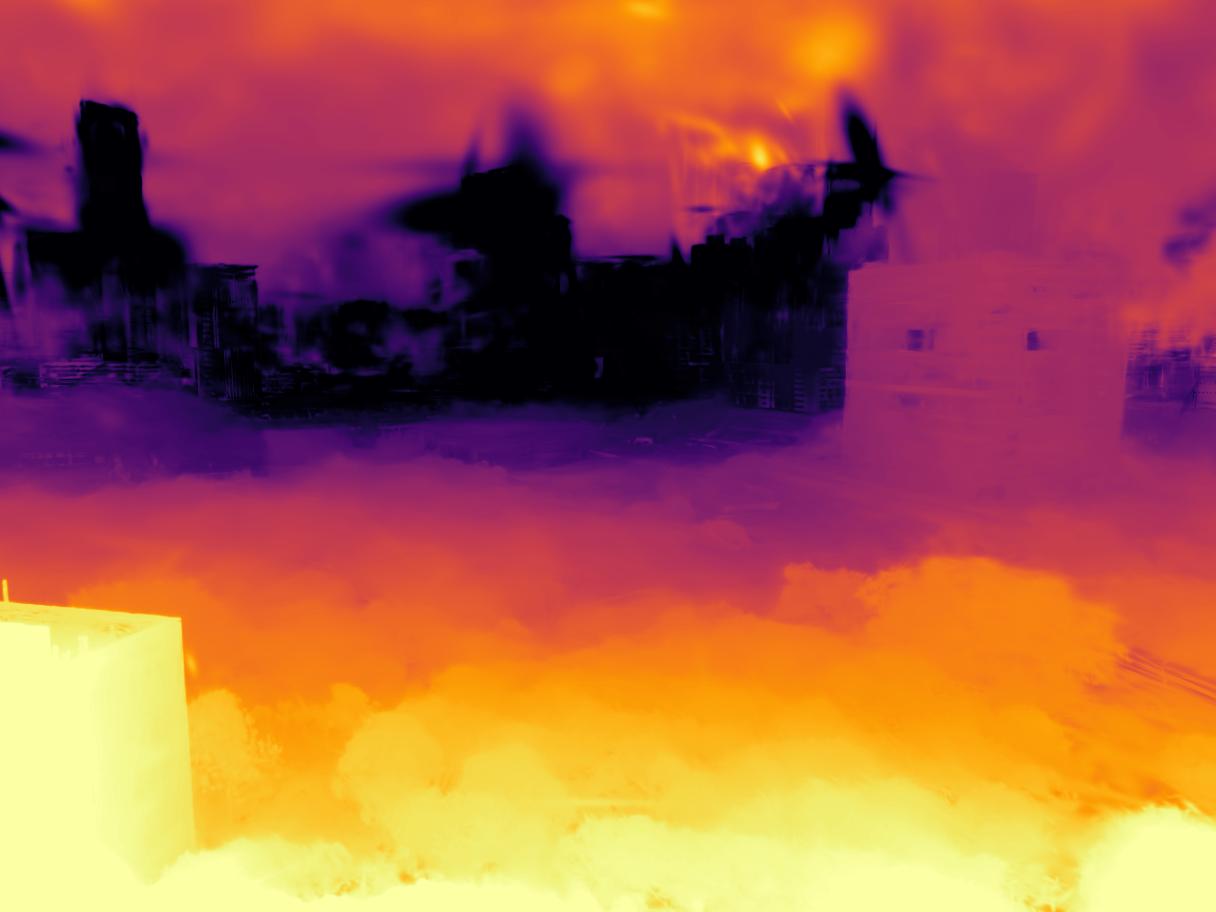}\\
        \multicolumn{2}{c}{(a) Centralized 3DGS} & \multicolumn{2}{c}{(b) Fed3DGS}
    \end{tabular}
    \caption{Top view of the Sci-Art scene and rendered images and depth images with the centralized 3DGS and Fed3DGS.}
    \label{fig:comp-topview}
\end{figure}

We show the top view of the reconstructed scenes by our models in Fig. \ref{fig:topviews}.
The scenes where our method shows the best SSIM are well reconstructed.
However, there are many noisy Gaussians for Quad 6k and Sci-Art.
In Quad 6k, there are many floating Gaussians as noise due to sky pixels, as shown in Fig. \ref{fig:quad}, which is the limitation of 3DGS~\cite{3dgs} because 3DGS with the centralized setting also has the floating Gaussians in Quad 6k (Fig. \ref{fig:quad}~(b)).
In Sci-Art, some Gaussians are sparsely scattered near the scene boundaries to represent the distant buildings.
Compared to a centralized approach, the number of viewpoints of such buildings per each local data is fewer, and the quality of the initial points reconstructed by SfM is poor.
As a result, our method shows somewhat poor results in Sci-Art.
In fact, the centralized 3DGS represents the distant objects well and shows better SSIM and LPIPS, as shown in Fig. \ref{fig:comp-topview} and Tab. \ref{tab:3dgs-res}.

\begin{figure}
    \centering
    \begin{tabular}{cccccc}
        \rotatebox{90}{Winter} &
        \includegraphics[clip,width=0.18\hsize]{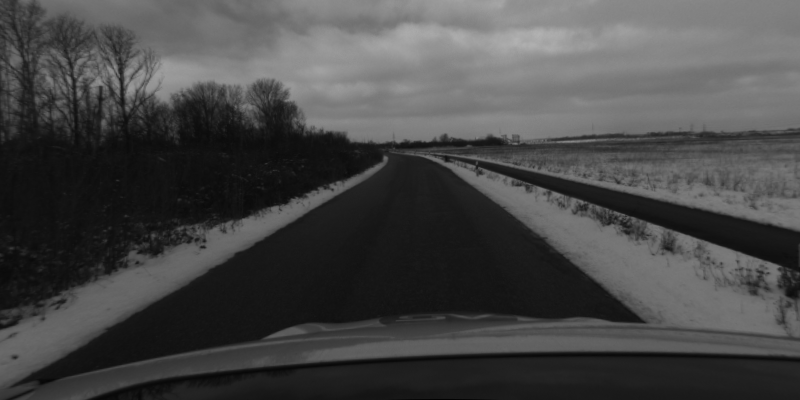} &  
        \includegraphics[clip,width=0.18\hsize]{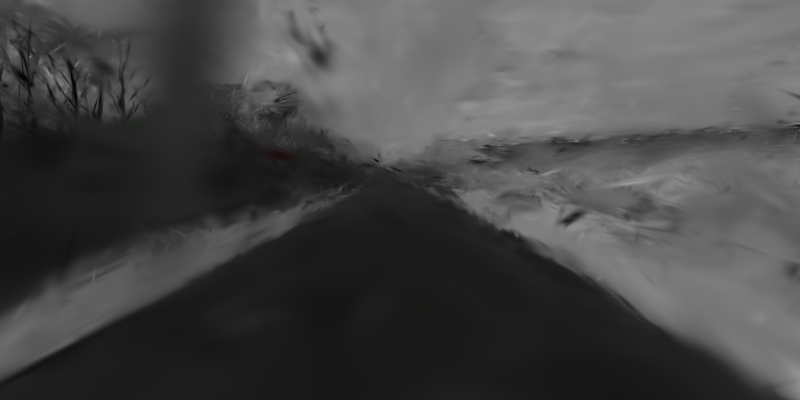} &  
        \includegraphics[clip,width=0.18\hsize]{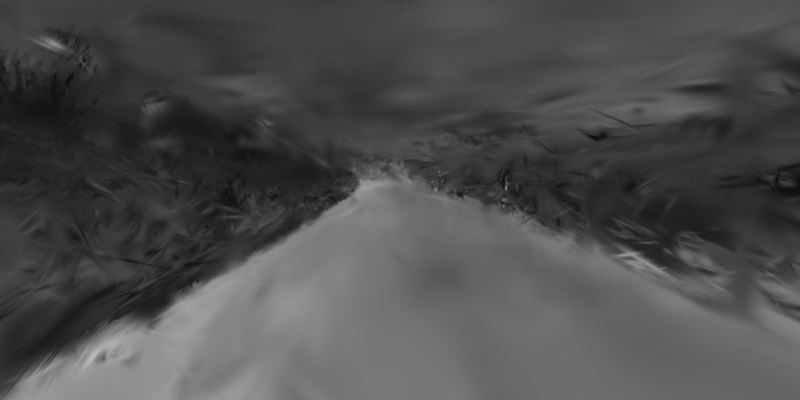} &  
        \includegraphics[clip,width=0.18\hsize]{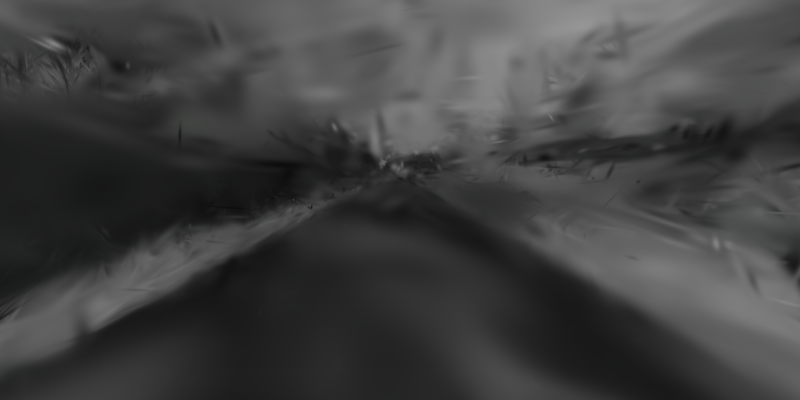} & 
        \includegraphics[clip,width=0.18\hsize]{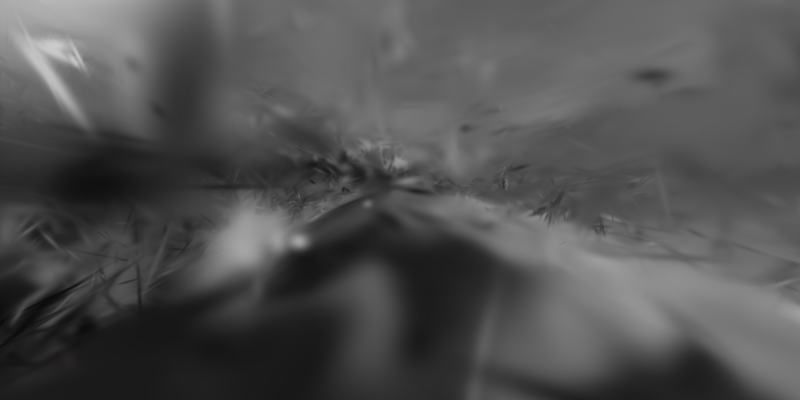} \\
        \rotatebox{90}{Summer} &
        \includegraphics[clip,width=0.18\hsize]{images/rend-images/4seasons/gt_summer.png} &  
        \includegraphics[clip,width=0.18\hsize]{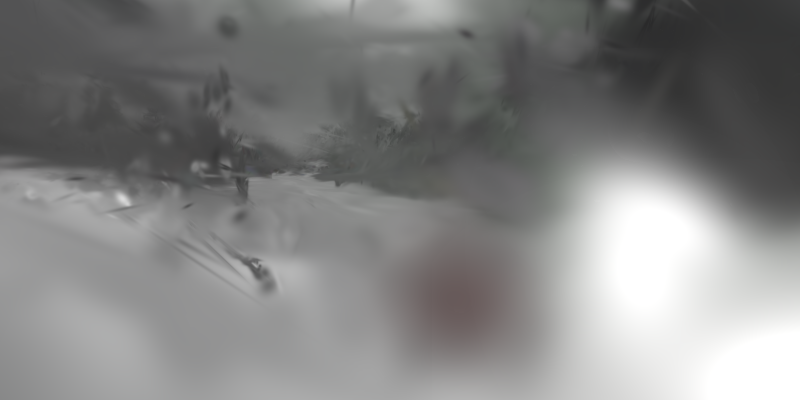} &  
        \includegraphics[clip,width=0.18\hsize]{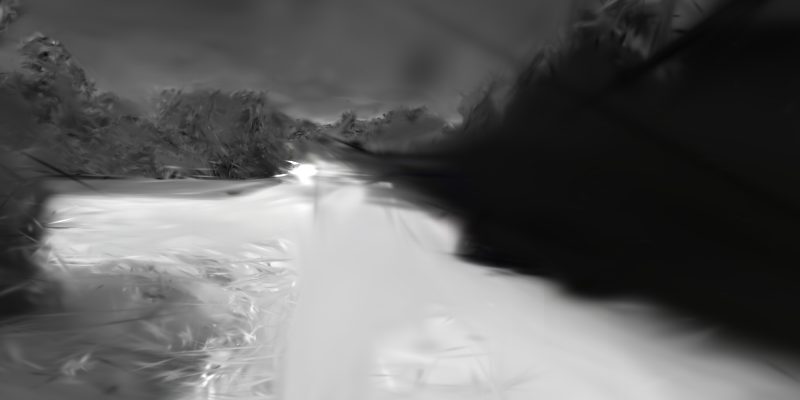} &  
        \includegraphics[clip,width=0.18\hsize]{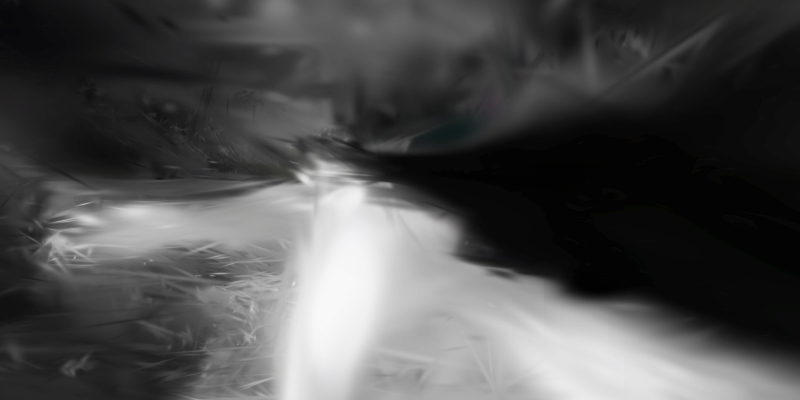} & 
        \includegraphics[clip,width=0.18\hsize]{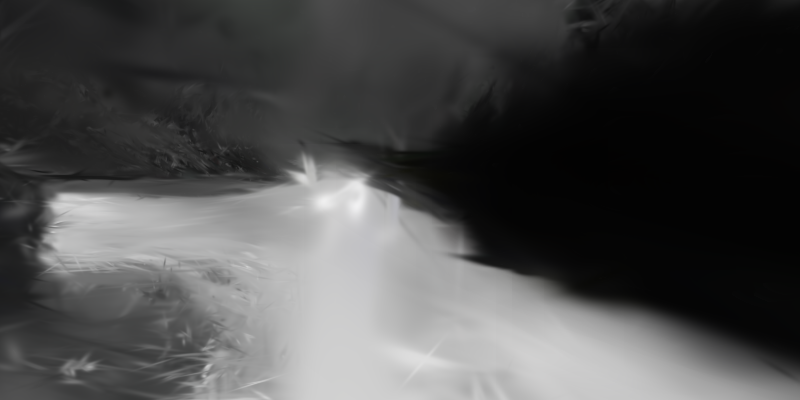} \\
        & (a) Ground-Truth & (b) $\mathcal{G}_g^W$ & (c) $\mathcal{G}_g^S$ & (d) $\mathcal{G}_g^{S\rightarrow W}$ & (e) $\mathcal{G}_g^{W\rightarrow S}$
    \end{tabular}
    \caption{Rendered images in the 4Seasons dataset.}
    \label{fig:4seasons-vis}
\end{figure}

We also show the rendered images through the models trained on 4Seasons in Fig. \ref{fig:4seasons-vis}.
$\mathcal{G}_g^{S\rightarrow W}$ and $\mathcal{G}_g^{W\rightarrow S}$ can represent appearance for both seasons (the color of the road surface is a clear example).
The image rendered by $\mathcal{G}_g^W$ for summer is corrupted because the model attempts to represent the color of the road surface.
It is important to note that some floating Gaussians exist due to pixels with infinite depth, as observed in Quad 6k, which can degrade the quality. Enhancing the performance for such scenes is a potential avenue for future work.

\begin{figure}
    \centering
    \begin{tabular}{cc}
         \includegraphics[clip,width=0.4\hsize]{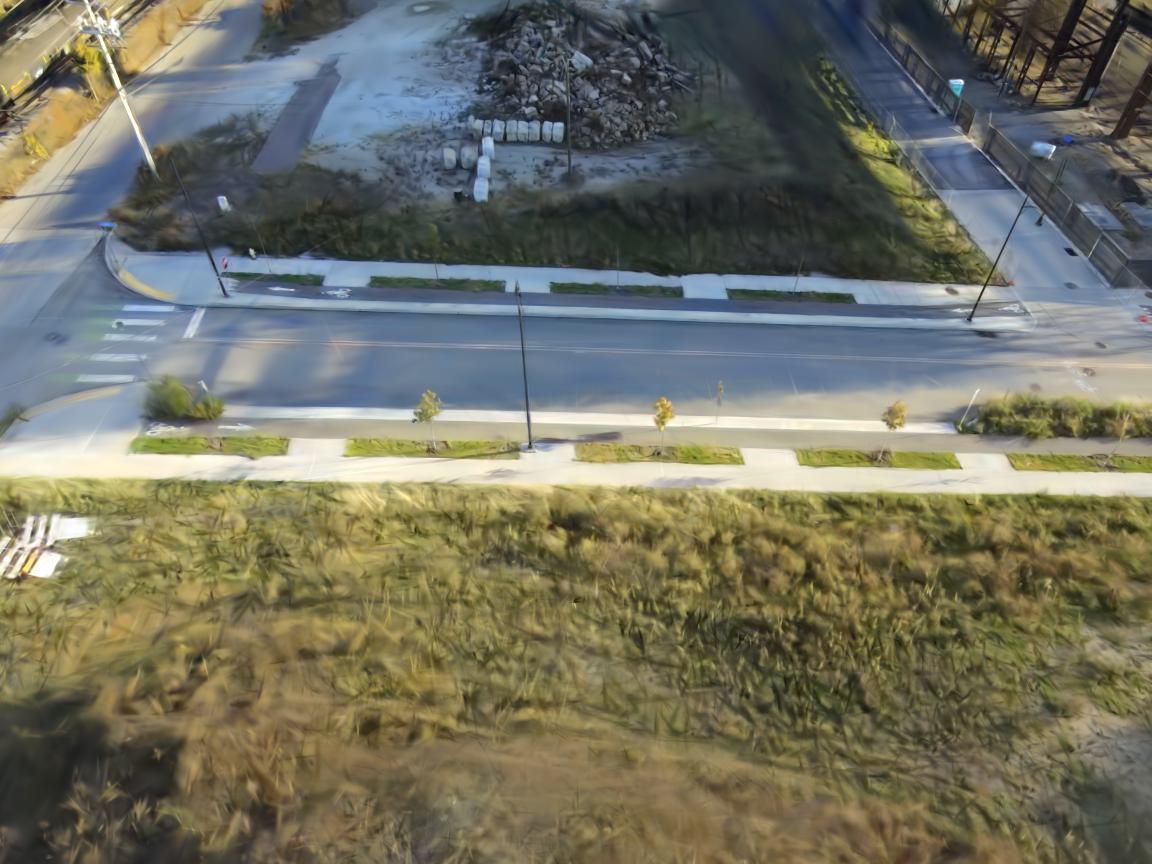} &  
         \includegraphics[clip,width=0.4\hsize]{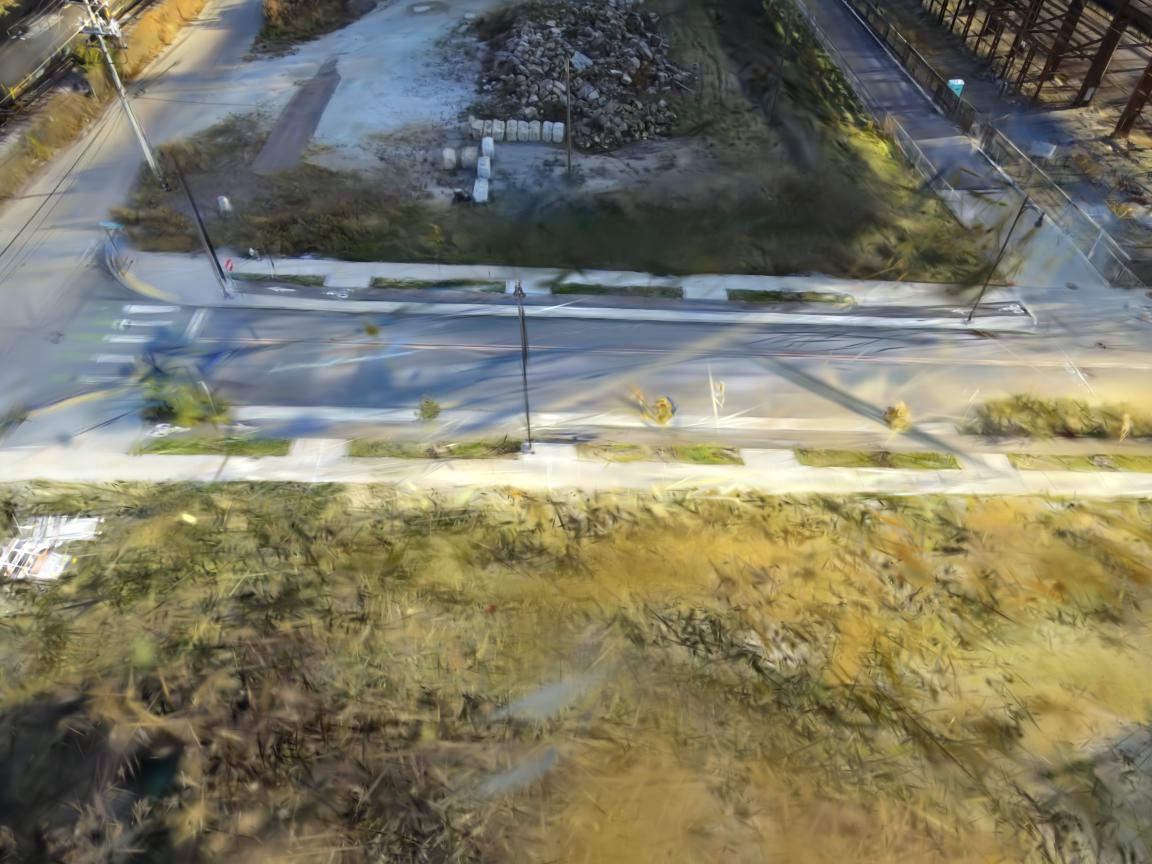} \\
         \includegraphics[clip,width=0.4\hsize]{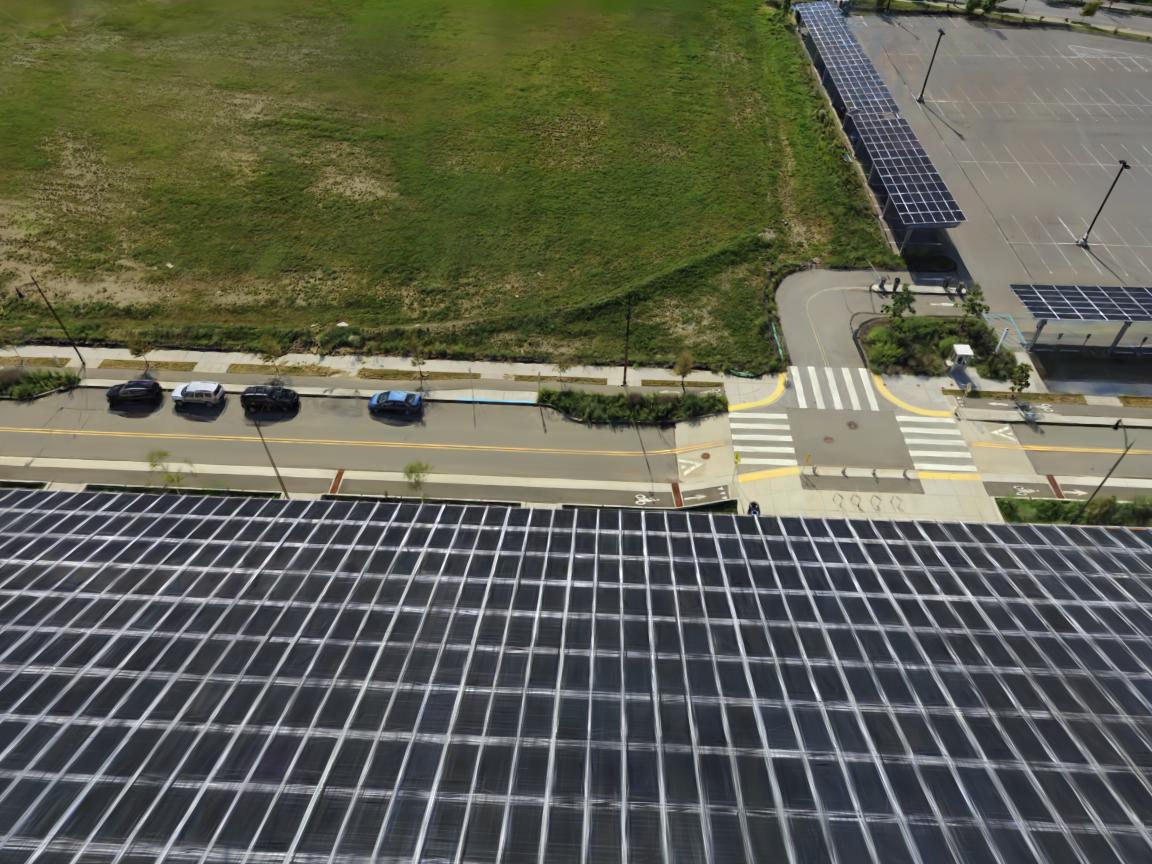} &  
         \includegraphics[clip,width=0.4\hsize]{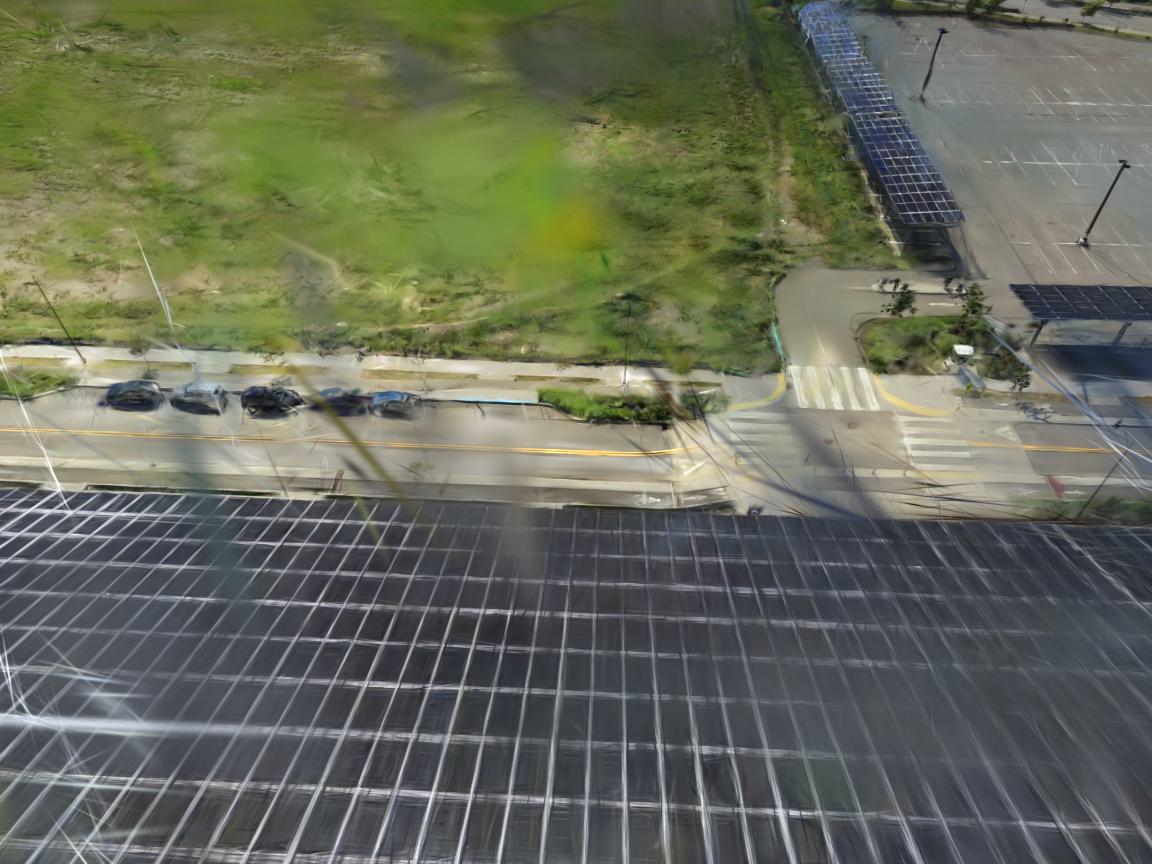} \\
         (a) & (b) 
    \end{tabular}
    \caption{Images rendered by the model trained (a) with eq. \eqref{eq:dist-obj} and (b) with eq. \eqref{eq:simple-dist-obj}.}
    \label{fig:w-wo-cam}
\end{figure}

We visually assess the effectiveness of the additional cameras in the distillation-based model update.
As shown in Fig. \ref{fig:w-wo-cam}, there are some floating Gaussians in the model updated only with the local camera. 
The incorporation of additional cameras contributes to the reduction of these floating Gaussians.

\begin{figure}
    \centering
    \begin{tabular}{ccc}
        \includegraphics[clip,width=0.3\hsize]{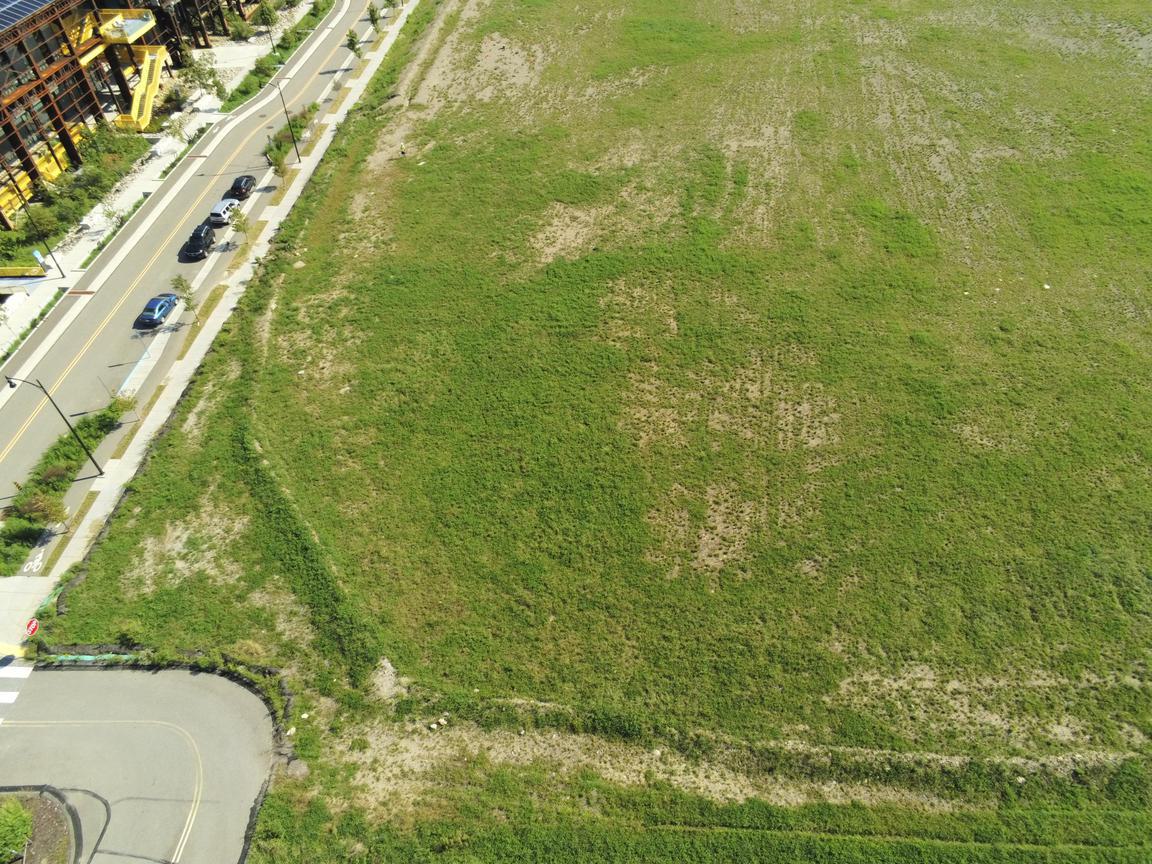} & 
        \includegraphics[clip,width=0.3\hsize]{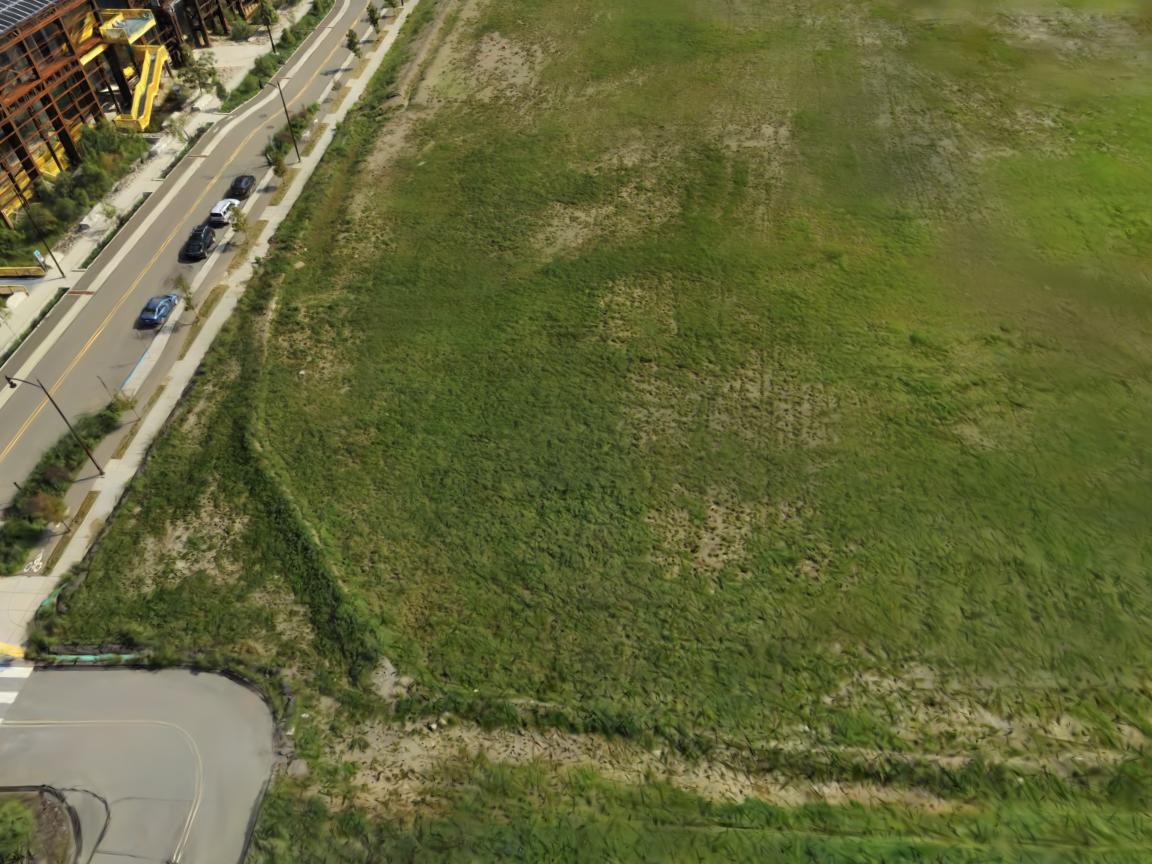} &
        \includegraphics[clip,width=0.3\hsize]{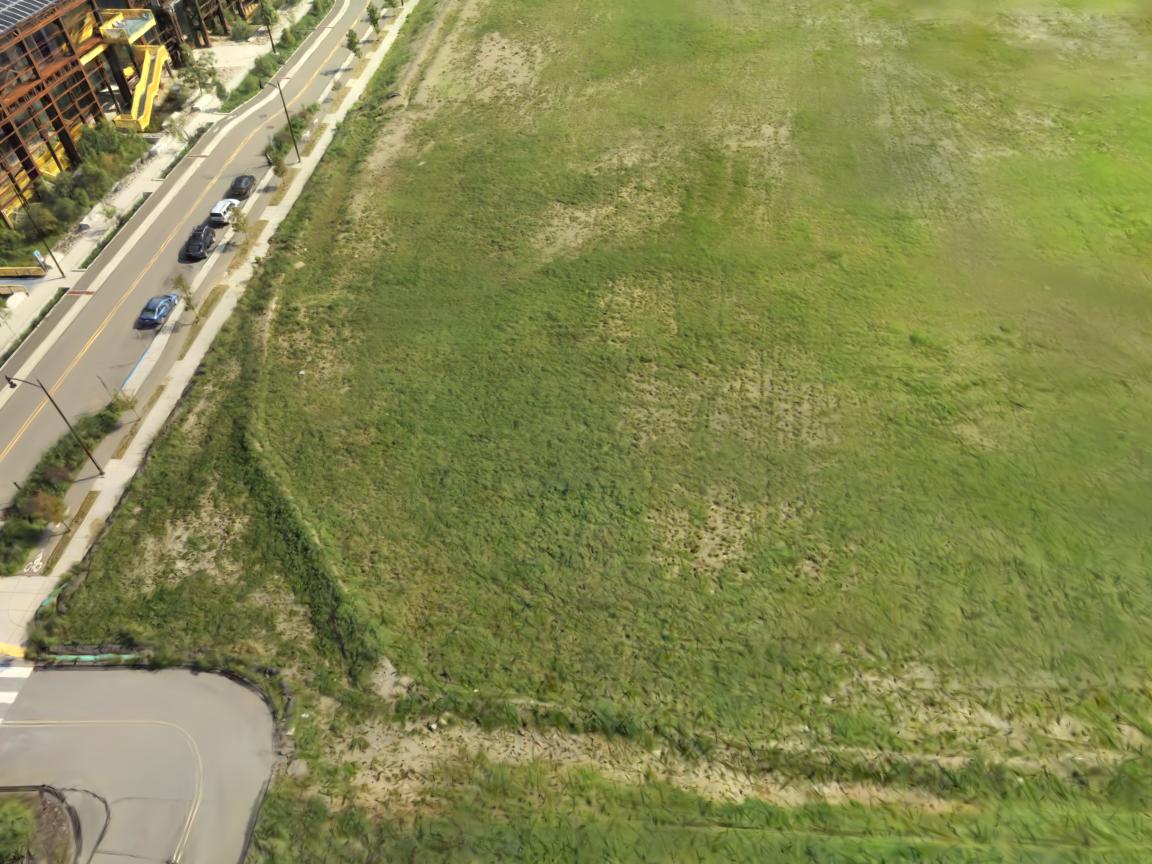}\\
        \includegraphics[clip,width=0.3\hsize]{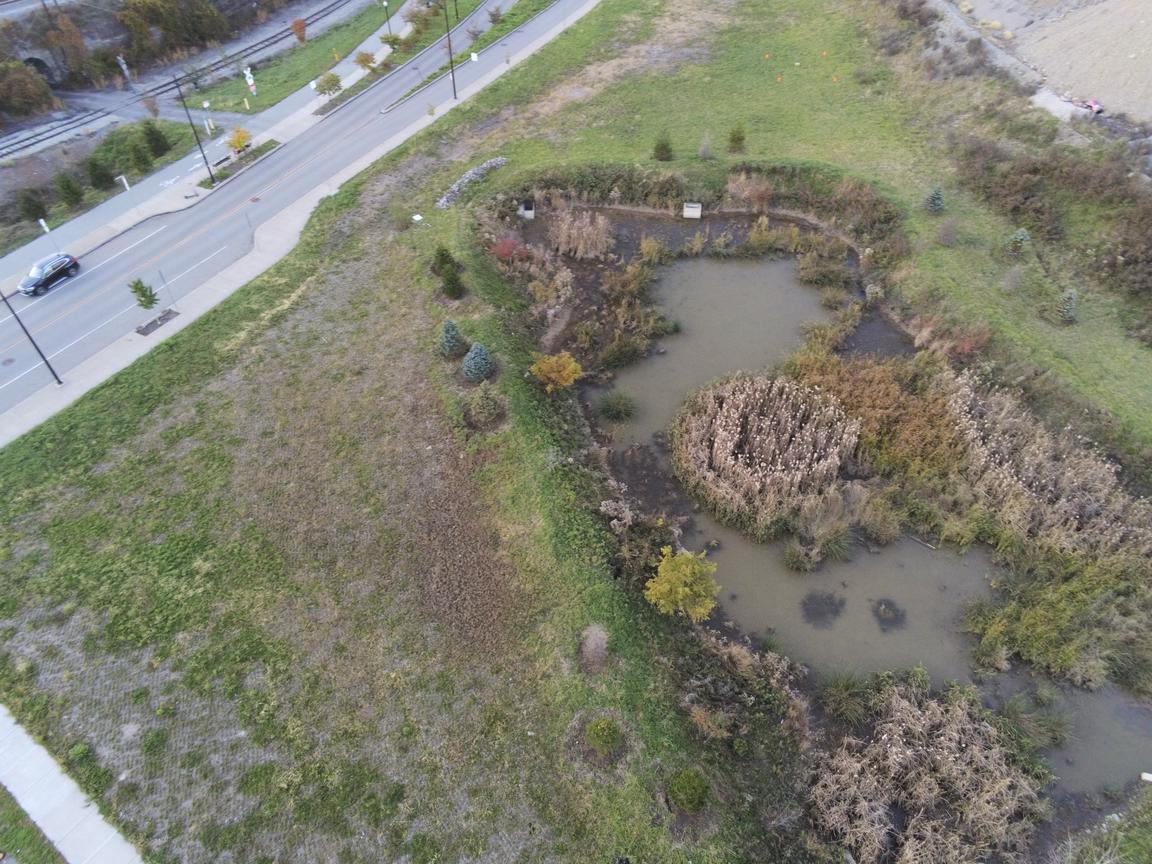} & 
        \includegraphics[clip,width=0.3\hsize]{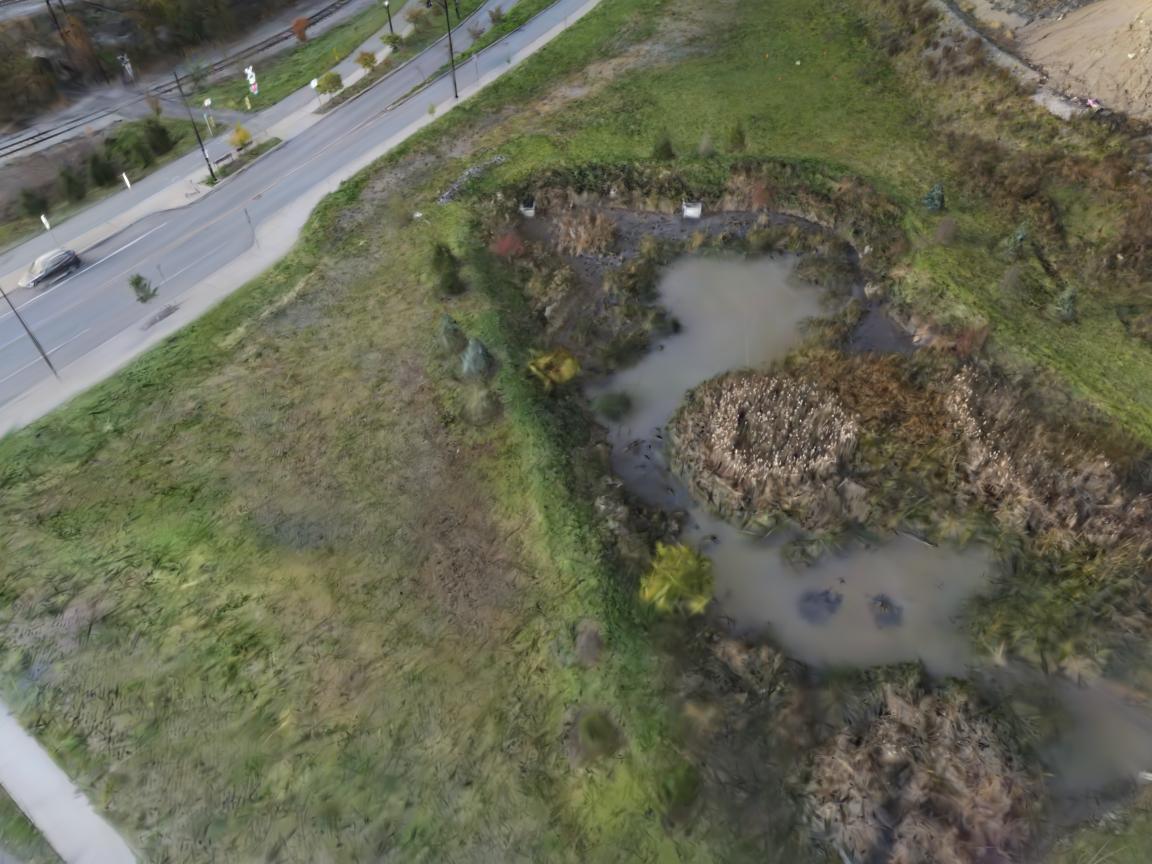} & 
        \includegraphics[clip,width=0.3\hsize]{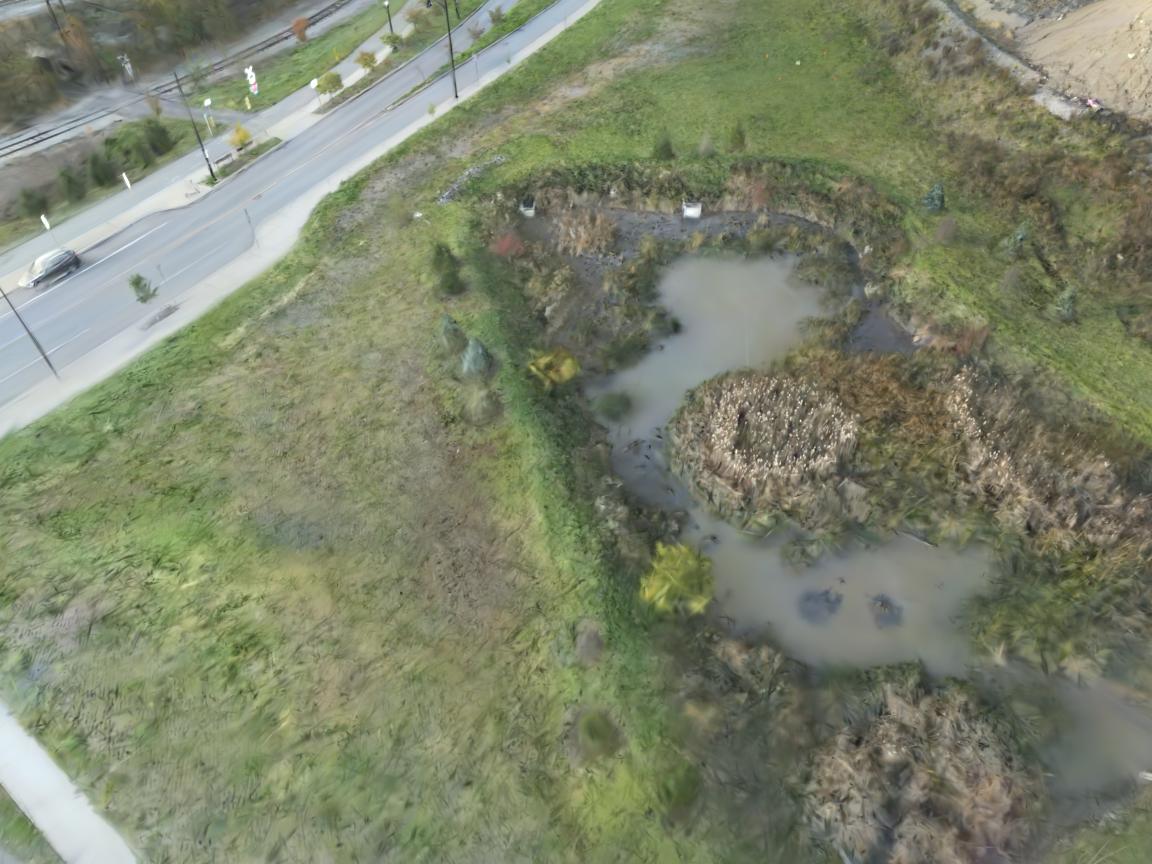} \\ 
        (a) Ground-Truth & (b) w/o $\phi$ & (c) w/ $\phi$
    \end{tabular}
    \caption{Rendered images by Fed3DGS with and without the appearance model $\phi$.}
    \label{fig:vis-appearance}
\end{figure}

We visually evaluate the appearance modeling.
We show the rendered images by Fed3DGS with and without the appearance model $\phi$ in Fig. \ref{fig:vis-appearance}.
As expected, the appearance model successfully adjusts the appearance to the target images.

\section{Additional Ablation Study}
\label{sec:add-ablation}
\subsection{The Effectiveness of Reset Opacity and Entropy Minimization for Other Datasets}
\begin{figure}[t]
    \centering
    \includegraphics[clip,width=0.8\hsize]{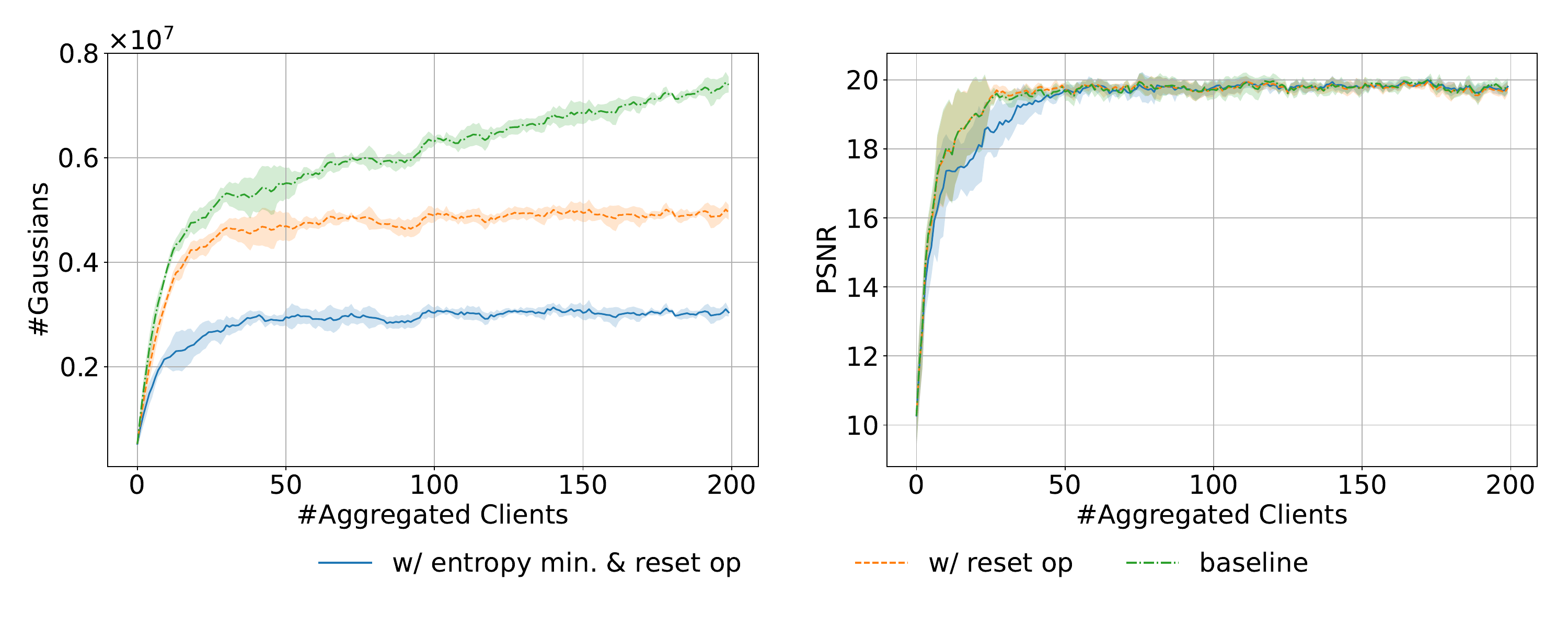} \\
    \includegraphics[clip,width=0.8\hsize]{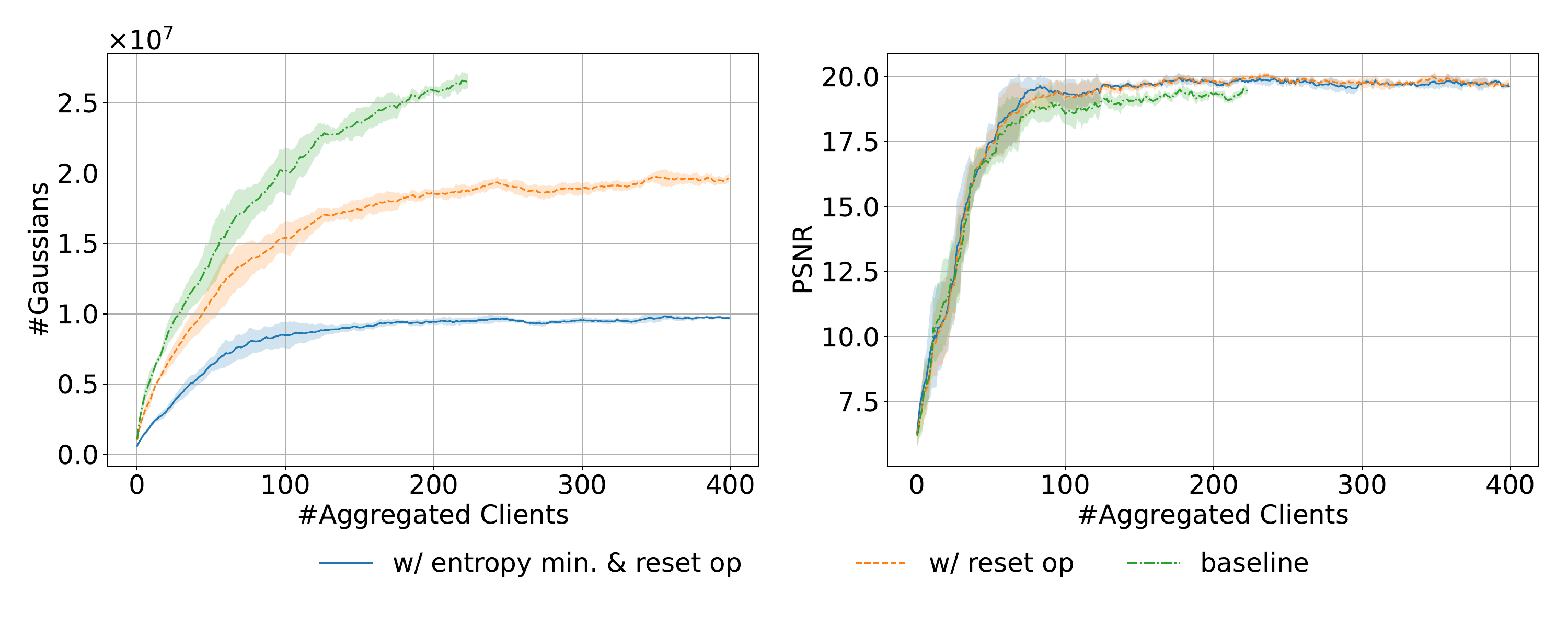}    
    \caption{The number of Gaussians and PSNR with and without reset opacity (reset op) and entropy minimization (entropy min.) on the rubble (upper) and campus (lower) scenes. We show the mean and standard deviation over 5 trials. Note that the results of baseline for Campus raise out-of-memory errors due to the monotonically increasing number of Gaussians.}
    \label{fig:transition-rubble-campus}
\end{figure}

We show the growth of the number of Gaussians and PSNR for the rubble and campus scenes in Fig. \ref{fig:transition-rubble-campus}.
For both datasets, the reset opacity prevents the monotonically increasing number of Gaussians, and the entropy minimization reduces the number of Gaussians while keeping PSNR.

\subsection{Global Pose Alignment}
We evaluate the global pose alignment with Fed3DGS.
We conduct the experiments on the building and rubble scenes.
We use the global model generated in the main experiments.
We randomly select a client model and then add noise to the 3D position and covariance matrix of a local model's Gaussians and local cameras in a range of ($-$20 m, 20 m) and ($-$20$^\circ$, 20$^\circ$) for translation and rotation.
To align the global pose of the client model, we use the alignment method proposed in FedNeRF~\cite{fednerf}.
The optimization procedure, including hyperparameters, is the same as in FedNeRF.

We show the mean of the rotation and translation errors over 10 client models in Tab. \ref{tab:alignment}.
For both datasets, the vision-based alignment works well for out method.
\begin{table}[]
    \centering
    \caption{Rotation and translation errors after global pose alignment.}
    \label{tab:alignment}
    \begin{tabular}{ccc}\toprule
           & \hspace{1em} Rotation Err. ($^\circ$) & \hspace{1em} Translation Err. (m)\\ \hline
Building   &             0.264                     &             1.150                \\
Rubble     &             0.743                     &             0.777                \\ \hline
    \end{tabular}

\end{table}

\subsection{Model Size vs. Data Size}
We show the average of the model sizes and the local data sizes for the building and rubble scenes in Tab. \ref{tab:comp-size}.
Note that the local image data are encoded by JPEG.
The local model size is much smaller than the data size.
In addition, there are ongoing efforts to further reduce the size of 3DGS up to 0.1$\times$ based on recent studies~\cite{lee2023compact,fan2023lightgaussian}.
\begin{table}[]
    \centering
    \caption{Data size and model size averaged over clients.}
    \label{tab:comp-size}
    \begin{tabular}{ccc} \toprule
                  & \hspace{1em}Local Data Size & \hspace{1em}Local Model Size \\ \hline
         Building &      814.4 MB       &    180.7 MB   \\
         Rubble   &      849.2 MB       &    91.77 MB   \\ \hline
    \end{tabular}

\end{table}

\subsection{Comparison between Fed3DGS and FedNeRF}
The results of FedNeRF reported in \cite{fednerf} are obtained with 100 clients, but we use 200 clients in the experiments.
Thus, we report the performance of Fed3DGS with 100 clients for the building and rubble scenes in Tab. \ref{tab:100clients}.
For 100 clients, our method shows better PSNR than FedNeRF.
In fact, as we can see from Figs. \ref{fig:transition} and \ref{fig:transition-rubble-campus}, the performance is saturated around 50--100 clients for both scenes.
\begin{table}[]
    \centering
    \caption{PSNR of the global model updated with 100 clients.}
    \label{tab:100clients}
    \begin{tabular}{ccc} \toprule
                               & \hspace{1em}Building & \hspace{1em}Rubble \\ \hline
        FedNeRF~\cite{fednerf} &   17.51  &  20.12 \\
        Ours                   &   18.43  &  20.51 \\ \hline
    \end{tabular}

\end{table}

\subsection{Comparison between Fed3DGS and Centralized 3DGS}
\label{sec:fed3dgs-vs-3dgs}
We present the results of the centralized 3DGS (\textit{i.e.,} the results of the original 3DGS~\cite{3dgs}) in Tab. \ref{tab:3dgs-res}.
Our approach demonstrates performance comparable to 3DGS.
Comparing 3DGS w/ $\phi$ to 3DGS, our appearance model improves the performance of 3DGS.
In particular, PSNR of 3DGS w/ $\phi$ is comparable to that of Mega-NeRF~\cite{Turki_2022_CVPR} and Switch-NeRF~\cite{mi2023switchnerf}.
This indicates that our appearance modeling works well, and the limited performance of Fed3DGS in terms of PSNR is due to the difficulty of the appearance modeling in the federated learning setting.
In terms of computational efficiency, centralized approaches fail to model the campus and residence scenes due to out-of-memory errors, while Fed3DGS can effectively model them.
The performance of 3DGS for Quad 6k is worse than other methods, such as Mega-NeRF, even when the appearance model is used.
Mega-NeRF incorporates a background model to handle pixels with infinite depth, such as sky pixels, while 3DGS lacks such a model, leading to the suboptimal performance.
Therefore, addressing the modeling of pixels with infinite depth is one of the challenges for 3DGS.

\begin{table*}

\caption{Comparison of Fed3DGS with 3DGS~\cite{3dgs} 3DGS w/ $\phi$ denotes 3DGS with the proposed appearance modeling, and \texttt{OOM} denotes the failure case due to out-of-memory errors.}
\label{tab:3dgs-res}
\resizebox{\textwidth}{!}{
\begin{tabular}{lccccccccc}
\toprule 
&\multicolumn{3}{c}{Building} & \multicolumn{3}{c}{Rubble} & \multicolumn{3}{c}{Quad 6k} \\
& $\uparrow$PSNR & $\uparrow$SSIM & $\downarrow$LPIPS & $\uparrow$PSNR & $\uparrow$SSIM & $\downarrow$LPIPS & $\uparrow$PSNR & $\uparrow$SSIM & $\downarrow$LPIPS \\ \hline

\multicolumn{4}{l}{\small\textit{- Centralized Training}}\\
\hspace{1em}3DGS~\cite{3dgs}\xspace & 18.56 & 0.629 & 0.401
& 21.50 & 0.681 & 0.359
& 15.62 & 0.538 & 0.490  \\
\hspace{1em}3DGS w/ $\phi$\xspace & 21.11 & 0.688 & 0.337
& 23.47 & 0.716 & 0.332
& 16.20 & 0.541 & 0.493  \\
\hline

\multicolumn{4}{l}{\small\textit{- Federated Learning}} \\
\hspace{1em}Fed3DGS\xspace & 18.66 & 0.602 & 0.362
& 20.62 & 0.588 & 0.437
& 15.41 & 0.528 & 0.485 \\

\toprule
&\multicolumn{3}{c}{Residence} & \multicolumn{3}{c}{Sci-Art} & \multicolumn{3}{c}{Campus} \\
& $\uparrow$PSNR & $\uparrow$SSIM & $\downarrow$LPIPS & $\uparrow$PSNR & $\uparrow$SSIM & $\downarrow$LPIPS & $\uparrow$PSNR & $\uparrow$SSIM & $\downarrow$LPIPS \\ \hline

\multicolumn{4}{l}{\small\textit{- Centralized Training}} \\
\hspace{1em}3DGS~\cite{3dgs}\xspace & \multicolumn{3}{c}{\texttt{\#\#\#OOM\#\#\#}} 
& 18.34 & 0.782 & 0.275
& \multicolumn{3}{c}{\texttt{\#\#\#OOM\#\#\#}} \\

\hspace{1em}3DGS w/ $\phi$ \xspace &  \multicolumn{3}{c}{\texttt{\#\#\#OOM\#\#\#}}
& 24.78 & 0.834 & 0.248
&  \multicolumn{3}{c}{\texttt{\#\#\#OOM\#\#\#}}   \\
\hline

\multicolumn{4}{l}{\small\textit{- Federated Learning}} \\
\hspace{1em}Fed3DGS\xspace & 20.00 & 0.665 & 0.344
& 21.03 & 0.730 & 0.335
& 21.64 & 0.635 & 0.436 \\

\bottomrule
\end{tabular}
}

\end{table*}

\end{document}